\documentclass[]{interact}

\usepackage[hyphens]{url} 

\usepackage{comment}
\usepackage{enumerate}

\DeclareMathOperator*{\argmax}{argmax}
\DeclareMathOperator*{\argmin}{argmin}

\newcommand{\minimize}{\mathop{\rm minimize}\limits}
\usepackage{siunitx}
\usepackage{mathtools}

\usepackage{algorithm} 
\usepackage{algpseudocode} 

\usepackage{color}    

\usepackage{epstopdf}
\usepackage[caption=false]{subfig}

\usepackage[numbers,sort&compress]{natbib}
\bibpunct[, ]{[}{]}{,}{n}{,}{,}
\renewcommand\bibfont{\fontsize{10}{12}\selectfont}
\makeatletter
\def\NAT@def@citea{\def\@citea{\NAT@separator}}
\makeatother

\theoremstyle{plain}

\theoremstyle{definition}

\theoremstyle{remark}

\begin{document}

\articletype{ARTICLE TEMPLATE}









\title{
Active Exploration based on Information Gain by Particle Filter \\for Efficient Spatial Concept Formation
}

\author{
    Akira Taniguchi$^{a \ast}$\thanks{$^\ast$Corresponding author. Email: a.taniguchi@em.ci.ritsumei.ac.jp\vspace{6pt}},
    Yoshiki Tabuchi$^{b}$, Tomochika Ishikawa$^{a}$,\\
    Lotfi El Hafi$^{c}$, Yoshinobu Hagiwara$^{c}$, and Tadahiro Taniguchi$^{a}$\\
    \vspace{6pt}
    $^{a}${\em{College of Information Science and Engineering, Ritsumeikan University, Shiga, Japan}};\\
    $^{b}${\em{Graduate School of Information Science and Engineering, Ritsumeikan University, Shiga, Japan}};\\
    $^{c}${\em{Research Organization of Science and Technology, Ritsumeikan University, Shiga, Japan}}\\
}

\maketitle

\begin{abstract} 

Autonomous robots need to learn the categories of various places by exploring their environments and interacting with users. 
However, preparing training datasets with linguistic instructions from users is time-consuming and labor-intensive. 
Moreover, effective exploration is essential for appropriate concept formation and rapid environmental coverage. 
To address this issue, we propose an active inference method, referred to as spatial concept formation with information gain-based active exploration (SpCoAE) that combines sequential Bayesian inference using particle filters and information gain-based destination determination in a probabilistic generative model. 
This study interprets the robot's action as a selection of destinations to ask the user, `What kind of place is this?' in the context of active inference.
This study provides insights into the technical aspects of the proposed method, including active perception and exploration by the robot, and how the method can enable mobile robots to learn spatial concepts through active exploration.
Our experiment demonstrated the effectiveness of the SpCoAE in efficiently determining a destination for learning appropriate spatial concepts in home environments. 
\end{abstract}



\section{Introduction}
\label{sec:introduction}

\begin{figure}[!tb]
    \centering
	\includegraphics[width=0.6\linewidth]{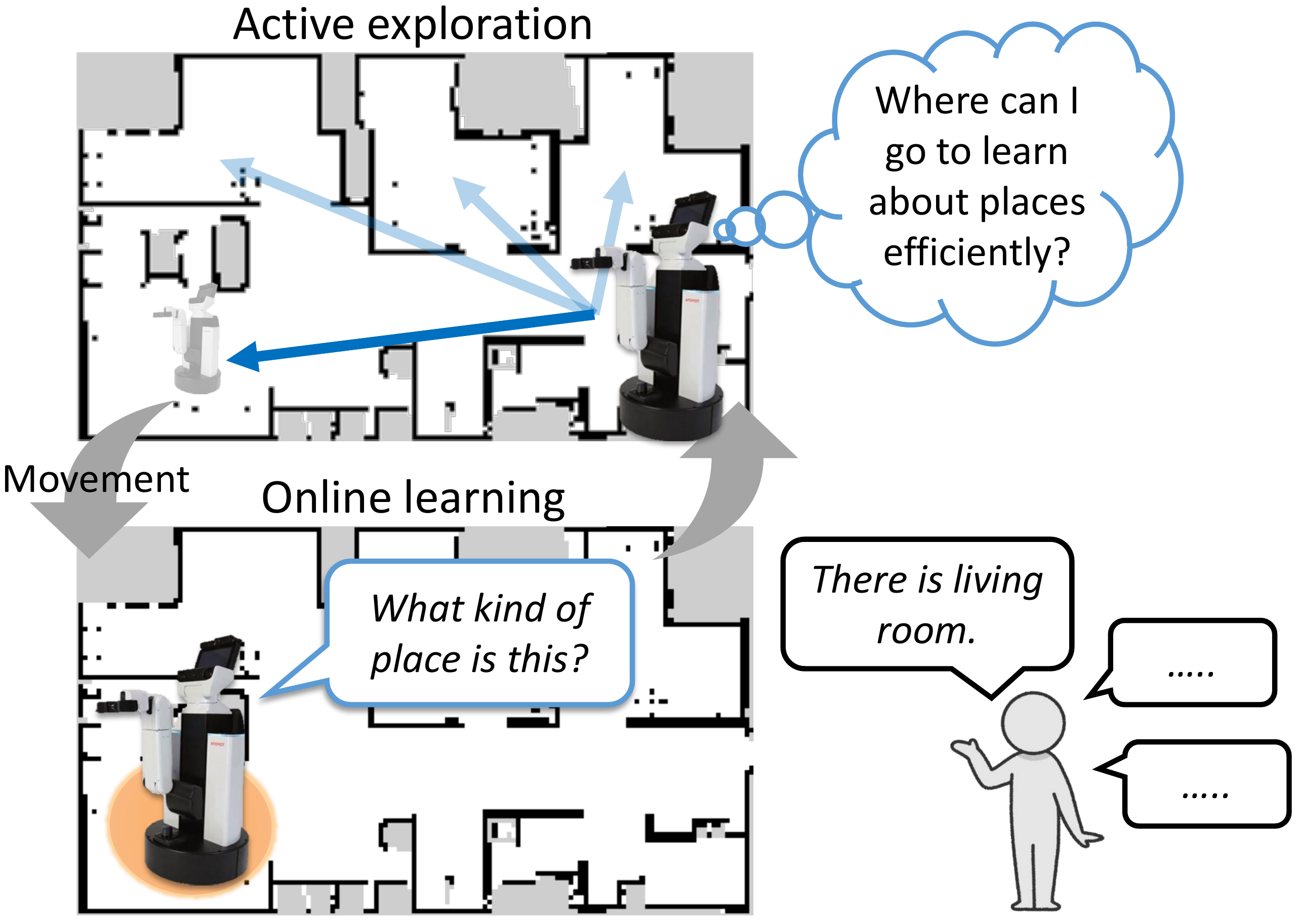}
	\caption{
    	Active exploration and learning of spatial concepts by the robot.
    	The robot explores a position that would provide the most information and reduce uncertainty.
    	First, the robot decides which destination to explore from the candidate destinations in the environment.
        After moving, the robot asks the user, `What kind of place is this?' to observe the position and words.
		Subsequently, the robot learns spatial concepts based on the observations.
		Following this, the robot decides where to go next based on the spatial concepts it has formed.
		The above process is repeated. 
	}
	\label{fig:master_thesis_overview}
\end{figure} 

Service robots deployed in office and home settings, where they should autonomously categorize and identify various locations through interactions with their surrounding environment and users.
Semantic mapping, which assigns a label to an environmental map~\cite{kostavelis2015semantic, Garg2020}, was recently proposed.
In contrast, spatial concept-based approaches allow unsupervised learning frameworks to categorize unknown places and flexible word assignments from user-language interactions~\cite{ataniguchi_IROS2017,ataniguchi2020spcoslam2}.
The spatial concept is defined as the abstracted categorical knowledge of locations derived from the multimodal observations gathered through spatial perception in robots.
However, to collect the training data, the user follows and manipulates the robot, moves, and speaks multiple times at each location to be taught.
To solve these problems, robots should actively determine and move autonomously toward their destinations.
Learning through active exploration requires the user to speak only when asked by the robot, which reduces the burden on the user.

Our challenge is task-independent and interactive knowledge acquisition, in which the robot asks the user questions.
At the interface between artificial intelligence and computational neuroscience, \textit{free-energy principle (FEP)-based }\textit{active inference (AIF)}~\cite{friston2017active} has gained attention as an approach through which agents actively explore and acquire knowledge.
The expected free energy based on AIF theoretically encompasses \textit{information gain (IG)}, which is commonly used in traditional active perception/learning in robotics, and provides further inspiration.
The application of AIF theories in robotics has gained increasing importance~\cite{friston2021woldmodel,Lanillos2021}.
AIF encompasses active exploration and online learning loops, which serve as the foundation for our study.
In the field of robotics, studies have been conducted on active exploration for \textit{simultaneous localization and mapping (SLAM)}~\cite{thrun2005probabilistic}, that is, active SLAM~\cite{Stachniss2005activeslam,Mu2016,Placed2022} and active perception/learning for multimodal categorization~\cite{Taniguchi2015yoshino,Yoshino2021}. 
Our study integrated these approaches, leading to active semantic mapping~\cite{Veiga2019,Chaplot2021}.
This study differs from \textit{vision-and-language navigation (VLN)}~\cite{anderson2018vision, Chen2020a, park2022vlnsurvey}, which uses task-dependent knowledge acquisition without AIF as its theoretical foundation. 
We focus on active exploration inspired by AIF for grounding a spatial lexicon in a mobile robot.

The learning procedure that uses the active exploration approach is illustrated in Figure~\ref{fig:master_thesis_overview}.
While moving in any environment, robots can acquire multimodal data on places, such as the names spoken by users and their locations.
The robot can actively explore uncertain locations and ask the user questions.
This allowed the robot to learn spatial concepts efficiently.
However, learning spatial concepts through active exploration has not been achieved in previous studies~\cite{ataniguchi_IROS2017,ataniguchi2020spcoslam2}.
In particular, it is important to conduct efficient exploration that leads to appropriate concept formation and quickly encompasses the environment rather than haphazard exploration.

This study addresses three main aspects: learning accuracy, learning efficiency, and movement efficiency.
\begin{enumerate}[(i)]
    \item \textbf{Learning accuracy}:
    Selecting data in an order that allows learning to be stable and accurate is important.
    In online unsupervised learning, such as particle filtering, the order of the observed data affects the estimation accuracy~\cite{ulker2010sequential}.
    \item \textbf{Efficiency of learning}:
    In active exploration, it is efficient to adapt to the environment quickly.
    It is required to reach a sufficiently accurate learning result that encompasses the environment in a few steps.
    To reduce the uncertainty of learning results concerning the environment, learning by exploring using information--theoretical criteria is expected to be effective~\cite{Taniguchi2015yoshino,Yoshino2021}.
    \item \textbf{Efficiency of movement}:
    For mobile robots, the costs associated with travel distance must also be considered.
    Exploring a distant location requires more time to move.
\end{enumerate}

This study aims to improve the efficiency of learning spatial concepts through autonomous active exploration using a mobile robot.
Therefore, we propose an AIF-inspired method that combines sequential Bayesian inference based on particle filters and destination determination based on IG called \textit{spatial concept formation with IG-based active exploration (SpCoAE)}.
By considering the active exploration of spatial concept formation as an optimization problem, we realize autonomous decision-making for the questioning behavior of the robot.

This study also considers aspects of a constructive approach in cognitive developmental robotics and symbolic emergence in robotics~\cite{asada2009cognitive,cangelosi2015developmental,taniguchi2018TCDSsurvey}.
A \textit{probabilistic generative model (PGM)} for spatial concept formation was proposed as a world model~\cite{Ha2018}.
Learning and inferring world models is important for building human-like intelligent machines~\cite{friston2021woldmodel}.
The implementation of AIF based on FEP, which is proposed as a fundamental principle of the brain~\cite{fep,fep1,fep2,fep3,friston2017active}, and demonstrating its feasibility is challenging in the robotics field~\cite{Lanillos2021}.

The main contributions of this study are as follows.
\begin{enumerate}[1.]
    \item We show that a robot can probabilistically relate flexible location-related words to a map through place categorization by actively determining the destination and asking the user there.
    \item We realize active inference for efficient spatial concept formation by leveraging the posterior distribution estimated by particle filter-based online learning for IG-based active exploration.
    \item We demonstrate that the utility function, including the IG and travel distance, achieves more efficient exploration compared to the baselines, suggesting a correspondence with the expected free energy.
\end{enumerate}

The remainder of this paper is organized as follows:
Section~\ref{sec:related_work} presents research on spatial concept acquisition and active exploration.
Section~\ref{sec:aif} describes the background of AIF based on FEP.
Section~\ref{sec:proposed} describes the proposed method, SpCoAE, and the online learning of spatial concepts.
Section~\ref{sec:exp1} presents experiments performed using a simulator in multiple home environments.
Section~\ref{sec:exp2} discusses experiments performed in real environments.
Finally, Section~\ref{sec:conclusion} concludes the paper.

\section{Related work}
\label{sec:related_work}

As research related to this study, we describe spatial perception, spatial concept formation and semantic mapping in Section~\ref{sec:related_work:semantic} and active spatial perception, exploration, and learning in Section~\ref{sec:related_work:active}.

\subsection{Spatial perception and semantic mapping}
\label{sec:related_work:semantic}

Recently, semantic mapping has been emphasized~\cite{kostavelis2015semantic, Garg2020}.
However, several studies have provided preset labels for specific map areas.
For example, LexToMap~\cite{rangel2016lextomap} assigns convolutional neural network-recognized lexical labels to a topological map.
Voxblox++~\cite{Grinvald2019} and Kimera~\cite{Rosinol2021} constructed dense metric-semantic maps using RGBD (or dense-stereo) cameras.
By contrast, our approach allows unsupervised learning to categorize unknown places and flexible word assignments.

The robot must learn the spatial concepts online to select its next destination.
In \textit{online spatial concept formation and lexical acquisition with SLAM (SpCoSLAM)}~\cite{ataniguchi_IROS2017,ataniguchi2020spcoslam2}, online learning is achieved by estimating parameters representing spatial concepts using particle filters.
Incidentally, neuroscientific findings suggest that spatial cognition and inference take place in the hippocampal formation of the brain~\cite{Tolman1948a, Okeefe1978placecells}.
Online learning with models such as spatial concept formation has been suggested to be consistent with the function of the hippocampal formation~\cite{Taniguchi2021hpf-pgm}.
Therefore, similar to SpCoSLAM, the proposed method learns spatial concepts through online learning using particle filters.

Considering the burden on the user, the robot was required to explore the environment by moving actively.
Thus far, the applications developed related to spatial concepts include 
action selection for object tidy-up~\cite{ataniguchi2020TidyUpHere}, 
\textit{spatial concept-based navigation by using speech instructions (SpCoNavi)}~\cite{ataniguchi2020spconavi}, 
knowledge transfer across multiple environments~\cite{Katsumata2020SpCoMapGAN,Katsumata2021slamgan,Hagiwara2021transfer} and
the relative location concept formation~\cite{Sagara2021}.
However, previous studies on spatial concept learning have used passive learning methods for robots, in which a user manipulates the robot or the robot moves in the environment by following the user.
The IG-based active exploration method proposed in this study has the potential to be applied to these PGMs.

The VLN is the latest applied research that focuses on the boundary area between computer vision and natural language processing for mobile robots~\cite{anderson2018vision, Chen2020a}.
Typical conventional VLNs provide detailed linguistic texts that specify a route to a goal. Accordingly, they performed navigation task-oriented knowledge acquisition through numerous trial-and-error attempts using benchmark simulators~\cite{wang2019vln,park2022vlnsurvey}.
By contrast, our approach to spatial concept formation allows for non-task-oriented spatial knowledge acquisition, including bottom-up lexical acquisition, even from small amounts of real-world data.

\subsection{Active spatial perception, exploration, and learning}
\label{sec:related_work:active}

Active SLAM selects the next destination when the robot performs SLAM in an environment~\cite{Stachniss2005activeslam,Placed2022}.
IG-based active SLAM~\cite{Stachniss2005activeslam} uses entropy to determine move destinations such that the uncertainty is reduced with FastSLAM, a particle-filter-based online SLAM method~\cite{montemerlo2003fastslam,gridbasedfastslam2005}.
Owing to the numerous candidate destinations, a frontier approach~\cite{frontier} was used to limit the candidate search points.
In another approach, graph structures have been proposed for fast exploration~\cite{Mu2016,Placed2021}.
Active Neural SLAM~\cite{Chaplot2020} and hierarchical AIF-based SLAM~\cite{Catal2021a} have also been used for deep learning-based SLAM.
However, active SLAM does not involve learning place-related words or categories.

Active semantic mapping is a challenging task that involves efficiently exploring space while accurately capturing the semantics of the environment~\cite{Placed2022}.
\textit{Self-supervised embodied active learning (SEAL)} uses perception models trained on internet images for active exploration policies~\cite{Chaplot2021}, while our study employs unsupervised learning without pre-training datasets.
Veiga et al. proposed information-reward models based on partially observable Markov decision processes~\cite{Veiga2019}. 
Unlike room-specific models, our study employed a unified model for spatial concept uncertainty. 
Instead of defining rewards and policies~\cite{Chaplot2021,Veiga2019}, our study directly determines actions using IG based on PGM.
Semantic octree mapping and Shannon mutual information computations were proposed for autonomous robot operations in unstructured and unknown environments~\cite{Asgharivaskasi2023}.
This study is positioned as an active semantic mapping approach that learns through active exploration to make a linguistic sense of the map, rather than mapping.

Active robot learning approaches for language acquisition and understanding have been developed.
Efficient natural language understanding through interactive spatial concept learning that considers user instructions and the environment is important~\cite{Paul2018,Patki2019}.
Active learning was introduced to estimate the command ambiguity in tabletop object manipulation in robot language acquisition~\cite{Sugiura2010}.
Efficient cross-situational object-word learning was achieved by actively selecting the order of the training samples~\cite{chen2016experimental}.
These studies have implications for constructive models for children autonomously acquiring language.
This study aims to achieve efficient spatial language acquisition by robots through active action selection.

Online learning is appropriate for learning through active exploration because new data are obtained each time.
Active perception/learning, in which a robot actively selects the next piece of information to observe when recognizing object categories, has been proposed~\cite{Taniguchi2015yoshino,Yoshino2021}.
Based on a \textit{multimodal hierarchical Dirichlet process (MHDP)}~\cite{nakamura2011}, which is a hierarchical Bayesian model of multimodal categorization, a robot selects an action corresponding to a sensory modality such as tactile, visual, and auditory modalities.
However, MHDP-based active perception/learning methods~\cite{Taniguchi2015yoshino,Yoshino2021} use a batch-learning algorithm based on Gibbs sampling.
These conventional methods compute the IG by Monte Carlo approximation with new sampling, whereas the proposed method uses the results of online estimation using particle filters to compute IG.

Many active-action selection methods use the IG maximization criterion, which is known to satisfy submodularity~\cite{Taniguchi2015yoshino}.
Obtaining words related to a place from a user at the position where the IG is maximized is expected to efficiently reduce the uncertainty in spatial concept formation.

The latest research on active exploration and visual goal navigation leverages semantics as an approach to learning through exploration based on curiosity and other factors~\cite{Chaplot2020c,Georgakis2022iclr}.
In addition, recent studies on VLN have used deep and reinforcement learning~\cite{anderson2018vision, Chen2020a}. 
By contrast, the proposed method is an unsupervised learning approach based on a PGM; it conducts active exploration using probabilistic inference based on information-theoretic criteria.
The proposed method uses abstracted locations in the ground language, while ensuring extensibility to vision.

In the field of embodied computer vision, curiosity, novelty, coverage, and reconstruction have been defined independently for exploration~\cite{Ramakrishnan2020a}.
In contrast, AIF can naturally combine these indicators into a single principle.
In addition, applications of AIF to robots include the estimation and control of body movements~\cite{Oliver2022} and multimodal affective human-robot interactions~\cite{Horii2021}.
This study is also significant because it applies AIF to a robot that collects data in its environment by moving with an embodied body.

\section{Foundational Concepts: Active inference based on free energy principle with world model} 
\label{sec:aif}

The FEP provides a unified explanation for the mechanisms of action, perception, and learning~\cite{fep}.
Various theories of the brain, such as the Bayesian brain hypothesis~\cite{doya2007bayesian} and motor control, can be explained in a unified manner by the FEP~\cite{fep1}.
When humans estimate the state of the external world based on their perceptions, they actively select the action most likely to provide the most information regarding that state (i.e., they may maximize the expected free energy)~\cite{fep2,fep3}.
This is called the AIF~\cite{friston2017active}, which infers what to do next to resolve the uncertainty.

A robot should acquire knowledge about its environment through active inference~\cite{friston2021woldmodel}.
Forming a representation of the world (i.e., an internal representation of the perception of the world) is necessary to appropriately promote action generation.
This abstract model of the environment is called the world model~\cite{Ha2018,Hafner2019,Okada2020a,Ball2020,friston2021woldmodel}.
In particular, predictive coding~\cite{Rao1999}, which learns to predict future observations, can efficiently reduce uncertainty in the knowledge of the environment. 
In this study, we constructed a PGM for spatial concept formation as a world model.

FEP is realized by simultaneous or reciprocal iterative inference of perceptual inference and AIF~\cite{friston2017active}.
In the perceptual inference formula proposed by Friston et al., variational free energy is defined, and an approximate posterior distribution is obtained by variational inference.
In this section, we assume that $Z$ is the set of the hidden states (latent variables), $X$ is the set of observations, and $q(Z)$ is the variational approximate distribution.
In the FEP, the general equations (Eqs.~(\ref{eq:FEP_VFE}) and (\ref{eq:FEP_EFE})) for the variational and expected free energies have been proposed~\cite{friston2017active,fep}. 
Here, the probability distributions, $p(Z \mid X)$ and $q(Z)$, are arbitrary.
\\
\textbf{The variational free energy} for perceptual inference and learning is described as follows: 
\begin{align}
F(q, p; X)
&=   D_{\rm{KL}}\left[ q(Z) \| p(Z \mid X) \right]  -  \log p(X).
\label{eq:FEP_VFE}
\end{align}
where $D_{\rm{KL}}\left[ q \| p \right] $ is the Kullback--Leibler (KL) divergence between the distributions $q$ and $p$.
It is inferred that moves $q(Z)$ closer to the posterior distribution $p(Z \mid X)$.
\\
\textbf{The expected free energy} for AIF is stated as follows:
\begin{align}
G(\pi, \tau)
&=  - \underbrace{\mathbb{E}_{q(X_{\tau} \mid \pi)}\left[  D_{\rm{KL}}\left[ q(Z_{\tau} \mid X_{\tau}, \pi) \| q(Z_{\tau} \mid \pi) \right] \right] }_{\text{information gain}} - \underbrace{ E_{q(X_{\tau} \mid \pi)} \left[ \log p(X_{\tau})  \right] }_{\text{expected log-evidence}}
\label{eq:FEP_EFE}
\end{align}
where policy $\pi$ indexes a sequence of control states from current time $t$ to the future. The future time was $\tau > t$. 
The notation $E_{q(X)}[f(X)]$ represents the expected value of a function $f(X)$ with respect to the distribution $q(X)$.

IG (also called mutual information~\cite{Placed2022}) is responsible for the epistemic value term in the expected free energy of the AIF.
The epistemic value is also called the value of information or intrinsic value.
Epistemic value can be interpreted as the resolution of uncertainty by motivating curiosity and novelty-seeking actions.
The expected log evidence is responsible for pragmatic value (extrinsic value).

In this study, the robot explored and learned spatial concepts autonomously through repeated perceptual-based online learning and active exploration.
The perceptual inference and learning of the model parameters correspond to online learning using particle filters.
The variational inferences from Eq.~(\ref{eq:FEP_VFE}) and the particle filter attempt to obtain the same posterior distribution.
This posterior distribution was approximated using multiple samples (i.e., particles).
IG maximization corresponds to policy selection in AIF.
If we assume the pragmatic value, whose prior distribution of the observed data is uniform, the IG maximization becomes equivalent to the expected free energy minimization.
Instead of inferring the distribution of policy $\pi$ and then determining action $a$ based on it as in Friston et al.'s formulation, we determine the action directly from IG maximization.

\section{Proposed method: Active Exploration for Spatial Concept Formation} 
\label{sec:proposed}

\begin{figure}[!tb]
    \centering
	\includegraphics[width=1.00\linewidth]{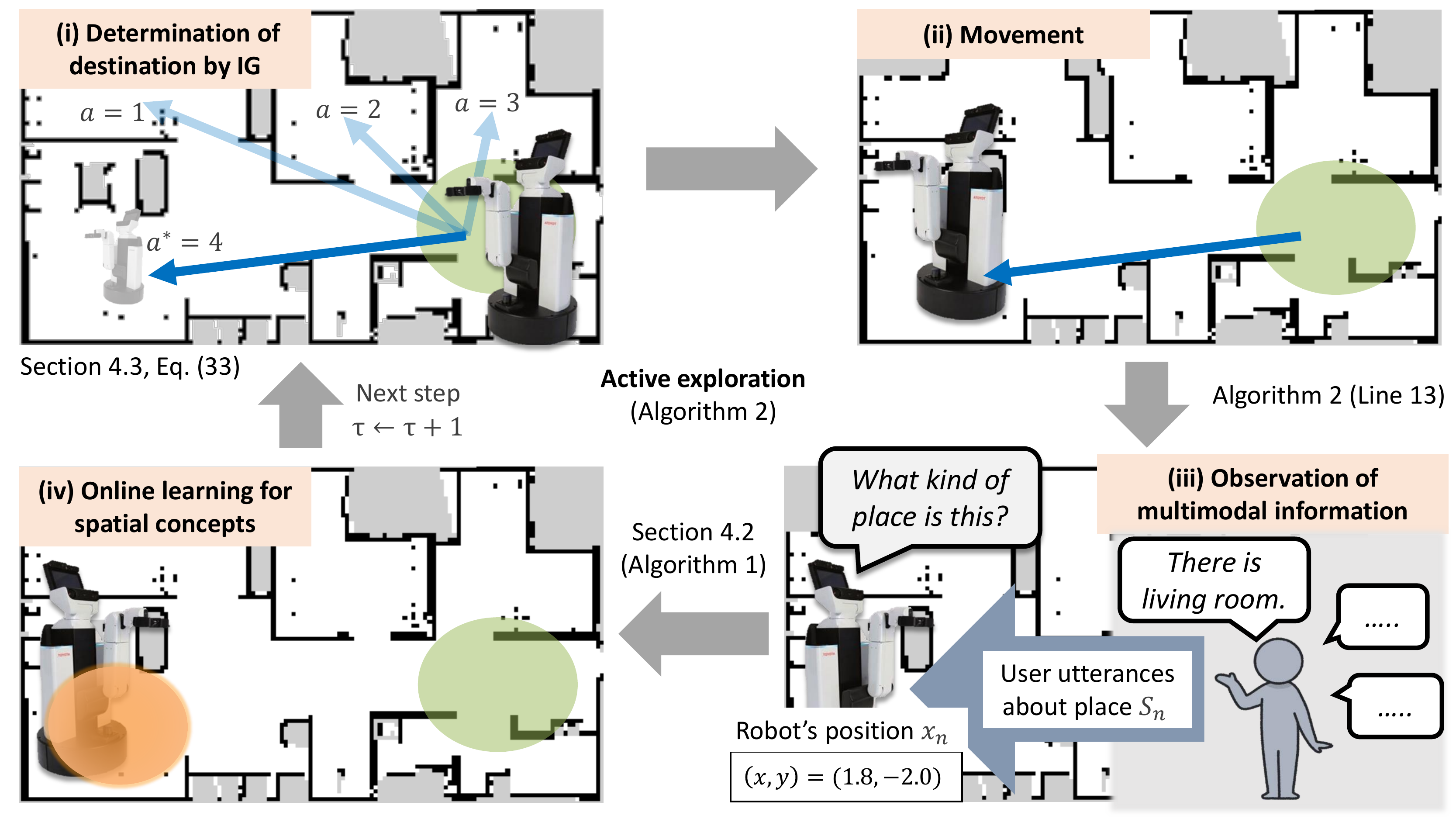}
	\caption{
    	Overview of flow of the proposed method (See Algorithm~\ref{alg:ActiveSpCoA} for details.).
     (i) The robot determines next destination $a^{\ast}$ using the utility function based on IG (See Section~\ref{sec:proposed:active_exploration}; especially Eq.~(\ref{eq:argmax_IG_cost})).
     (ii) The robot moves to the position $x_{a^{\ast}}$ of the destination (See Algorithm~\ref{alg:ActiveSpCoA}, line 13).
     (iii) The robot asks the user, `What kind of place is this?' to observe the position $x_n$ and words $S_n$ (Here, $n=a^{\ast}$) as multimodal observation of the PGM in Figure~\ref{fig:ActiveSpCoA}.
     (iv) The robot learns spatial concepts based on the observations (See Section~\ref{sec:proposed:online_learning} and Algorithm~\ref{alg:onlineSpCoA}).
     The above process is repeated until potential destinations are over or until learning is completed by sufficient exploration.
     In this scenario, it is assumed that a map has been pre-generated using SLAM, allowing for accurate self-localization and navigation to the destination.
	}
	\label{fig:method_flow}
\end{figure} 

In this study, we propose SpCoAE, a method for a robot to select the next destination in the online learning of spatial concepts.
Figure~\ref{fig:method_flow} illustrates the flow of the proposed method.
SpCoAE is an AIF-inspired method that combines online learning with particle filters (described in Section~\ref{sec:proposed:online_learning}) and active exploration based on IG (formulated in Section~\ref{sec:proposed:active_exploration}) in a PGM for spatial concept formation (defined in Section~\ref{sec:proposed:generative_model}).

Section~\ref{sec:proposed:generative_model} presents model definitions.
In this section, we describe the definition of the generative process of random variables in the proposed PGM for spatial concept formation.
Based on this PGM, online learning in Section~\ref{sec:proposed:online_learning} and active exploration in Section~\ref{sec:proposed:active_exploration} are executed.
Section~\ref{sec:proposed:online_learning} discusses the learning of spatial concepts.
In this section, we present the mathematical derivation of the particle filter in the proposed PGM.
Section~\ref{sec:proposed:active_exploration} describes the destination determination through active exploration.
In this section, we describe the formulation of IG-based active exploration and the procedure for deriving the algorithm.

\subsection{Probabilistic generative model}
\label{sec:proposed:generative_model}

\begin{figure}[tb]
    \centering
	\includegraphics[width=1.00\linewidth]{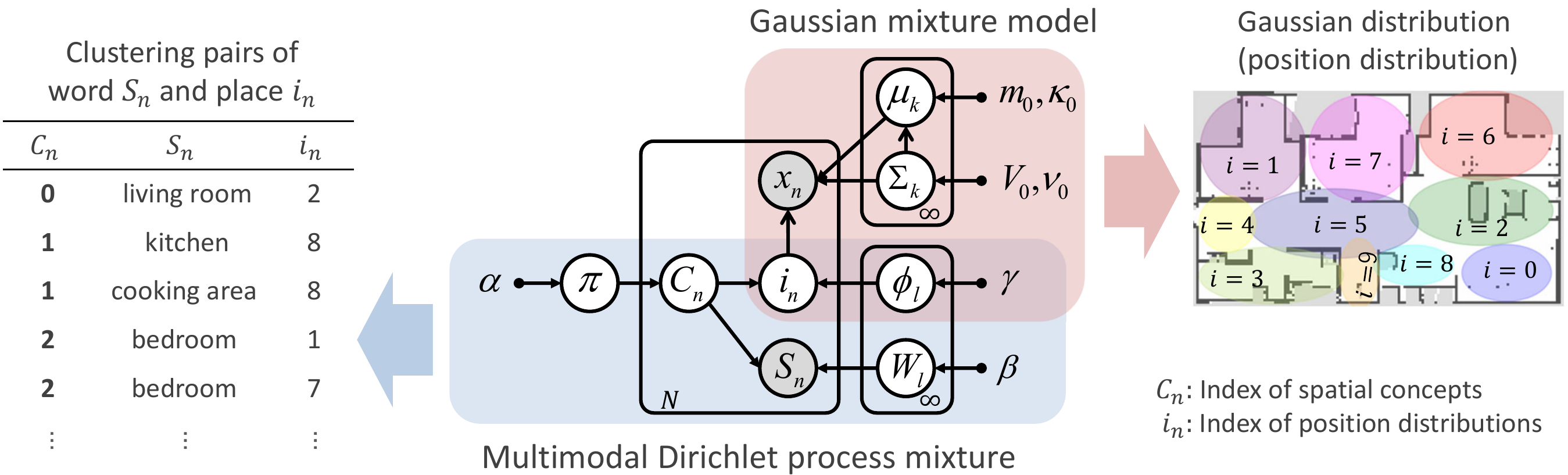}
	\caption{
    	Graphical model representation for spatial concept formation.
        The graphical model represents conditional dependency between random variables.
        Gray and white nodes represent observations and unobserved latent variables, respectively.
        This model is integrated by a multimodal Dirichlet process mixture whose emission distributions are multinomial distributions based on a Gaussian mixture model.
        Table~\ref{tab:element_of_graphical_model} summarizes the descriptions of the variables in the model.
	}
	\label{fig:ActiveSpCoA}
\end{figure}

\begin{table}[tb]
    \tbl{
        Description of the variables in the generative model.
    }{
		\begin{tabular}{cl}
			\hline\noalign{\smallskip}
			\textbf{Symbol} & \textbf{Definition} \\
			\noalign{\smallskip}\hline\noalign{\smallskip}
			$x_{n}$  & Position of robot {($(x,y)$-coordinates of floor plane)}\\
			$S_{n}$  & Words representing place corresponding to position {$x_{n}$} (Bag-of-Words)\\ 
			$C_{n}$  & Latent variable for index of spatial concepts \\ 
			$i_{n}$  & Latent variable for index of position distributions  \\ 
			$\pi$    & Parameters of multinomial distribution for index $C_{n}$ of spatial concepts \\ 
			$\phi_l$ & Parameters of multinomial distribution for index $i_{n}$ of position distribution \\ 
			$W_l$    & Parameters of multinomial distribution for observing $S_{n}$ \\ 
			$\mu_k,\Sigma_k$         & Parameters of Gaussian distribution (position distribution) for observation of $x_{n}$ \\ 
			$\alpha,\beta,\gamma,$   & Hyperparameters of Dirichlet prior distribution \\
			$m_0,\kappa_0,V_0,\nu_0$ &  Hyperparameters of Gaussian and inverse Wishart prior distributions \\ 
			\noalign{\smallskip}\hline\noalign{\smallskip}
			{$n$} & {Index number of training data ({$n \in \{ 1,2, \dots, N\}$})} \\
			{$l$} & {Index number of spatial concepts  ({$l \in \{ 1,2, \dots, L\}$})} \\
			{$k$} & {Index number of position distributions  ({$k \in \{ 1,2, \dots, K\}$})} \\
			{$N$} & {Total number of training data} \\
			{$L$} & {Upper limit of number of spatial concepts} \\
			{$K$} & {Upper limit of number of position distributions} \\
			\noalign{\smallskip}\hline
		\end{tabular}
    }
    \label{tab:element_of_graphical_model}
\end{table}

The capabilities of the proposed PGM are as follows.
(i) Categorization is performed using unsupervised learning based on multimodal observations. 
(ii) There is many-to-many correspondence between the words and places.
Moreover, the model does not require a prior manual setting of the vocabulary labels or categories.

Figure~\ref{fig:ActiveSpCoA} shows a graphical representation of spatial concept formation, and Table~\ref{tab:element_of_graphical_model} summarizes the descriptions of the variables in the model.
The generative process of the model is as follows.
\begin{align}
\pi        &\sim \operatorname{DP}(\alpha) \label{eq:pi}\\
\phi_{l}   &\sim \operatorname{DP}(\gamma) & l &= 1, 2, \dots, \infty \\
W_{l}      &\sim \operatorname{Dir}(\beta) \\ 
\Sigma_{k} &\sim \mathcal{IW}(V_0, \nu_0) & k &= 1, 2, \dots, \infty \\
\mu_{k}    &\sim \mathcal{N}( m_0,  \Sigma_{k} / \kappa_0 ) \\ 
C_{n}      &\sim \operatorname{Cat}(\pi) & n &= 1, 2, \dots, N \\
i_{n}      &\sim \operatorname{Cat}(\phi_{C_{n}}) \\ 
S_{n}      &\sim \operatorname{Mult}(W_{C_{n}}) \\ 
x_{n}      &\sim \mathcal{N}( \mu_{i_{n}}, \Sigma_{i_{n}}) \label{eq:x} 
\end{align}
where $\operatorname{DP}()$ denotes the prior distribution in the Dirichlet process. 
$\operatorname{Dir}()$ is a Dirichlet, $\operatorname{Cat}()$ is a categorical, $\operatorname{Mult}()$ is multinomial, $\mathcal{IW}()$ is an inverse-Wishart, and $\mathcal{N}()$ is a Gaussian distribution.
We refer to the literature on machine learning~\cite{murphy2012machine} for specific formulas of the above probability distributions.
In this study, the Dirichlet process is represented as a \textit{stick-breaking process (SBP)}~\cite{sethuraman1994constructive,ref:ishwaran2001gibbs}.
We adopted a weak-limit approximation~\cite{fox2011sticky} for SBP.
Appropriate numbers of spatial concepts and position distributions are probabilistically determined by learning based on observations.

\subsection{Online learning algorithm}
\label{sec:proposed:online_learning}

In this study, the robot learns spatial concepts from observations to select the next move.
Therefore, we introduced online learning using a \textit{Rao-Blackwellized particle filter (RBPF)}~\cite{doucet2000rao} as in SpCoSLAM.
This method estimates the posterior distributions of the parameters of a spatial concept using words and positions as multimodal observations of a place.
The learning algorithm is presented as Algorithm~\ref{alg:onlineSpCoA}.

The joint posterior distributions of all the parameters to be estimated when learning spatial concepts and their factorization are as follows:
\begin{align}
&p(\Theta, C_{1:n}, i_{1:n} \mid x_{1:n}, S_{1:n}, h) \nonumber \\
&= p(\Theta \mid C_{1:n}, i_{1:n}, x_{1:n}, S_{1:n}, h) 
p(C_{1:n}, i_{1:n} \mid x_{1:n}, S_{1:n}, h)
\label{eq:learning_algorithm}
\end{align}
where the parameter set of each spatial concept is $\Theta = \{ \{ \mu_k \}, \{ \Sigma_k \}, \{ \phi_l \}, \{ W_l \}, \pi\}$ and the set of hyperparameters is $h = \{ \alpha, \beta, \gamma, m_0, \kappa_0, V_0, \nu_0\}$.
Appendix~\ref{subsec:Theta} explains the calculation procedure for the posterior distribution of the model parameter $\Theta$.

\begin{algorithm}[tb]
    \caption{Online learning algorithm for spatial concepts.} 
    \label{alg:onlineSpCoA}          
    \begin{algorithmic}[1]
        \State{$Z_{n-1} = \{ {C}_{1:n-1}^{[r]}, i_{1:n-1}^{[r]}, \Theta_{n-1}^{[r]} \}_{r=1}^{R}$, $X_{1:n} = \{ x_{1:n},S_{1:n}, h \}$}
        \Procedure{${\textsc{Online\_Learning}}$}{$Z_{n-1},X_{1:n}$}
            \State{$\bar{Z}_{n} = Z_{n} =  \emptyset$}
            \Comment{Initialize set of particles}
            \For{$r=1$ to $R$}
                \State{$C_{n}^{[r]},i_{n}^{[r]} \sim p(C_{n},i_{n} \mid  C_{1:n-1}^{[r]},i_{1:n-1}^{[r]},x_{1:n},S_{1:n},{h})$}
                \Comment{Sampling}
                \State{$\omega_{n}^{[r]} = {p(x_{n}, S_{n} \mid C_{1:n-1}^{[r]}, i_{1:n-1}^{[r]}, x_{1:n-1}, S_{1:n-1}, h)}$}
                \Comment{Importance}
                \State{$\Theta_{n}^{[r]} = {E}[p(\Theta \mid {C}_{1:n}^{[r]}, i_{1:n}^{[r]}, x_{1:n}, S_{1:n}, {h})]$}
                \Comment{Calculation of posterior parameters}
                \State{$\bar{Z}_{n} = \bar{Z}_{n} \cup \langle {C}_{1:n}^{[r]}, i_{1:n}^{[r]}, \Theta_{n}^{[r]}, \omega_{n}^{[r]} \rangle$}
            \EndFor
            \For{$r=1$ to $R$}
                \State{draw $j$ with probability $\propto \{ \omega_{n}^{[j]} \}$}
                \Comment{Resampling}
                \State{add $\langle {C}_{1:n}^{[j]}, i_{1:n}^{[j]}, \Theta_{n}^{[j]} \rangle$ to $Z_{n}$}
            \EndFor
            \State{\textbf{return} $Z_{n}$}
        \EndProcedure
    \end{algorithmic}
\end{algorithm}

\subsubsection{Derivation and process in Rao-Blackwellized particle filter}
\label{subsec:online_algorithm_derivation}

The second term, $p(C_{1:n}, i_{1:n} \mid x_{1:n}, S_{1:n}, h)$, in Eq.~(\ref{eq:learning_algorithm}) was calculated using the RBPF. 
The particle filter algorithm is based on sampling importance resampling.
The process can be summarized by the following steps: 

\textbf{Sampling}: 
The latent variables $C_{n}, i_{n}$ are sampled simultaneously as the proposal distribution, $q_{n}$, as follows\footnote{The intermediate equation is provided in Appendix~\ref{apdx:online:q_n}.}$^{,}$\footnote{Details of the simultaneous sampling of $C_{n}$ and $i_{n}$ are presented in Appendix~\ref{subsec:C_n_tau,i_n_tau}.}:
\begin{align}
C_{n}, i_{n} &\sim  p(C_{n}, i_{n} \mid C_{1:n-1}, i_{1:n-1}, x_{1:n}, S_{1:n}, h)  \\ 
&\propto  p(x_{n} \mid x_{1:n-1},i_{1:n}, h)p(S_{n} \mid S_{1:n-1}, C_{1:n}, \beta) 
p(C_{n}, i_{n} \mid C_{1:n-1}, i_{1:n-1}, \alpha, \gamma).
\label{eq:qn}
\end{align}

\textbf{Importance Weighting}: 
Weight $\omega_{n}$ is expressed as follows:
\begin{align}
\omega_{n}^{[r]} &= \frac{p(C_{1:n}^{[r]}, i_{1:n}^{[r]} \mid x_{1:n}, S_{1:n}, h)}{q(C_{1:n}^{[r]}, i_{1:n}^{[r]} \mid x_{1:n}, S_{1:n}, h)} 
=\frac{P_{n}^{[r]}}{Q_{n}^{[r]}}\label{eq:omega}
\end{align}
where $r$ is the particle number, and $R$ is the number of particles.
The subsequent equations were computed for each particle $[r]$; however, the subscripts indicating the particle number were omitted.

Target distribution $P_n$ can be transformed as follows\footnote{The intermediate formulas are given in Appendix \ref{apdx:online:P_n}.}:
\begin{align}
{P_{n}} &\propto p(C_{n}, i_{n} \mid C_{1:n-1}, i_{1:n-1}, x_{1:n}, S_{1:n}, h) \nonumber \\
& \qquad p(x_{n} \mid C_{1:n-1}, i_{1:n-1}, x_{1:n-1}, h) \nonumber \\
& \qquad p(S_{n} \mid S_{1:n-1}, C_{1:n-1}, h) P_{n-1}. 
\label{eq:Pt}
\end{align}

The proposal distribution $Q_{n}$ can be transformed as follows:
\begin{align}
Q_{n} &= \underbrace{p(C_{n}, i_{n} \mid C_{1:n-1}, i_{1:n-1}, x_{1:n}, S_{1:n}, h)}_{q_{n}} \nonumber \\
&\qquad \underbrace{q(C_{1:n-1}, i_{1:n-1} \mid x_{1:n-1}, S_{1:n-1}, h)}_{Q_{n-1}} \\
&= q_{n}Q_{n-1} .
\label{eq:Qt}
\end{align}
Here, the proposal distribution $q_{n}$ is the marginal distribution of the parameter set $\Theta$ of latent variables $C_{n}, i_{n}$.

From Eqs.~(\ref{eq:omega})--(\ref{eq:Qt}), weight $\omega_{n}$ is expressed as follows:
\begin{align}
\omega_{n} &= 
p(x_{n} \mid C_{1:n-1}, i_{1:n-1}, x_{1:n-1}, h) p(S_{n} \mid S_{1:n-1}, C_{1:n-1}, h) 
\underbrace{\frac{P_{n-1}}{Q_{n-1}}}_{\omega_{n-1}}.
\label{eq:weight}
\end{align}

The new weight terms for $n$ in Eq.~(\ref{eq:weight}) are transformed and marginalized for $C_{n}$ and $i_{n}$ as shown in Eq.~(\ref{eq:weight_sum_ci}).
The terms in the sum operation are the same as those in Eq.~(\ref{eq:qn}). 
These values were proportional to the proposal distribution $q_{n}$.
These have already been computed when $C_{n}, i_{n}$ is sampled from $q_{n}$.
Therefore, all probabilities were added before normalization. 
\begin{align}
&p(x_{n} \mid C_{1:n-1}, i_{1:n-1}, x_{1:n-1}, h)p(S_{n} \mid S_{1:n-1}, C_{1:n-1}, h) \nonumber \\
&\propto p(x_{n}, S_{n} \mid C_{1:n-1}, i_{1:n-1}, x_{1:n-1}, S_{1:n-1}, h) \\
&= \sum_{C_{n}}\sum_{i_{n}}  p(x_{n} \mid x_{1:n-1}, i_{1:n}, h)p(S_{n} \mid S_{1:n-1},C_{1:n},  \beta) 
p(C_{n}, i_{n} \mid C_{1:n-1}, i_{1:n-1}, \alpha, \gamma).  
\label{eq:weight_sum_ci}
\end{align}

\textbf{Resampling}: 
As shown in lines 10--13 of Algorithm~\ref{alg:onlineSpCoA}, the particles are sampled again as $R$ particles, according to their weights $\omega_{n}$.
After resampling, all the particles have the same weights.

\subsection{Active exploration algorithm}
\label{sec:proposed:active_exploration}

The proposed method uses IG, an information-theoretic measure, to select and move towards a position in an environment that is most likely to reduce uncertainty. Moreover, it obtains observations, including user utterances.
The algorithm for the proposed method is presented as Algorithm~\ref{alg:ActiveSpCoA}.
The details of the derivation of the algorithm are provided in Sections~\ref{sec:proposed:selection} and \ref{sec:proposed:ig}.
We also describe a method for including costs related to travel distance, as presented in Section~\ref{sec:proposed:travelcost}.

\begin{algorithm}[tb]
    \caption{
        SpCoAE: Active exploration algorithm for spatial concept formation.
    } 
    \label{alg:ActiveSpCoA}          
    \begin{algorithmic}[1]
            \State{Initialize $n_{0}=\emptyset$, $ Z_{0}=\emptyset$, $X_{n_{0}}=\emptyset$} 
            \For{$\tau=1$ to $\mathcal{T}$} \Comment{Number of action trials} 
                \For{\textbf{all} $ \{ x_a \in \mathbb{R}^{D} \mid \text{free-space in a map} \}$} \Comment{Number of candidate positions} 
                    \For{$r=1$ to $R$} \Comment{Number of particles}
                        \For{$j=1$ to $J$} \Comment{Number of pseudo observations} 
                            \State{$X_{a}^{[r,j]} \sim p(X_{a} \mid Z_{\tau-1}^{[r]}, X_{n_0})$}
                        \EndFor
                    \EndFor
                    \State{$\displaystyle {\mbox{IG}}_{a} = \sum_{r=1}^{R} \sum_{j=1}^{J} \log \frac{p(X_{a}^{[r,j]} \mid Z_{\tau-1}^{[r]}, X_{n_0})}{\sum_{r^{\prime}=1}^{R} p(X_a^{[r,j]} \mid Z_{\tau-1}^{[r^{\prime}]}, X_{n_0}) }$} 
                \EndFor                    
                \State{$a^{\ast} = \argmax_{a} ( {\mbox{IG}}_{a} - \eta \, \mathrm{TravelCost}(a) )$} 
                \Comment{Select position $x_{a^{\ast}}$}
                \State{$n_{0} = n_{0} \cup \{ a^{\ast} \}$ }
                \State{Move to position $x_{a^{\ast}}$, and observe words $S_{a^{\ast}}$}
                \Comment{Observe $X_{n_{0}}$}
                \State{$Z_{\tau} = \textsc{Online\_Learning}(Z_{\tau-1},X_{n_0})$} 
            \EndFor
    \end{algorithmic}
\end{algorithm}

\subsubsection{Formulation of how to select destination}
\label{sec:proposed:selection}

This section discusses, in the first half, the conceptual and basic formulations that underlie active exploration and, in the second half, the practical formulation and derivation for executing the basic formulation with online learning.

\textbf{Basic formulation of active exploration:}
The proposed method uses a formulation that minimizes the KL divergence between the two types of posterior distributions\footnote{This formulation is similar to that in MHDP-based active perception/learning methods~\cite{Taniguchi2015yoshino,Yoshino2021}.}.
The first is the final posterior distribution when data are observed $N$ times, $p(\Theta, C_{1:N}, i_{1:N} \mid x_{1:N}, S_{1:N},h)$.
The second is the current posterior distribution when the data are observed as the next destination from the current step, $p(\Theta, C_{1:N}, i_{1:N} \mid x_{n_0\cup a}, S_{n_0\cup a},h)$.
This means determining the next observation so that the distribution at the current observations makes similar to the final posterior distribution after all observations.
This formulation aims to select data $a$.
The index set $\{ 1:N \}$ of the data points corresponding to multiple possible destinations for asking the user a question is defined as the action set for $a$.
In each step, as observed data is obtained, it is minimized as follows:
\begin{align}
    \minimize_{a} D_{\rm{KL}} [ p(\Theta,C_{1:N},i_{1:N} \mid x_{1:N},S_{1:N},h) 
    \| p(\Theta,C_{1:N},i_{1:N} \mid x_{n_0\cup a},S_{n_0\cup a},h) ].
    \label{eq:minimize_KL}
\end{align}
Note that $n_0$ is the set of already observed data indices; that is, the subset of indices of the plate representation from $1$ to $N$ in the graphical model for spatial concept formation in Figure~\ref{fig:ActiveSpCoA}.
In the first step, the case where data is unobserved is $n_0=\emptyset$.
However, Eq.~(\ref{eq:minimize_KL}) cannot be computed in online learning. This problem must be solved in some manner.
Note that neither true $\{ x_{1:N}, S_{1:N} \}$ nor $\{ x_{a}, S_{a} \}$ can be observed before moving to the next destination.

\begin{figure}[tb]
    \centering
	\includegraphics[width=0.92\linewidth]{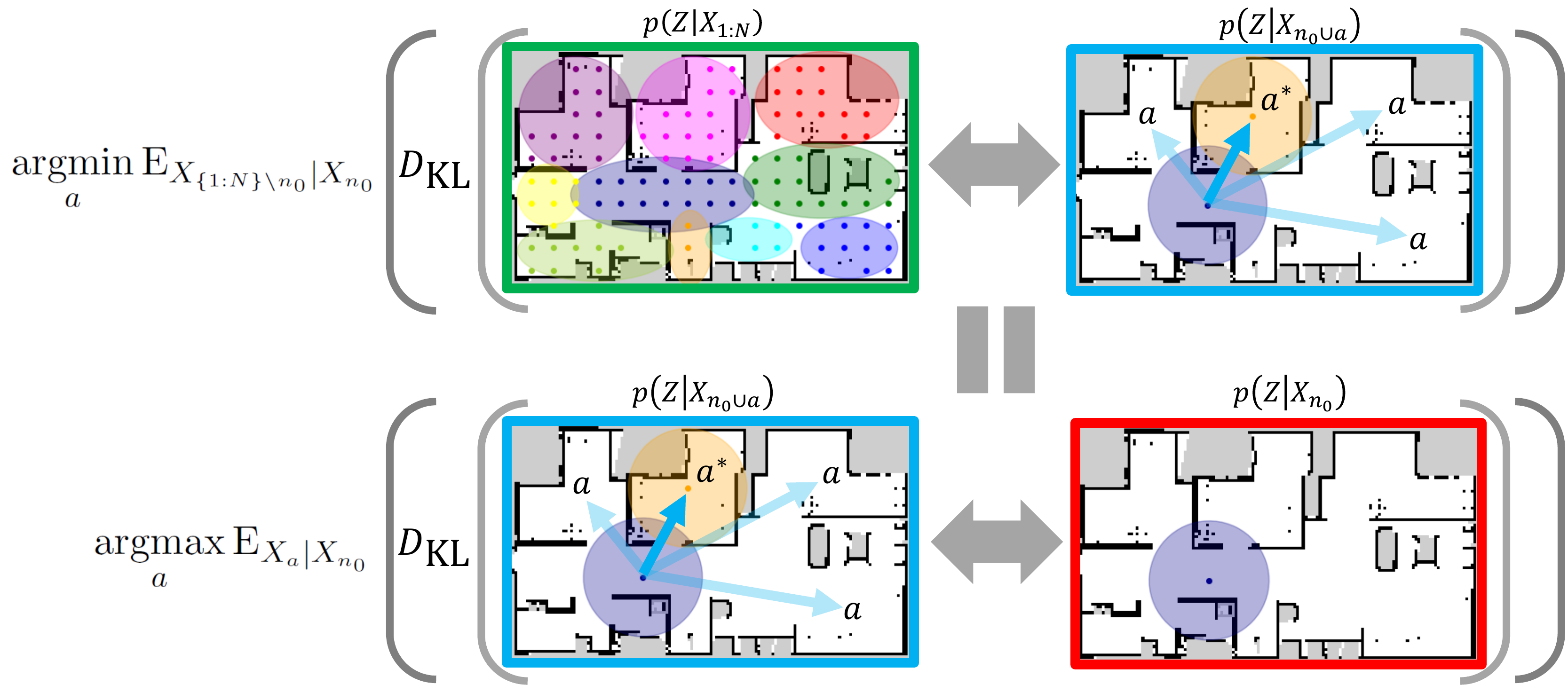}
	\caption{
    	Relationship diagram between Eqs.~(\ref{eq:KL_min}) and (\ref{eq:KL_max}).
        Each ellipse is drawn as a spatial concept.
        Top (Eq.~(\ref{eq:KL_min})) represents an action decision that moves the distribution when an action is taken (upper right) closer to the final posterior distribution (upper left).
        Bottom (Eq.~(\ref{eq:KL_max})) represents an action decision that results in a distribution (bottom right) that is the furthest away from the current distribution (bottom right).
        This implies that these are equivalent.
	}
	\label{fig:KLtoIG}
\end{figure} 

\textbf{Formulation for computational feasibility:}
As a suitable alternative, we calculated the expected value of the KL divergence using Eqs.~(\ref{eq:KL_min})--(\ref{eq:argmax_IG})\footnote{
The transformation details of Eqs. (\ref{eq:KL_min})--(\ref{eq:argmax_IG}); please refer to Appendix~\ref{apdx:active:IG_formula}.
This derivation is based on MHDP-based active perception/learning methods~\cite{Taniguchi2015yoshino,Yoshino2021}.}
to obtain the index of the candidate point $a^{\ast}$ of the data to be observed at the next destination as follows:
\begin{align}
    a^{\ast} &= \argmin_{a} \mathbb{E}_{X_{\{1:N\} \backslash n_0}  \mid X_{n_0}} 
    \left[ D_{\rm{KL}}\left[ p(Z \mid X_{1:N}) \| p(Z \mid X_{n_0\cup a}) \right] \right] \label{eq:KL_min} \\
    &= \argmax_{a} \mathbb{E}_{X_a \mid X_{n_0}}\left[  D_{\rm{KL}}\left[ p(Z \mid X_{{n_0}\cup a}) \| p(Z \mid X_{n_0}) \right] \right] \label{eq:KL_max} \\
    &= \argmax_{a} \mbox{IG}(Z;X_a \mid X_{n_0}),
    \label{eq:argmax_IG}
\end{align}
where
$a \in \{ 1:N \} \backslash n_0$ is the index of the data point to be observed at the next destination, excluding the previously observed data $n_0$.
Let $Z=\{C_{1:N},i_{1:N},\Theta\}$, $X_{n_0}=\{x_{n_0},S_{n_0},h\}$.
Figure~\ref{fig:KLtoIG} shows the relationship between Eqs. (\ref{eq:KL_min}) and (\ref{eq:KL_max}), respectively.
Eq.~(\ref{eq:KL_max}) expresses the maximization of the expected value of the KL divergence between the posterior distributions after observing the data at the next destination and the current step.
Based on the information theory, the expected value of the KL divergence in Eq.~(\ref{eq:KL_max}) is defined as IG in Eq.~(\ref{eq:argmax_IG}), i.e., $ \mbox{IG}(Z;X_a \mid X_{n_0}) \coloneqq \mathbb{E}_{X_a \mid X_{n_0}}\left[  D_{\rm{KL}}\left[ p(Z \mid X_{{n_0}\cup a}) \| p(Z \mid X_{n_0}) \right] \right]$.
IG is known as mutual information~\cite{Placed2022}.

\subsubsection{Derivation of approximate computation of IG with particle filter}
\label{sec:proposed:ig}

In this section, we describe a computationally efficient way to approximate IGs that takes advantage of particle filter results in online learning.
The formula for IG in Eq.~(\ref{eq:argmax_IG}) as follows:
\begin{align}
    &\mbox{IG}(Z;X_a \mid X_{n_0}) \label{eq:IG_yoshino} \nonumber \\
    &=\sum_Z \sum_{X_{a}} \left[ p(Z, X_{a} \mid X_{n_0}) \log\frac{p(Z, X_{a} \mid X_{n_0})}{p(Z \mid X_{n_0})p(X_{a} \mid X_{n_0})} \right] \\
    &\approx \sum_{r=1}^{R} \sum_{j=1}^{J} \left[ \omega_{n_0}^{[r]} \log\frac{p(X_{a}^{[j]} \mid Z^{[r]}, X_{n_0})}{\sum_{r^{\prime}=1}^{R} \left[p(X_a^{[j]} \mid Z^{[r^{\prime}]}, X_{n_0}) \omega_{n_0}^{[r^{\prime}]} \right]} \right],  \nonumber \\
    &\qquad \qquad Z^{[r]} \sim q(Z \mid X_{n_0}), \quad X_{a}^{[j]} \sim p(X_{a} \mid Z^{[r]}, X_{n_0}) 
    \label{eq:draw_Z_X_a} 
\end{align}
where $R$ is the number of particles, $J$ is the number of pseudo observations, and $\omega_{n_0}^{[r]}$ is the particle weight.
The equation above is approximated by sampling based on the predictive distribution $p(X_a \mid X_{n_0})$. 
There are two approximation levels.

First, $p(Z \mid X_{n_0})$ is approximated by the estimated samples in a particle filter based on the observed values as follows:
\begin{align}
p(X_a \mid X_{n_0})&=\sum_{Z} [p(X_a \mid Z, X_{n_0})p(Z \mid X_{n_0})] \\
&\approx\sum_{r=1}^{R} \left[p(X_a \mid Z^{[r]}, X_{n_0}) \omega_{n_0}^{[r]} \right],\quad Z^{[r]} \sim q(Z \mid X_{n_0}). 
\label{eq:pred_dist_X}
\end{align}

Next, $p(X_a \mid Z^{[r]}, X_{n_0})$ in Eq.~(\ref{eq:pred_dist_X}) can be approximated by sampling $J$ pseudo observations, where $X_a^{[j]}$ denotes the pseudo observations obtained at the next destination.
This distribution is expressed as follows:
\begin{align}
    p(X_a \mid Z, X_{n_0}) 
    &= p(x_{a}, S_{a} \mid C_{n_0}, i_{n_0}, x_{n_0}, S_{n_0}, h) .
    \label{eq:p_Xa}
    \end{align}
    When sampling the pseudo-observation $X_a^{[j]}$, assuming that $x_a$ is fixed for some $a$, only $S_a$ must be sampled.
    In this case, Eq.~(\ref{eq:p_Xa}) is identical to Eq.~(\ref{eq:weight_sum_ci}) for weight calculation in the particle filters.
    \begin{align}
    X_a^{[j]} = S_a \mid x_{a}
    &\sim p(x_{a}, S_{a} \mid C_{n_0}, i_{n_0}, x_{n_0}, S_{n_0}, h) \\
    &\quad = \sum_{C_{a},i_{a}} p(S_{a} \mid C_{n_0\cup a}, S_{n_0}, h) p(x_{a} \mid i_{n_0\cup a}, x_{n_0}, h) p(C_{a},i_{a} \mid C_{n_0}, i_{n_0}, h) .
\end{align}

In conventional similarity methods~\cite{Taniguchi2015yoshino,Yoshino2021}, the Monte Carlo approximation is performed again.
However, the proposed method can use the estimated results of existing particle filters, as expressed in Eq.~(\ref{eq:draw_Z_X_a}).
As $Z^{[r]}, \omega_{n_0}^{[r]}$ has already been estimated by the online learning algorithm, we can sample $X_{a}^{[j]}$ from $p(X_{a} \mid Z^{[r]}, X_{n_0})$.
Finally, we obtain probability $p(X_a^{[j]} \mid Z^{[r]}, X_{n_0})$ based on the sampled values.
Weight $\omega_{n_0}^{[r]}$ can be treated as a constant $1/R$ because it is resampled\footnote{Note that the weights need to be used if they are not resampled by devices such as effective sample size~\cite{Kong1994,Liu1996}.}.
Thus, computational reuse based on particle filters in learning can increase computational efficiency.

\subsubsection{Introduction of travel distance costs}
\label{sec:proposed:travelcost}

When selecting a movable/reachable destination and controlling the robot at that destination, IG (Eq.~(\ref{eq:draw_Z_X_a})). However, the cost of movement also needs to be considered.
In this study, the criterion of `\textit{utility}' that includes the travel cost of distance, which is used in active SLAM~\cite{Stachniss2005activeslam}, is also introduced into the IG of SpCoAE.
Therefore, the utility function for destination $a$ is defined as 
\begin{align}
    a^{\ast} &= \argmax_{a} \left( \mbox{IG}(Z;X_a \mid X_{n_0}) - \eta \, \mathrm{TravelCost}(a) \right)
    \label{eq:argmax_IG_cost}
\end{align}
where $\eta $ is a parameter of a weighting factor that trades off the cost with the IG and $\mathrm{TravelCost}(a)$ is the travel cost, which is the path length estimated by the A$^{\star}$ algorithm from the current position to the target position, $x_{a}$, on the map.

Thus, the IG and travel costs can be treated as a tradeoff.
Consequently, the destination ($a$) that maximizes the utility, represented by Eq.~(\ref{eq:argmax_IG_cost}) can be determined as the optimal destination $x_{a^{\ast}}$, and the robot can move to that point.

The introduction of travel costs has similar implications for the incorporation of state transitions, that is, movements and time steps.
The graphical model of SpCoAE does not include SLAM.
However, the consideration of state transitions is inherently desirable because the robot moves to different locations on the map.
Theoretically, we expect the travel cost to act as an approximation of the state transition probabilities in a graphical model such as SLAM.
We believe that a graphical model of SpCoAE can be integrated into SLAM in the future.

\section{Experiment I: Simulator environments}
\label{sec:exp1}

We evaluated whether active exploration using the proposed method enables efficient spatial concept formation.

\subsection{Condition}
\label{sec:exp1:condition}

\begin{figure}[tb]
	\centering
   	\subfloat[Simulated home environment]{
       	\resizebox*{5cm}{!}{\includegraphics[width=0.3\linewidth]{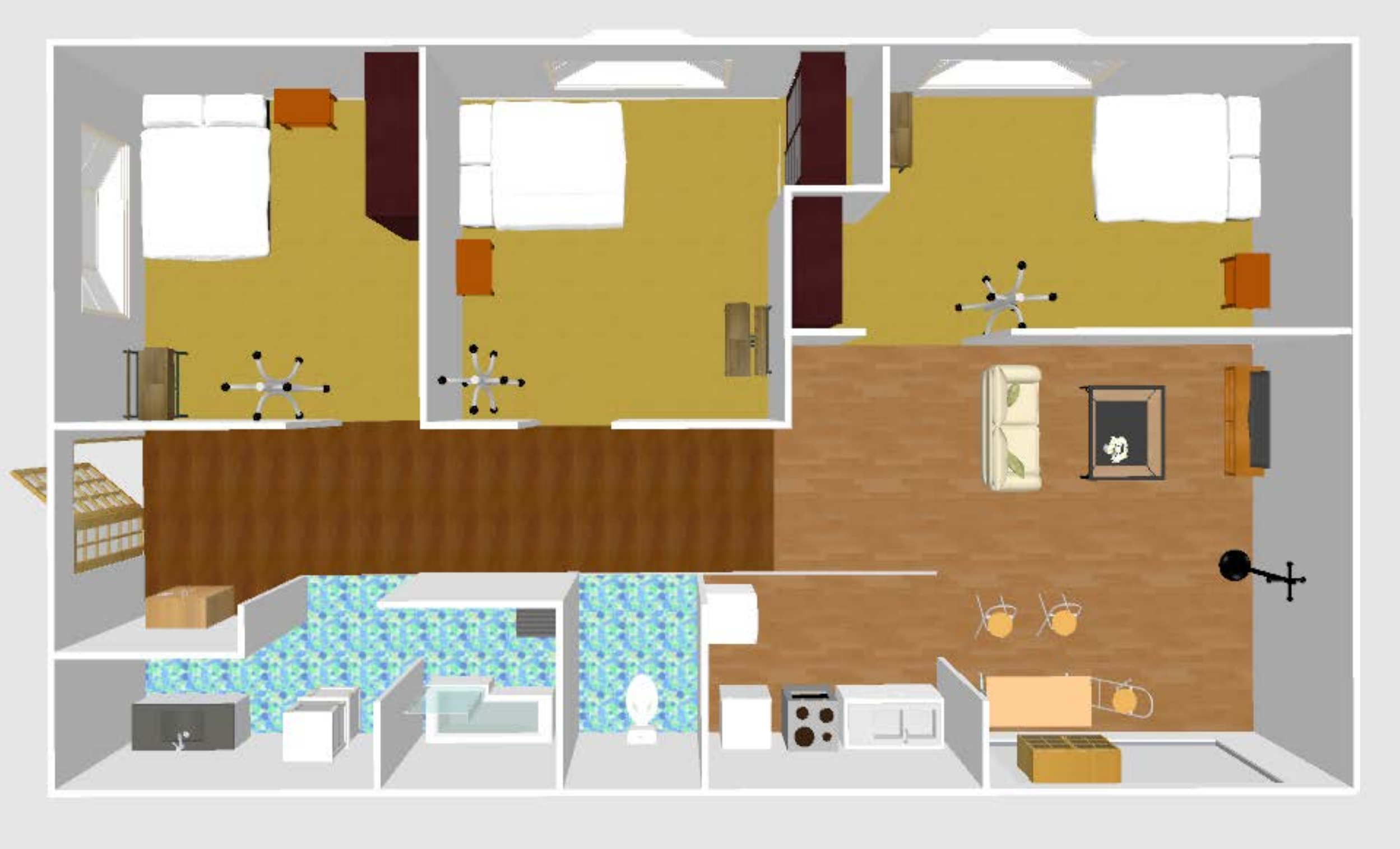}}
        \label{fig:exp_1_environment_sim}}
    \hspace{5pt}
	\subfloat[Map created by SLAM]{
      	\resizebox*{5cm}{!}{\includegraphics[width=0.3\linewidth]{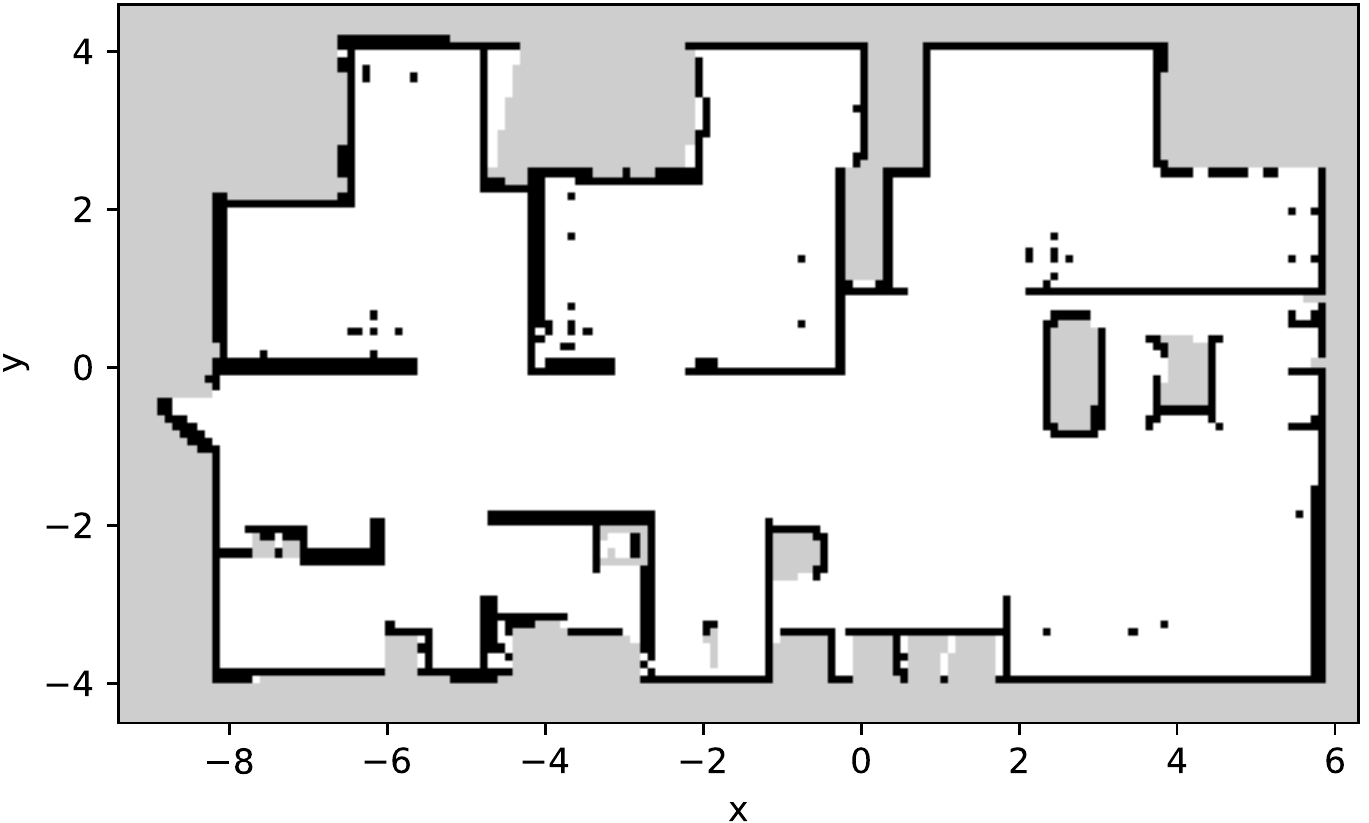}}
        \label{fig:exp_1_environment_map}}
   	\caption{Experimental environment on SIGVerse in Experiment I (Environment 5)}
    \label{fig:exp_1_environment}
\end{figure}

In this experiment, we used a simulator environment with a virtual Toyota \textit{human-support robot (HSR)}~\cite{HSR2019}.
We used a map (Figure~\ref{fig:exp_1_environment_map}) created in advance in the home environment (Figure~\ref{fig:exp_1_environment_sim}) constructed using the SIGVerse~\cite{inamura2021sigverse} simulator as the experimental environment.
SIGVerse is a virtual environment platform for robots that integrates Unity, a game development engine, and \textit{Robot Operating System (ROS)}~\cite{ROS}, a robot software platform.
For mapping, we used the ROS package {\tt gmapping}{\footnote{ROS Wiki: {\tt gmapping}, http://wiki.ros.org/gmapping}}, which was implemented based on the grid-based FastSLAM~2.0~\cite{gridbasedfastslam2005} algorithm. 
The inputs for the SLAM were the depth sensor data from the lidar on the robot as the measured values and odometry data as the control values.

Candidate position coordinates for data observation were set at 0.8 {m} intervals in the white free space in the map in Figure~\ref{fig:exp_1_environment_map}.
Within this space, if there is a wall or an obstacle within 0.5 {m} of the candidate positions, the candidates are deleted to provide space for the HSR with a diameter of 0.43 {m} to move.
In addition, a single exploration was performed for each candidate point, and once a candidate point was searched, it was assumed that it would not be searched again.
It is assumed that the travel to a destination by self-localization and path planning is accurate.

Figure~\ref{fig:exp_2_environment_5} shows the ideal form of the spatial concept and the position distribution set by the tutor.
The color of the candidate points where the data are observed represents the index of the spatial concept $C_{n}$, and the color of the ellipses corresponds to the index of the position distribution $i_{n}$.
The ideal model parameters are set such that for the spatial concept index $C_{n}$, both the index of the position distribution $i_{n}$ and the word of the place are assigned one-to-one. 
Specifically, it is assumed that the same word is observed at all candidate points and indicated by the same color.
The word given at a candidate point corresponding to place is 
\texttt{Living\_room}, \texttt{Dining\_room}, \texttt{Kitchen}, \texttt{Bedroom\_A}, \texttt{Bedroom\_B}, \texttt{Bedroom\_C}, \texttt{Corridor}, \texttt{Toilet}, \texttt{Bathroom}, or \texttt{Entrance}.
\texttt{Bedroom\_A}, \texttt{Bedroom\_B}, and \texttt{Bedroom\_C} are pseudo-representations of the unique names called at certain places in the onsite environment, e.g., \textit{Bob's room} or \textit{cats' playroom}.

The number of particles was set to $R=1000$, and the number of pseudo-observations was $J=10$.
The hyperparameters are set as $\alpha=1.0,\,\beta=0.01,\,\gamma=0.1,\,m_0=[0.0,0.0]^{\rm T},\,\kappa_0=0.001,\,V_0={\rm diag}(1.5,1.5),\,\nu_0=4.0$.
The upper limits for the number of categories were set as $K=10$ and $L=10$.

\subsection{Comparison methods}
\label{sec:experiments:comparison}

The following comparison methods were used:
\begin{enumerate}[(A)]
    \item \textbf{SpCoAE}: The proposed method (IG maximization). 
    \item \textbf{SpCoAE with travel cost}: The proposed method (maximization of the IG with the travel cost).
    The next destination point is determined using Eq.~(\ref{eq:argmax_IG_cost}) described in Section~\ref{sec:proposed:travelcost}.
    The weighting factor for the travel cost is $\eta = 0.005$.
    \item \textbf{Random}: A method to randomly select a destination point with a uniform distribution.
    \item \textbf{Travel cost}: A method to select a destination point in order to decrease travel cost.
    \item \textbf{IG min}: A method to select a destination point in order of decreasing IG.
\end{enumerate}

\subsection{Evaluation metrics}
\label{sec:experiments:evaluation}

\textbf{Evaluation of learning performance}:
As an evaluation metric, we adopted the \textit{adjusted Rand index (ARI)}, which measures clustering performance.
An ARI value close to 0.0 indicates random clustering, whereas an ARI value close to 1.0 indicates high clustering performance.
In this experiment, the ARI value between the estimated result for each particle and the ideal clustering result was calculated, and the sum of the values multiplied by the weight of each particle was evaluated as the weighted ARI.
We evaluated both the index of the spatial concept $C_{n}$ and the index of the position distribution $i_{n}$.
Because the proposed method involves online learning, two metrics are used: (i) ARI (step-by-step) for the observed data only at each step and (ii) ARI (predictive padding) for all data, that is, observed and unobserved data.
In predictive padding, the latent variables for unobserved data are predicted and filled based on the posterior predictive distribution.
This distribution uses the spatial concept parameters $\Theta = \{ \{ \mu_k \}, \{ \Sigma_k \}, \{ \phi_l \}, \{ W_l \}, \pi\}$ estimated at each step.
The ARI (step-by-step) can be expected to show a value close to 1.0 in the first few steps, owing to its nature; however, it does not necessarily indicate a high clustering performance.

\textbf{Evaluation of learning efficiency}:
Two metrics were used to evaluate the learning efficiency.
(i) \textit{Normalized minimum step above threshold (NMS)} is the minimum step that results in $\text{ARI (predictive padding)}\geqq 0.6$, normalized to $[0,100]$ for the final step.
NMS indicates that quick and high learning performance has been achieved.
(ii) \textit{Learning stability rate (LSR)} is the percentage of steps that result in $\text{ARI (predictive padding)}\geqq 0.6$ (i.e., the total number of steps that result in ARI higher than 0.6 / the final step in each environment).
LSR indicates that the learning performance is stable and high for multiple steps.

The lower the NMS and the higher the LSR, the higher the efficiency of quick environmental adaptation in learning.
For spatial concept formation, an ARI of approximately 0.6 or higher is sufficient to perform well on tasks such as navigation~\cite{ataniguchi2020spconavi,ataniguchi2022spcotmhp}.

\textbf{Evaluation of movement efficiency}:
The travel distance at each step was evaluated to compare the travel amount and time spent in the environment.
The travel distance is the total number of cells that move on the occupancy grid map.
The movement path is assumed to be obtained by path planning using the A$^{\star}$ algorithm from the current self-position to the destination position.

\subsection{Results}
\label{sec:exp1:result}

\begin{figure}[tb]
    \centering
	\subfloat[ARI (step-by-step) of~$C_{n}$]{
	    \includegraphics[width=0.30\linewidth]{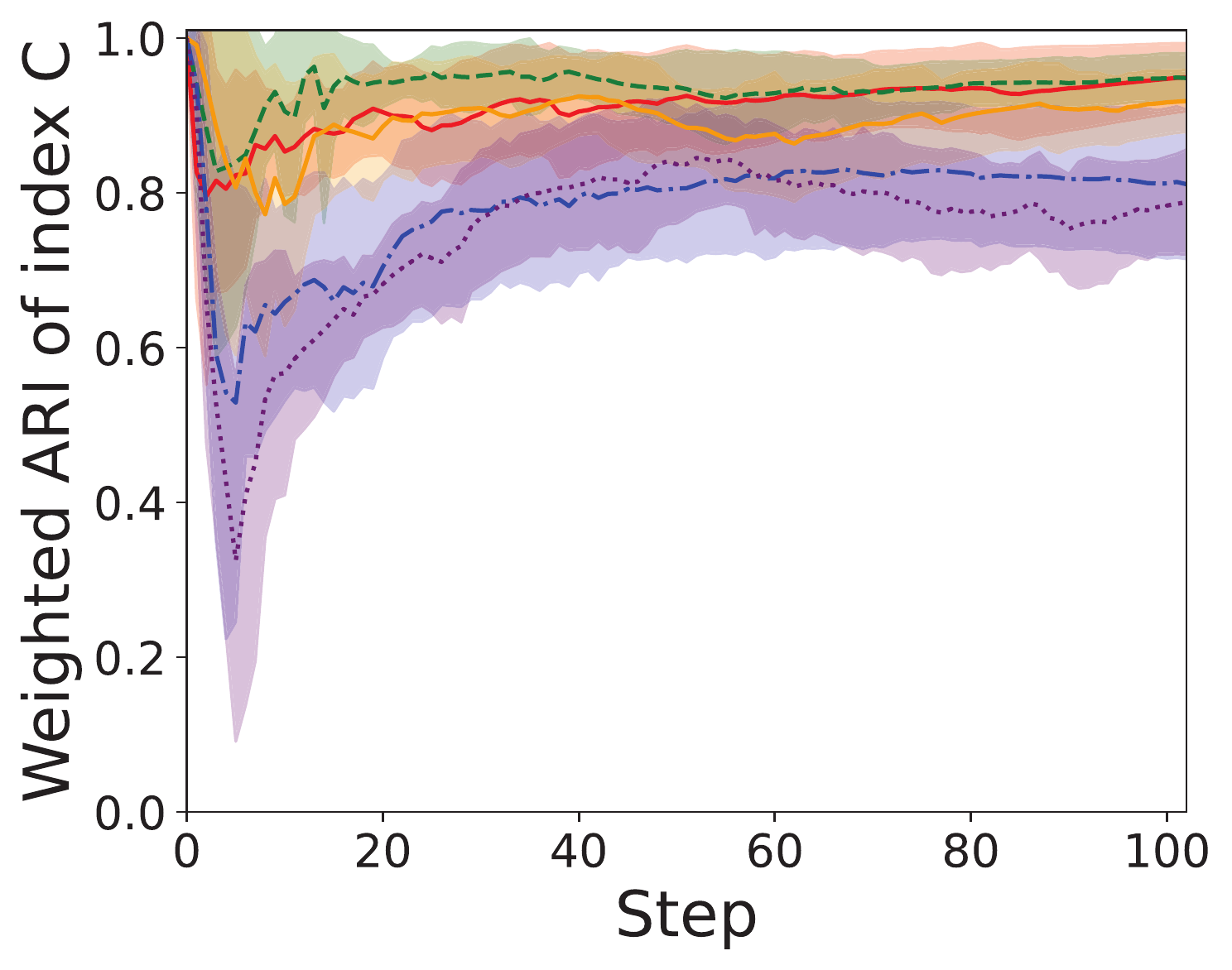}
	    \label{fig:env1_C}}
	\subfloat[ARI (predictive padding) of~$C_{n}$]{
		\includegraphics[width=0.30\linewidth]{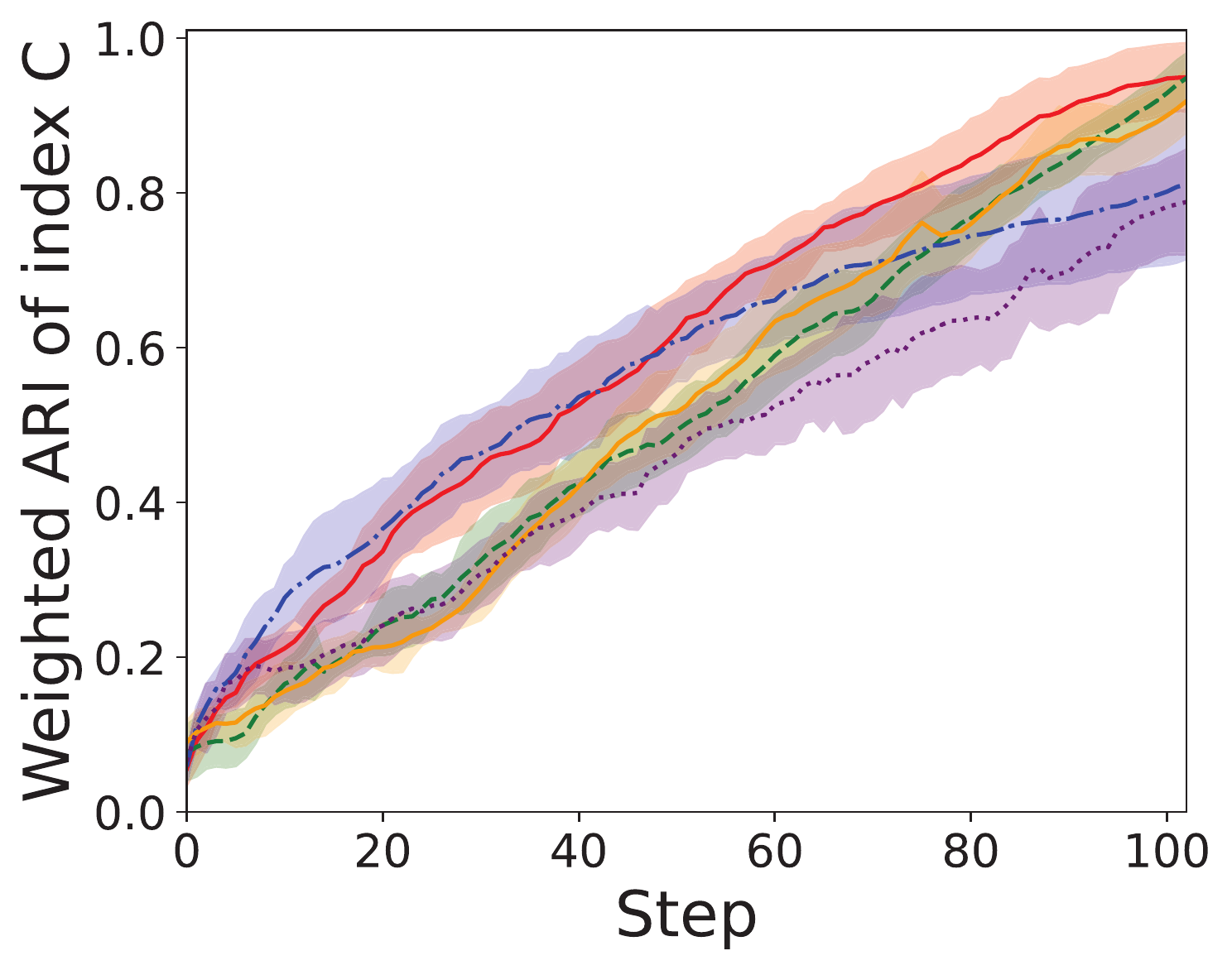}
	    \label{fig:env1_C_2}}	    
   	\subfloat[Cumulative travel distance]{ 
       	\includegraphics[width=0.32\linewidth]{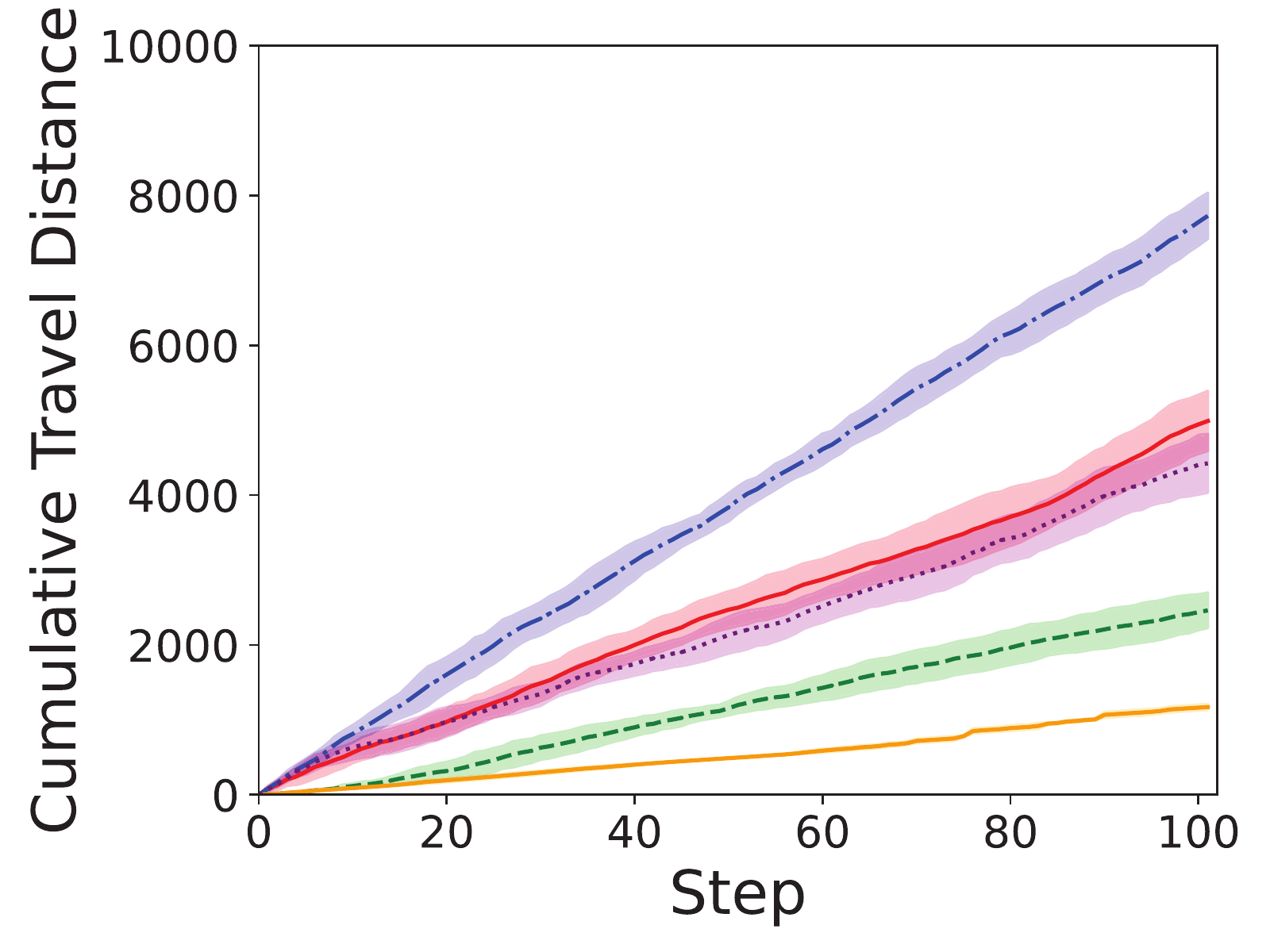}
        \label{fig:env1_travel}}
    \\
    \subfloat[ARI (step-by-step) of~$i_{n}$]{
		\includegraphics[width=0.30\linewidth]{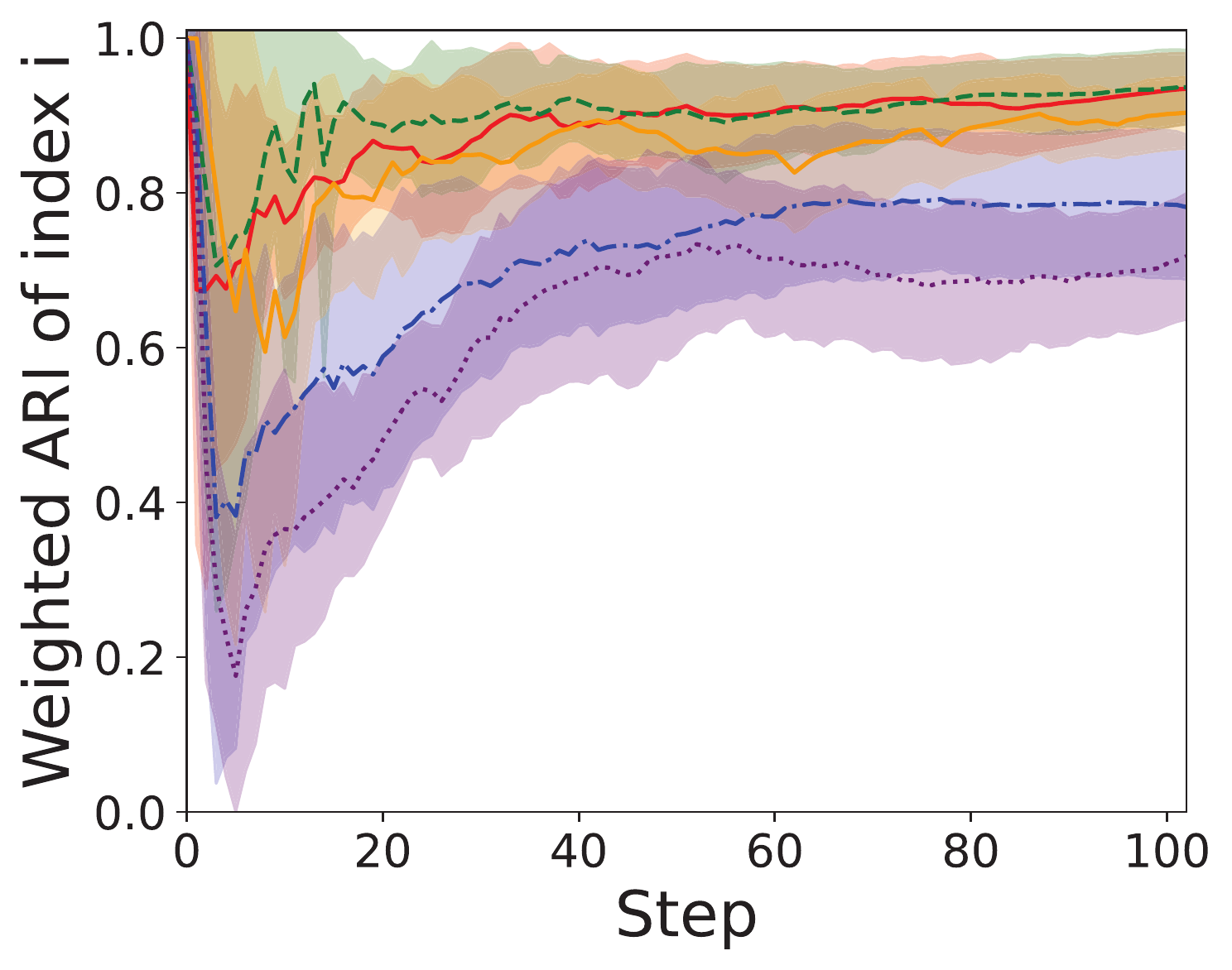}
	    \label{fig:env1_i}}
    \subfloat[ARI (predictive padding) of~$i_{n}$]{
		\includegraphics[width=0.30\linewidth]{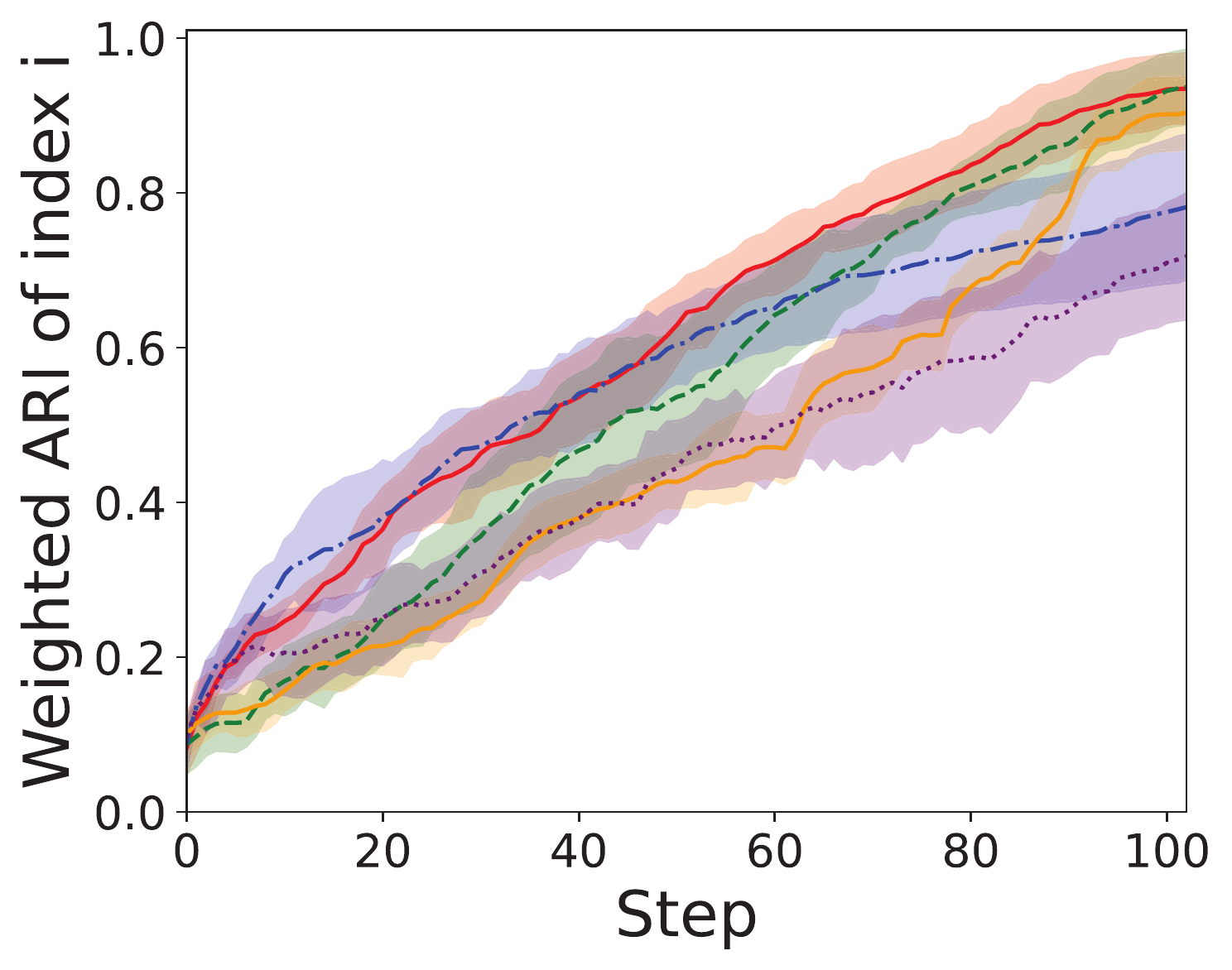}
	    \label{fig:env1_i_2}} 
   	\subfloat[Legends]{
   	    \raisebox{5mm}{
       	\includegraphics[width=0.30\linewidth]{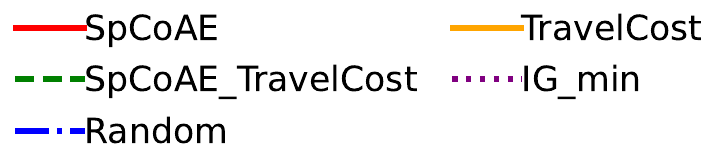}
       	}
        \label{fig:env1_legend}}
   	\caption{Evaluation values for each step (Environment 5)}
    \label{fig:exp_result}
\end{figure}
\begin{figure}[tb]
	\centering
	\subfloat[NMS$\downarrow$ of index~$C_{n}$]{
		\includegraphics[width=0.48\linewidth]{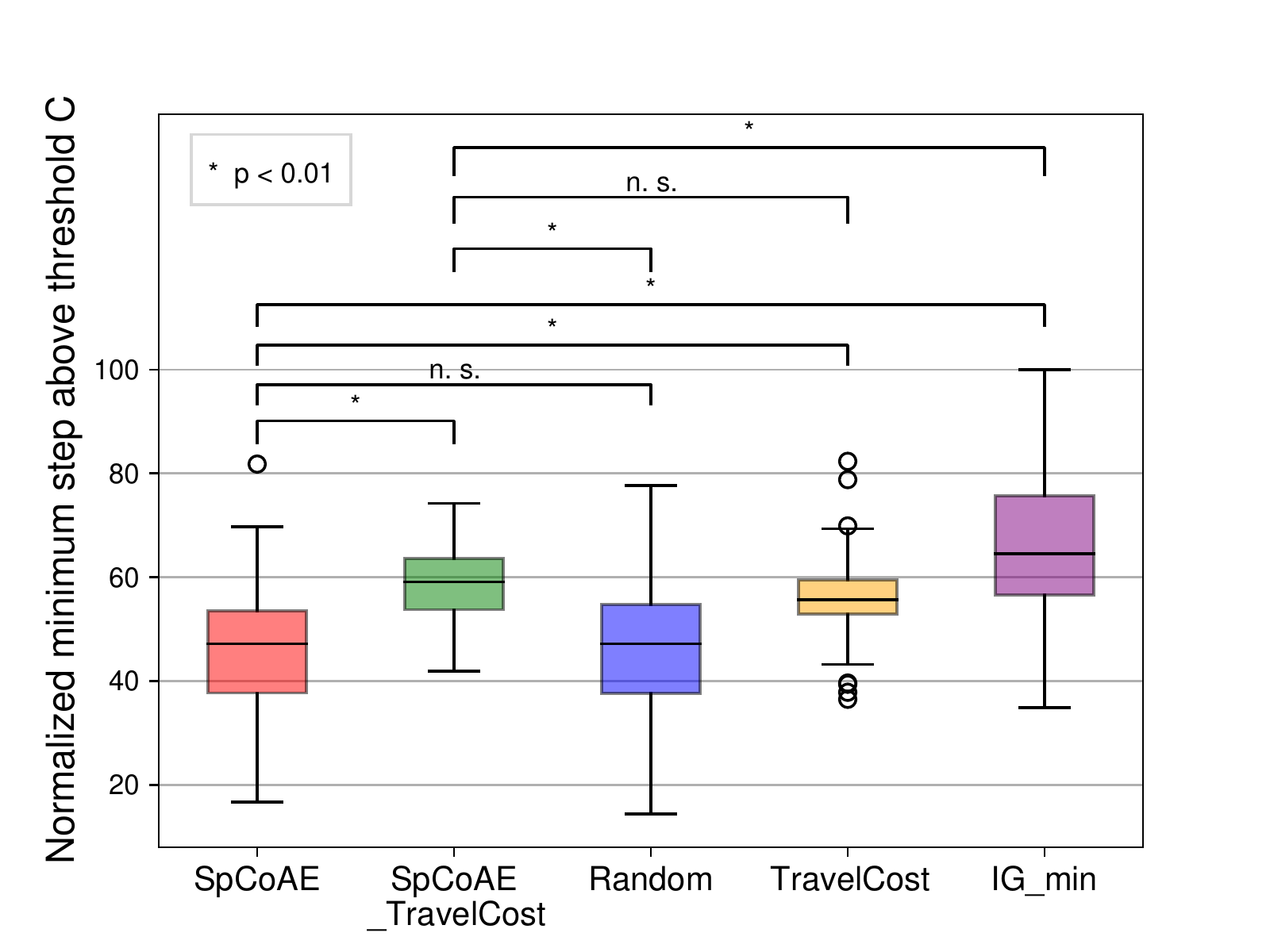}
	    \label{fig:nms_C}}
    \subfloat[NMS$\downarrow$ of index~$i_{n}$]{
		\includegraphics[width=0.48\linewidth]{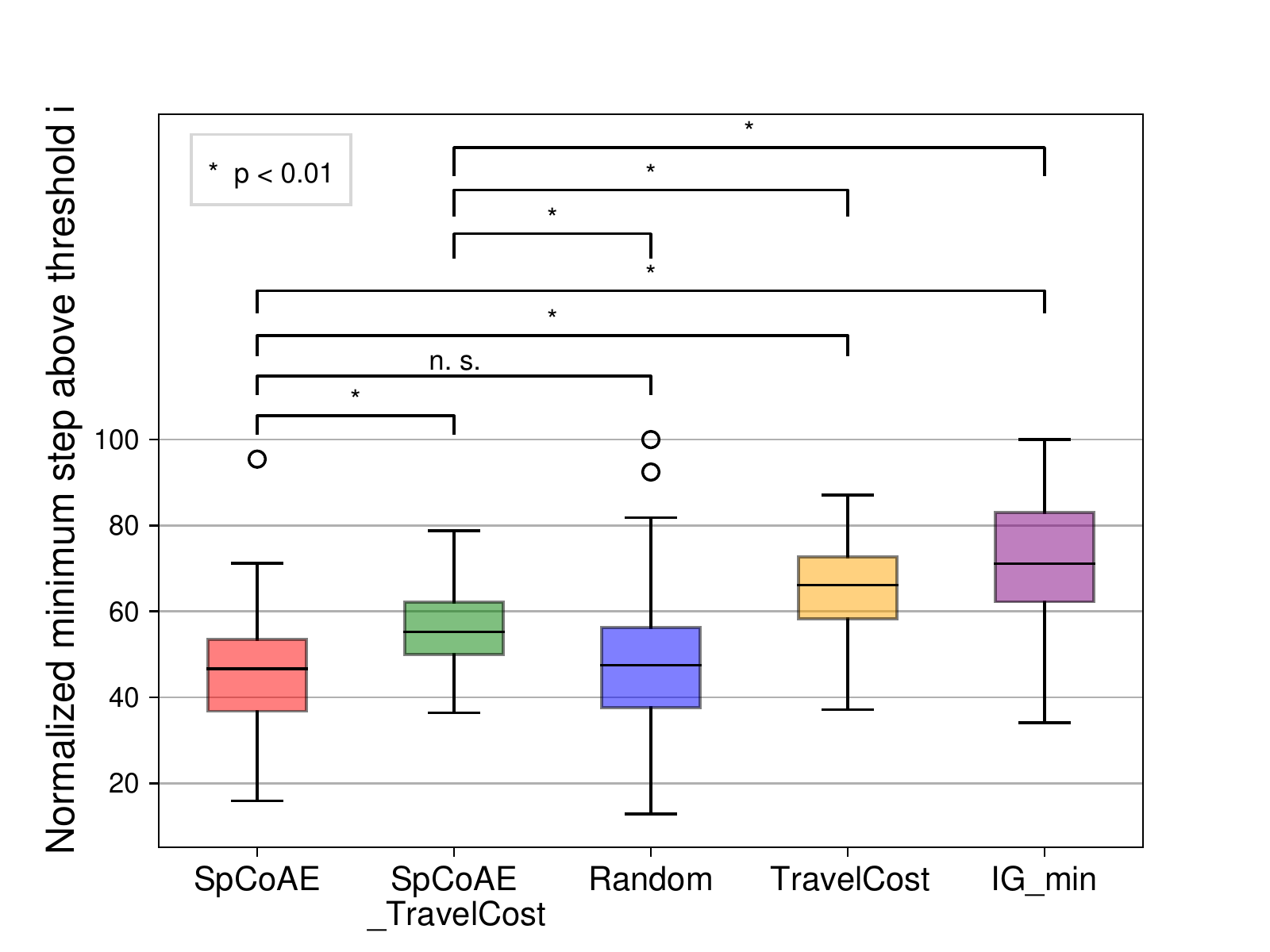}
	    \label{fig:nms_i}}
     \\
	\subfloat[LSR$\uparrow$ of index~$C_{n}$]{
		\includegraphics[width=0.48\linewidth]{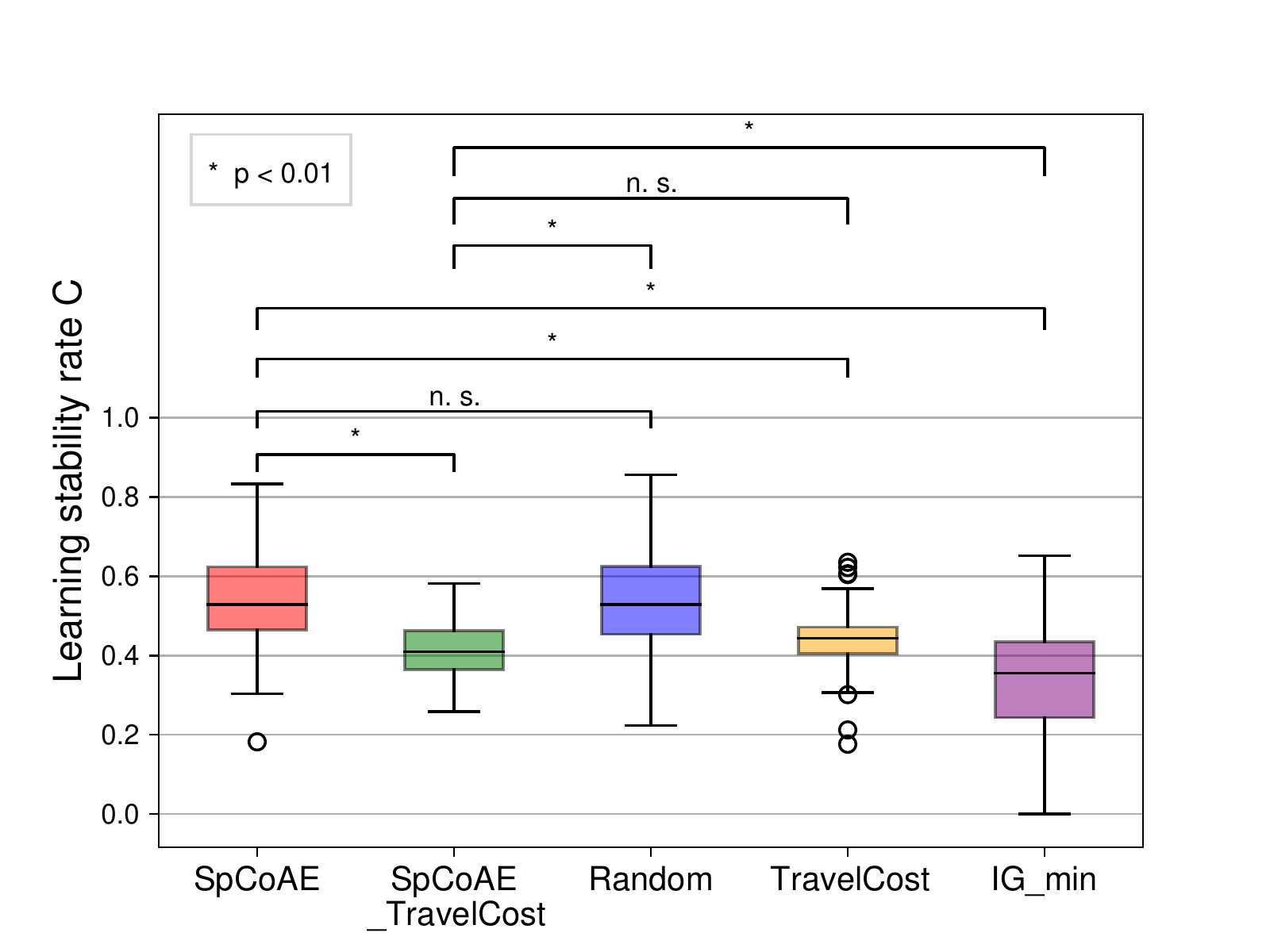}
        \label{fig:lsr_C}}
    \subfloat[LSR$\uparrow$ of index~$i_{n}$]{
		\includegraphics[width=0.48\linewidth]{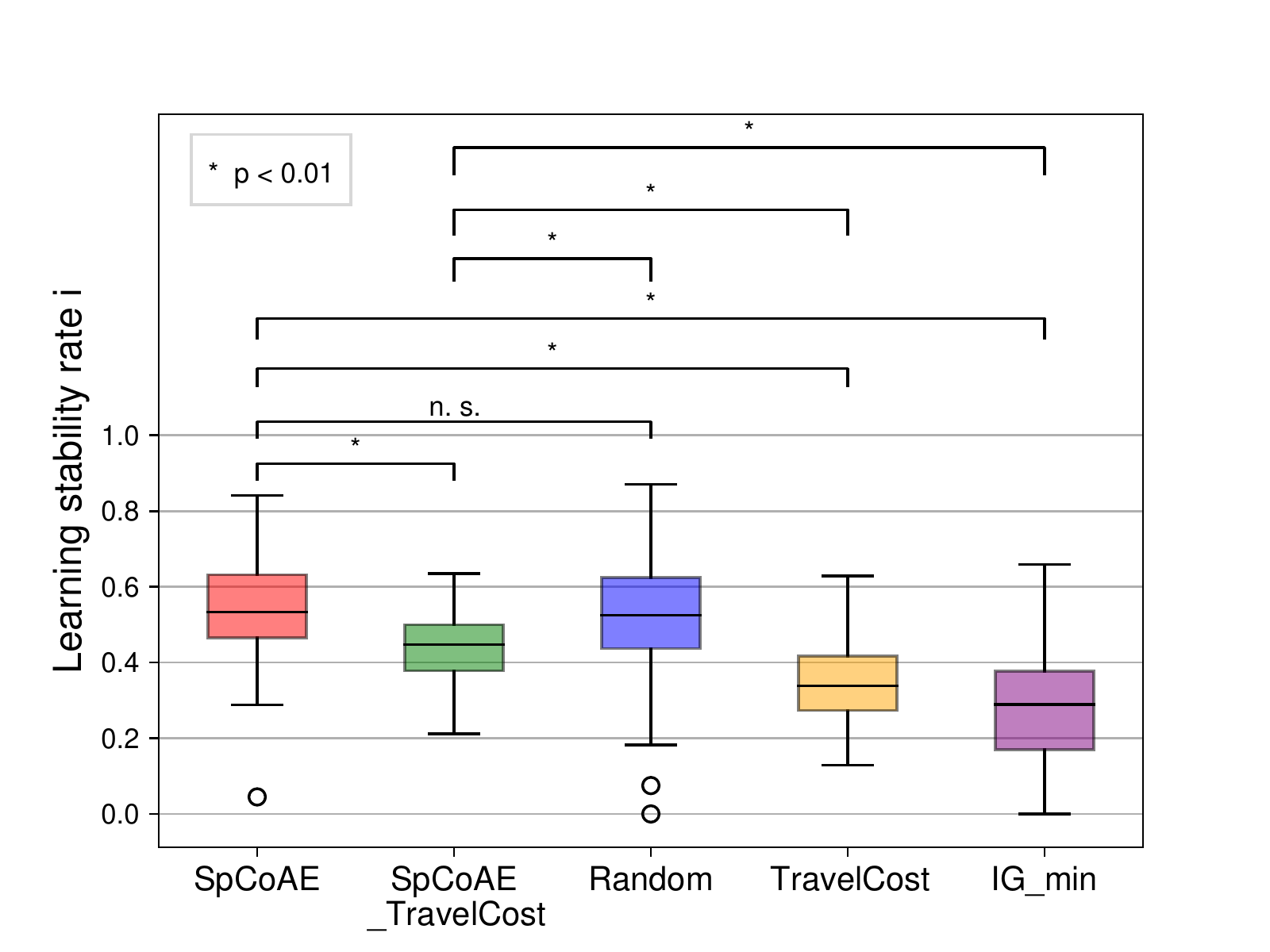}
	    \label{fig:lsr_i}}
   	\caption{
   	    Box-and-whisker plot of learning efficiency constructed using the evaluation values of ten trials gathered in each of the ten environments.
   	    The up or down arrows point in the direction of the desired value.
   	    The statistical significance between the methods was checked by Welch's t-test with Bonferroni's adjustment for multiple comparisons. 
   	    A p-value of less than 0.01 was considered statistically significant.
   	}
    \label{fig:exp_efficiency}
\end{figure}
\begin{table}[tb]
    \tbl{
        Evaluation results (mean and standard deviation) of ten trials in each of the ten environments. 
    }{
    \begin{tabular}{lccc} \hline
		& \multicolumn{2}{c}{\textbf{ARI}} & \textbf{Travel distance} \\  
    \textbf{Methods} & $C_{n}$ & $i_{n}$ & [grid/step]\\ \hline
    SpCoAE                  & \underline{0.942} (0.055) & \underline{0.920} (0.067) & 48.96 (7.90) \\
    SpCoAE with travel cost & \underline{\textbf{0.953}} (0.041) &  \underline{\textbf{0.938}} (0.055) & \underline{22.79} (4.36) \\
    Random                  & 0.878 (0.076) & 0.829 (0.087) & 71.37 (9.82) \\
    Travel cost             & 0.936 (0.053) & 0.919 (0.066) & \underline{\textbf{12.07}} (1.09) \\
    IG min                  & 0.832 (0.094) & 0.755 (0.116) & 40.86 (6.17) \\
    \hline 
    \end{tabular}
    }
    \label{tbl:hyouka_real_learning}    
\end{table}
%

\begin{figure}[tb]
	\centering
	\subfloat[Ideal]{
       	\includegraphics[width=0.30\linewidth]{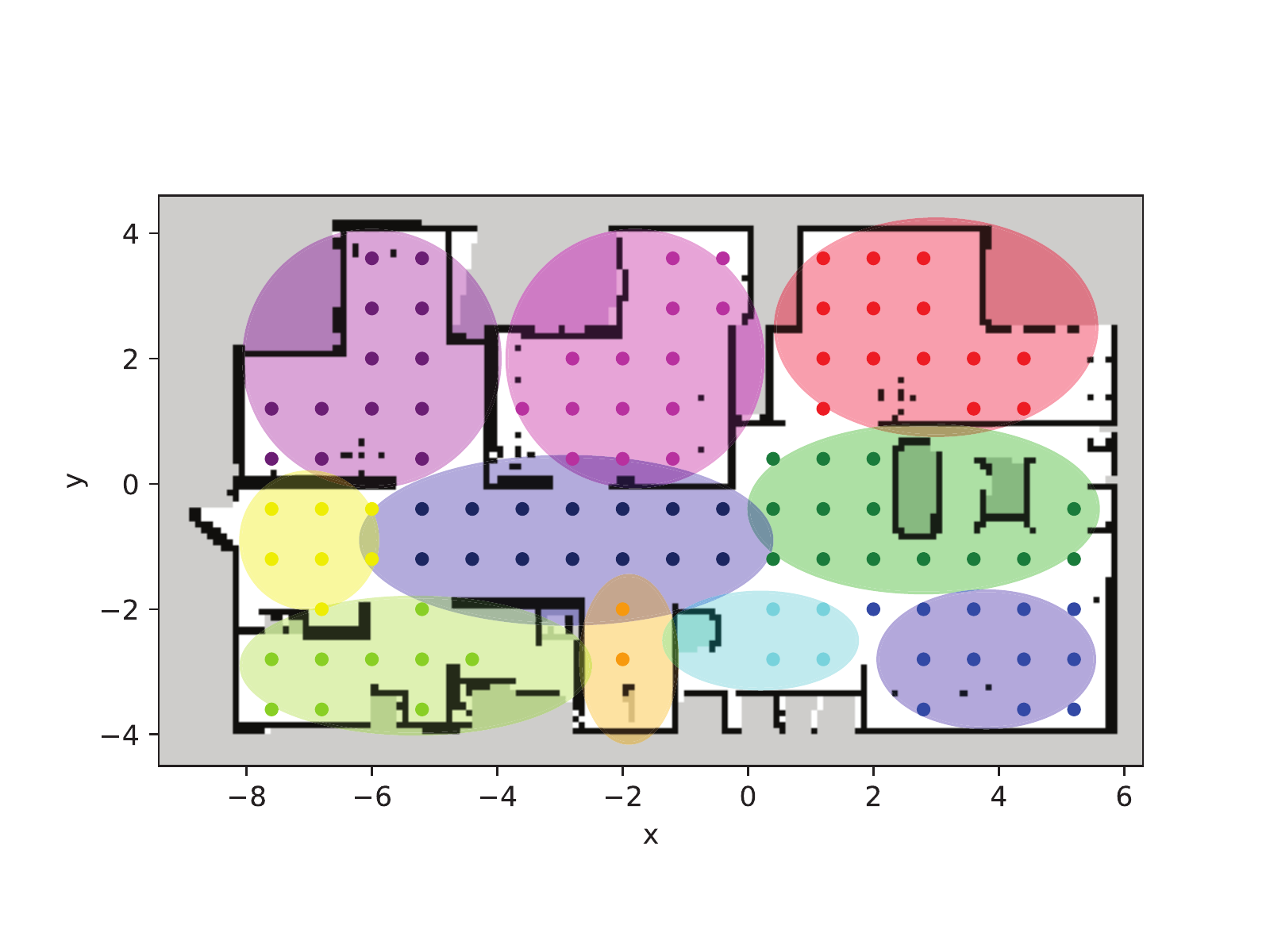}
        \label{fig:exp_2_environment_5}}
   	\subfloat[SpCoAE]{
       	\includegraphics[width=0.30\linewidth]{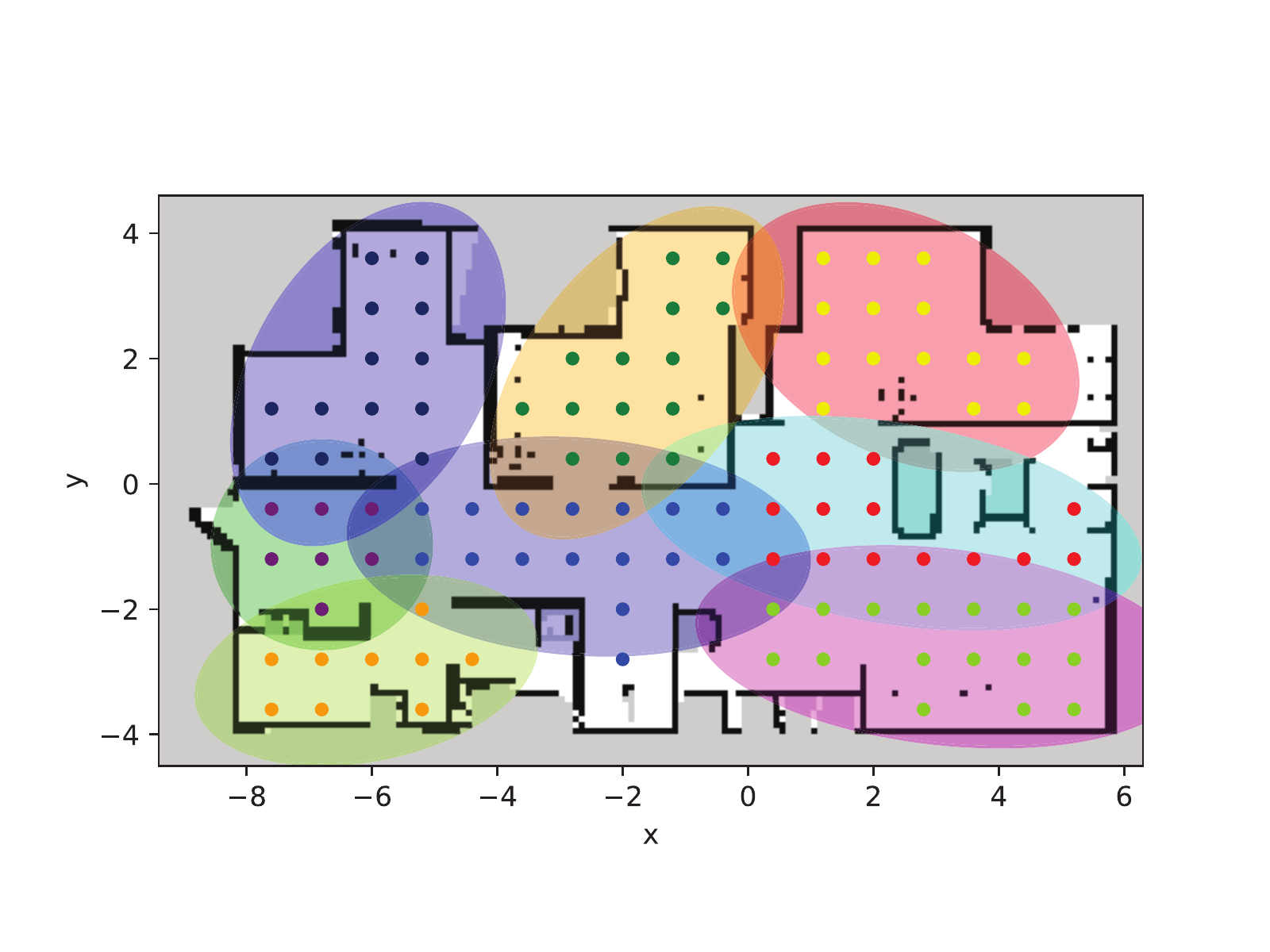}
        \label{fig:exp_1_result_env5_SpCoAE}
        }
	\subfloat[SpCoAE with travel cost]{
      	\includegraphics[width=0.30\linewidth]{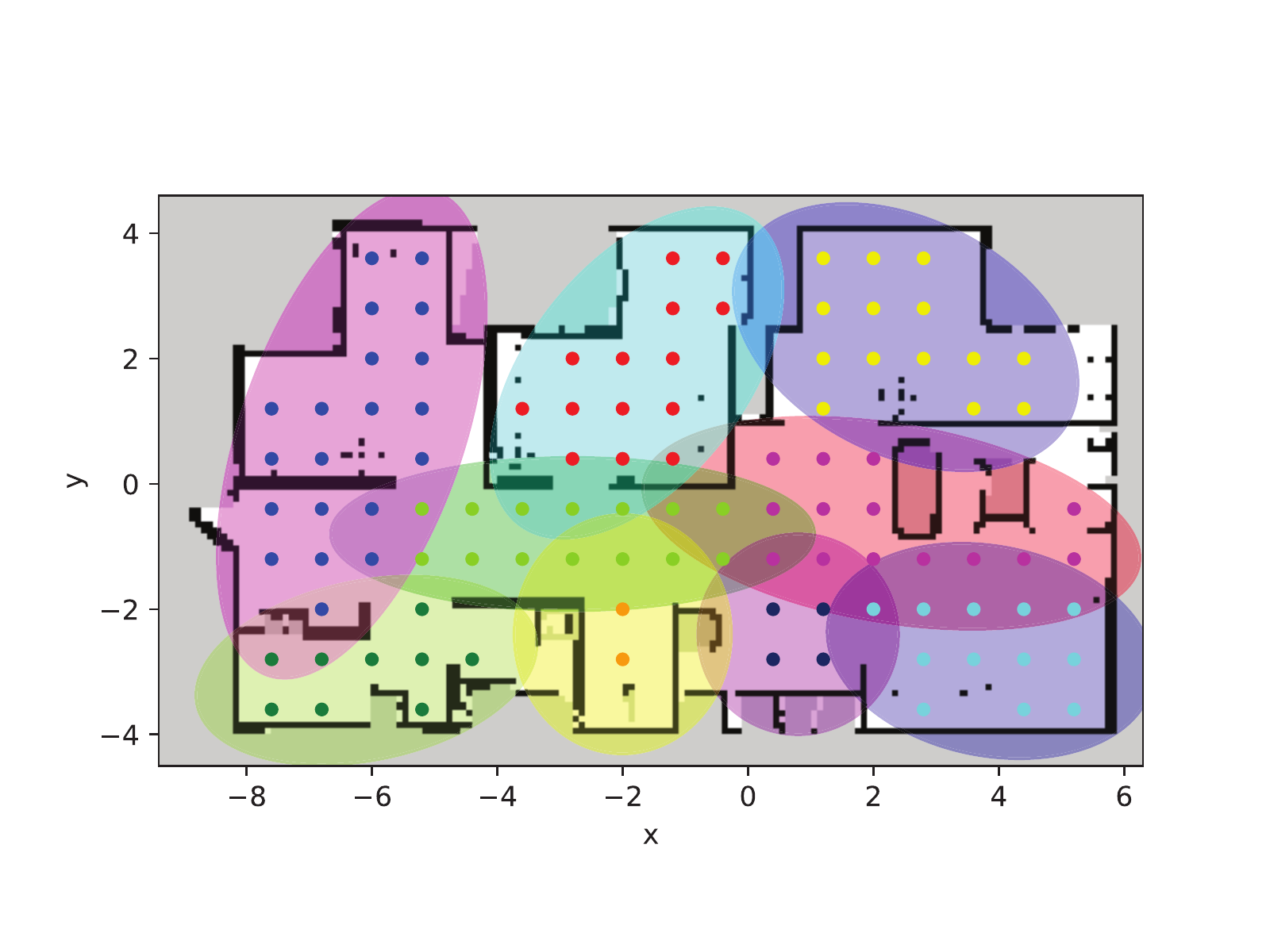}
       \label{fig:exp_1_result_env5_SpCoAE_travel}
        }
        \\
   	\subfloat[Random]{
       	\includegraphics[width=0.30\linewidth]{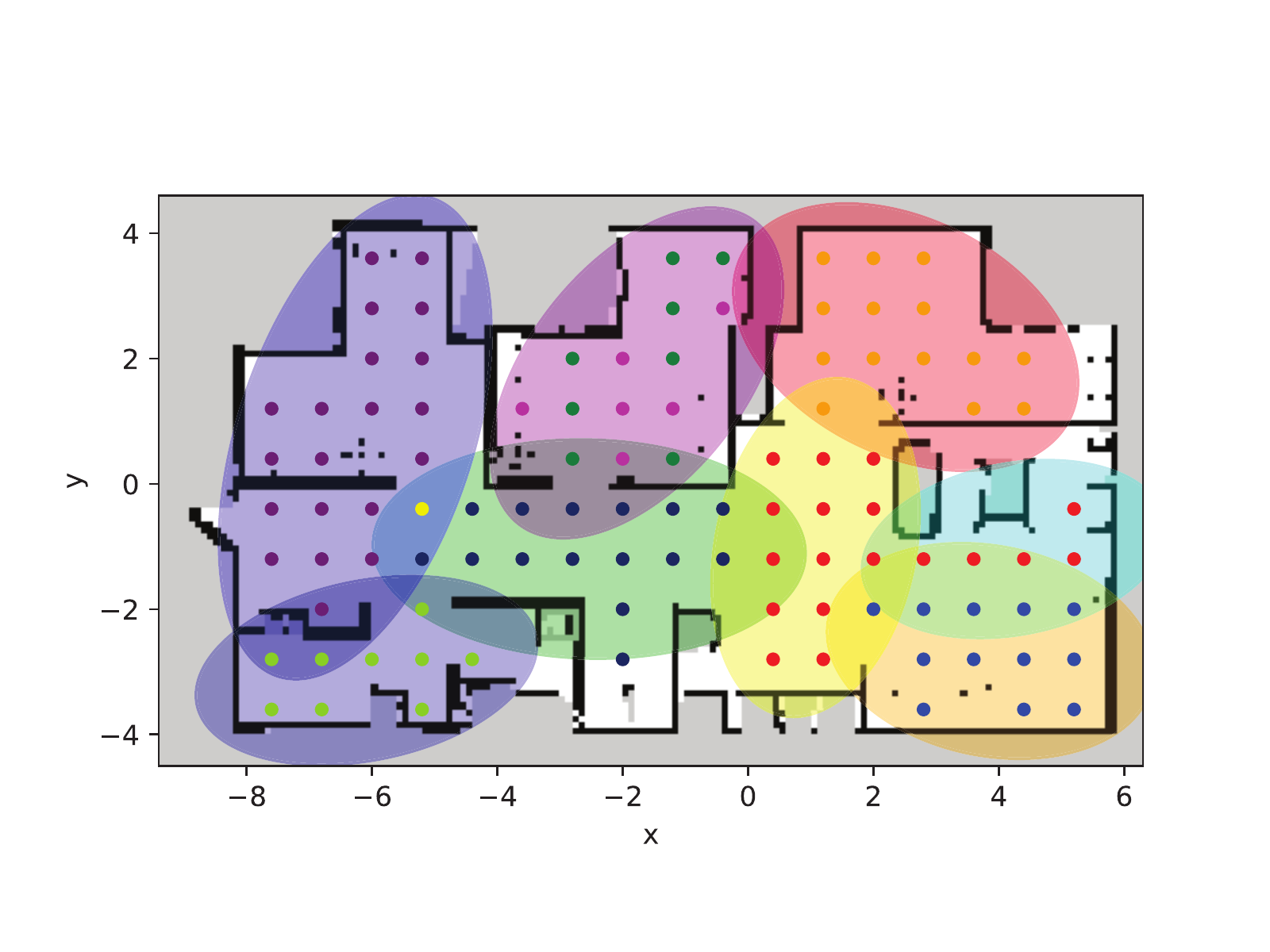}
        \label{fig:exp_1_result_env5_random}
        }
	\subfloat[Travel cost]{
      	\includegraphics[width=0.30\linewidth]{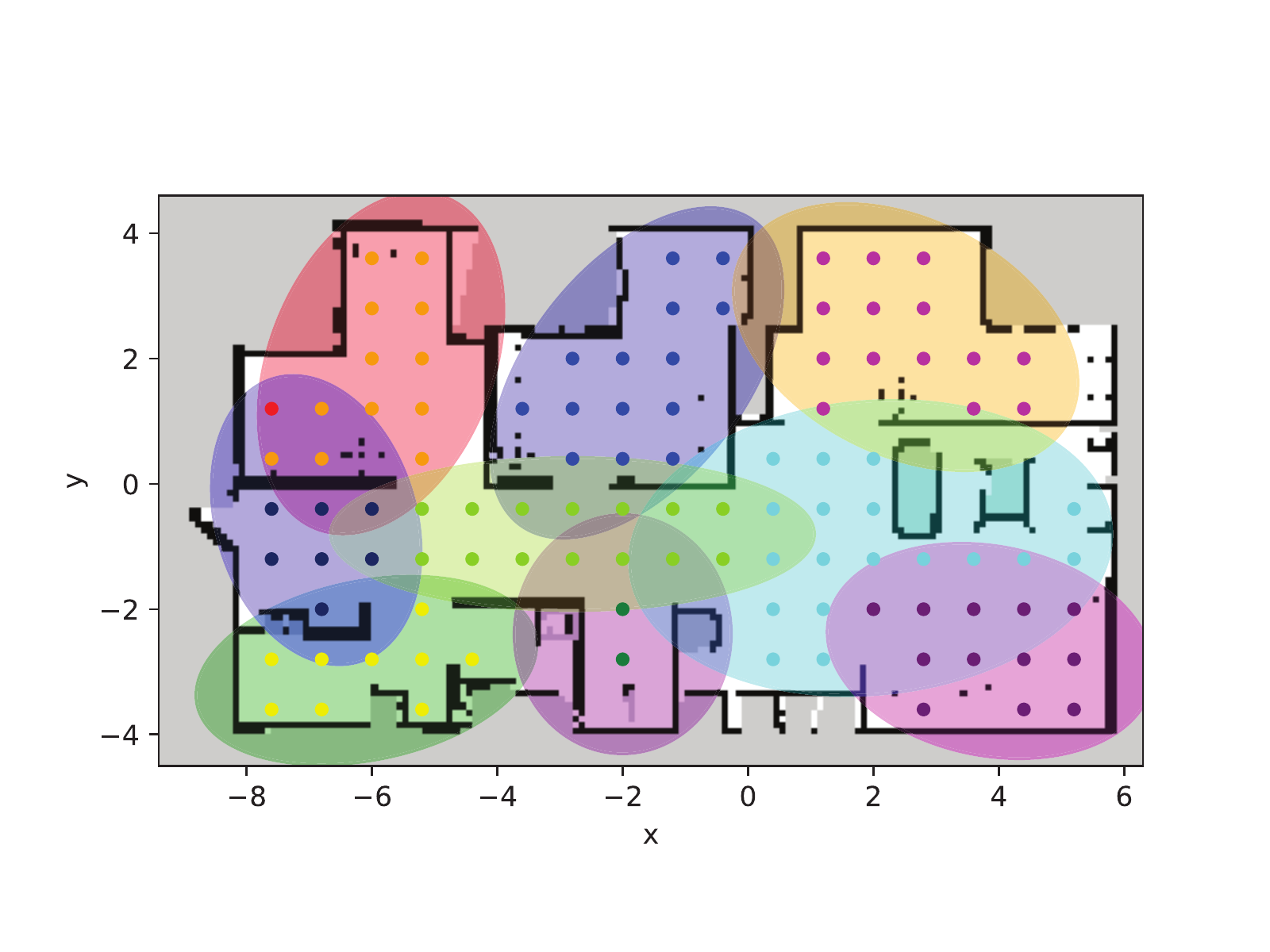}
       \label{fig:exp_1_result_env5_travel}
        }
	\subfloat[IG~min]{
       	\includegraphics[width=0.30\linewidth]{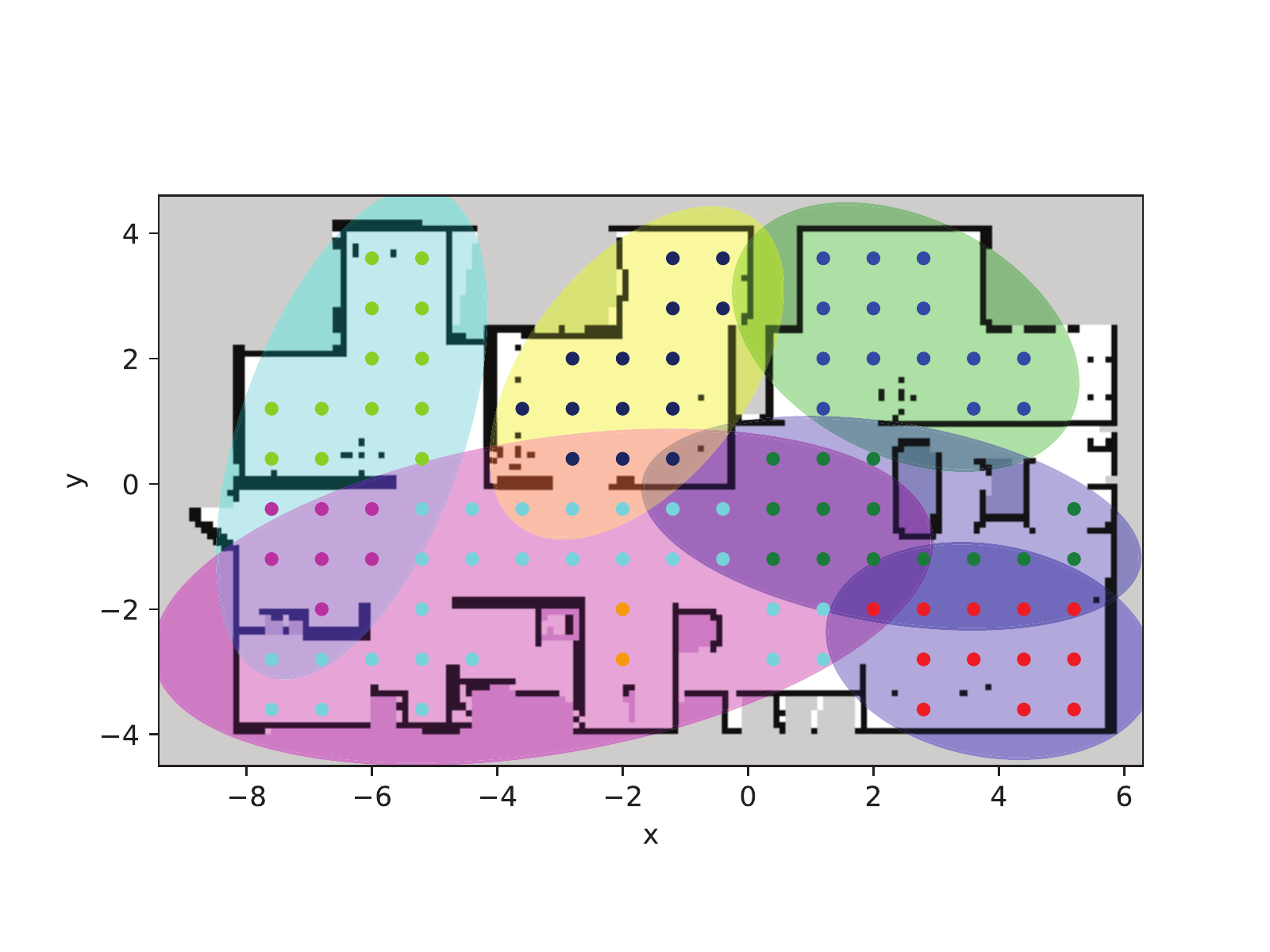}
        \label{fig:exp_1_result_env5_IGmin}
        }
   	\caption{
		Examples of learning results for each method (Environment 5): 
		These results are for the highest weighted particle in the final step.
		Ellipses drawn on map are position distributions $\{ \mu_{k}, \Sigma_{k} \}$.
        Colors of ellipses indicate indices~$i_{1:n}$ of position distributions.
		Colors of candidate points indicate indices~$C_{1:n}$ of spatial concepts. 
		Colors of each ellipse and each candidate point are randomly determined for each trial.
		Sub-figure (a) shows the ideal form envisioned by the tutor.
		This means that clustering results close to (a) are desirable.
   	}
    \label{fig:exp_1_result_env5}
\end{figure}

\begin{figure}[tb]
    \centering
	\includegraphics[width=0.68\linewidth]{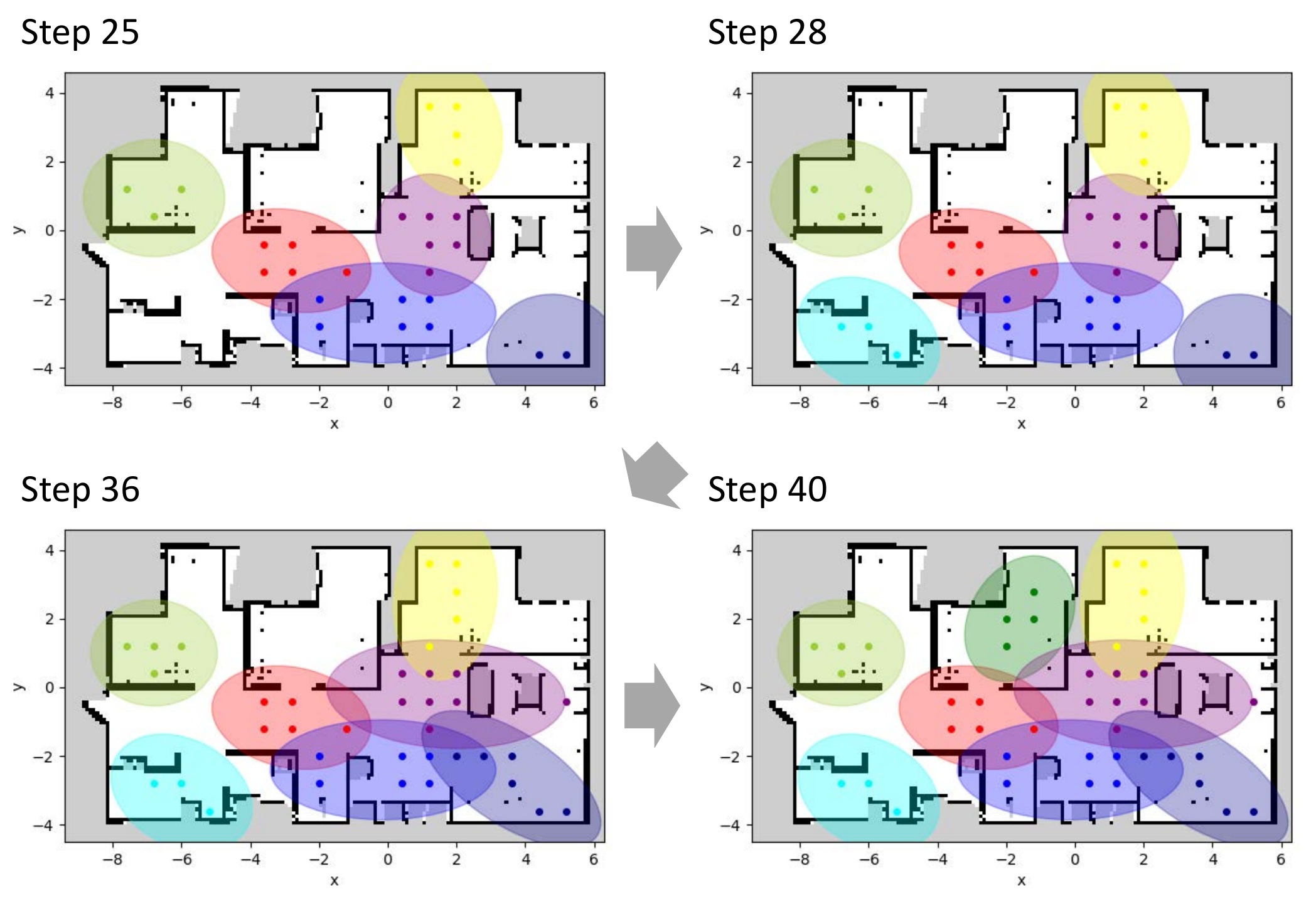}
	\caption{
        Learning process of spatial concepts and position distributions by active exploration of the proposed method in several steps (Environment 5).
        Each point indicates the position from which the robot has acquired data thus far.
    }
	\label{fig:result_example}
\end{figure} 
%

\textbf{Evaluation values for each step}:
Figure~\ref{fig:exp_result} shows the mean and error bars for the standard deviation of each evaluation value for all ten trials\footnote{The spatial concept learning results and evaluation values per environment for the ten environments are presented in Appendix~\ref{apdx:exp1:results}.}.
In terms of the ARI (step-by-step), overall, SpCoAE showed higher values than the other methods.
This result is attributed to the fact that SpCoAE can select a destination that allows spatial concepts to be learned more accurately in each step.
In addition, SpCoAE with travel cost had consistently higher values than SpCoAE in steps 10 -- 40.
In terms of the ARI (predictive padding), SpCoAE had slightly inferior values compared with the random method in the initial steps, whereas it showed higher values in the later steps.
SpCoAE is more effective for learning precise spatial concepts.
In addition, SpCoAE with travel costs exhibited the highest value in the final step, although it was inferior to SpCoAE in the first half.
These results suggest that searching for one place after another in the environment is advantageous for predicting unexplored regions during the learning process.
In terms of the cumulative travel distance, introducing travel cost into SpCoAE resulted in a smaller value.
The random method was less efficient for movement because the travel distance was greater.

\textbf{Evaluation result of learning efficiency}: 
Figure~\ref{fig:exp_efficiency} compares the methods for LER and NMS with $C_{n}$ and $i_{n}$.
The results show that SpCoAE and the random method have higher learning efficiencies than the other methods.
This indicates that these methods are effective for rapidly covering an entire environment.

\textbf{Evaluation values at the last step}: 
Table~\ref{tbl:hyouka_real_learning} summarizes the evaluation results for all ten environments.
The travel distance was normalized by the number of candidate search points (i.e., the number of final steps) to adjust for the impact of the scale of each environment.
Consequently, the proposed method shows a higher ARI than the other methods; in particular, SpCoAE with travel costs has the highest value.
Furthermore, in terms of travel distance, the introduction of travel costs is effective, even in environments with different structures.
Thus, overall, IG-based SpCoAE, specifically SpCoAE with travel costs, is shown to be effective.
%

\textbf{Qualitative results}: 
Figure~\ref{fig:exp_1_result_env5} shows the results of learning spatial concepts for each exploration method in Environment 5.
The proposed methods were closer to the ideal form than the other methods. However, the Random and IG min methods have a higher tendency to combine multiple locations into a single category than the proposed methods.
For example, in Figure~\ref{fig:exp_1_result_env5_random}, the bedroom and entrance, as well as the hallway and bathroom, had the same distribution.
In Figure~\ref{fig:exp_1_result_env5_IGmin}, a large distribution can be observed across the bathroom, toilet, hallway, and kitchen.
This indicates that the order of exploration affects the learning of spatial concepts.
%
Figure~\ref{fig:result_example} shows an example of sequential spatial concept formation constructed using SpCoAE.
In Steps 25--28, the robot searches the three bottom-left locations consecutively.
Steps 36--40 focused on the upper-middle room.
This indicates that SpCoAE can sufficiently reduce the uncertainty regarding a location by searching several times in a row in areas where the location name is not provided and where there is uncertainty.
In other words, the robot can reduce the entropy of the word distribution for a location by obtaining multiple instructions in close positions.
It then searches for a different location.
This means that the robot has obtained enough observations about the location.

\textbf{Discussion}:
According to our results, SpCoAE with travel costs achieved a higher ARI than SpCoAE. 
This can be attributed to the exploration tendency of pure IG, which tends to focus on the outer areas of the environment, potentially hindering the exploration of the central area. 
By incorporating the travel cost, we consider that the exploration candidate points were more evenly distributed throughout the environment, including in the central areas. 
Based on the AIF framework, the travel cost can be considered a pragmatic/extrinsic value with a preference, and can be regarded as an appropriate approximation of the expected free energy. 
This finding highlights the importance of considering both IG and distance in exploration and learning strategies for effective semantic mapping.

\section{Experiment II: Real environment}
\label{sec:exp2}

The effectiveness of the proposed method was investigated in a realistic environment.
This experiment relaxed some of the limitations of Experiment I and demonstrated the method in a more realistic scenario.

\subsection{Condition}

\begin{figure}[tb]
	\centering
	\subfloat[HSR~\cite{HSR2019}]{
       	\includegraphics[width=0.18\linewidth]{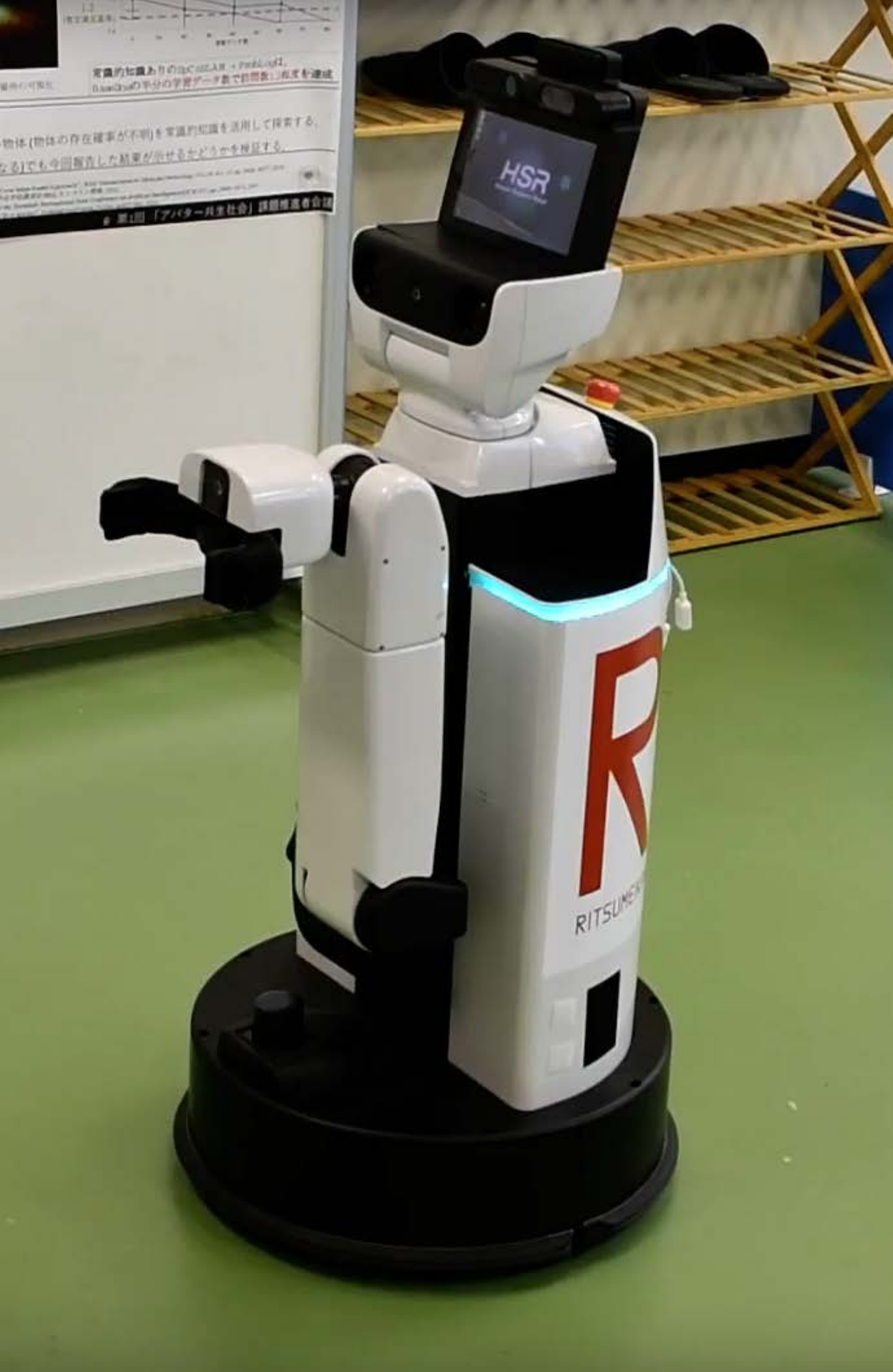}
        \label{fig:exp_2_environment_hsr}}	
    \subfloat[Real environment]{
   	    \includegraphics[width=0.40\linewidth]{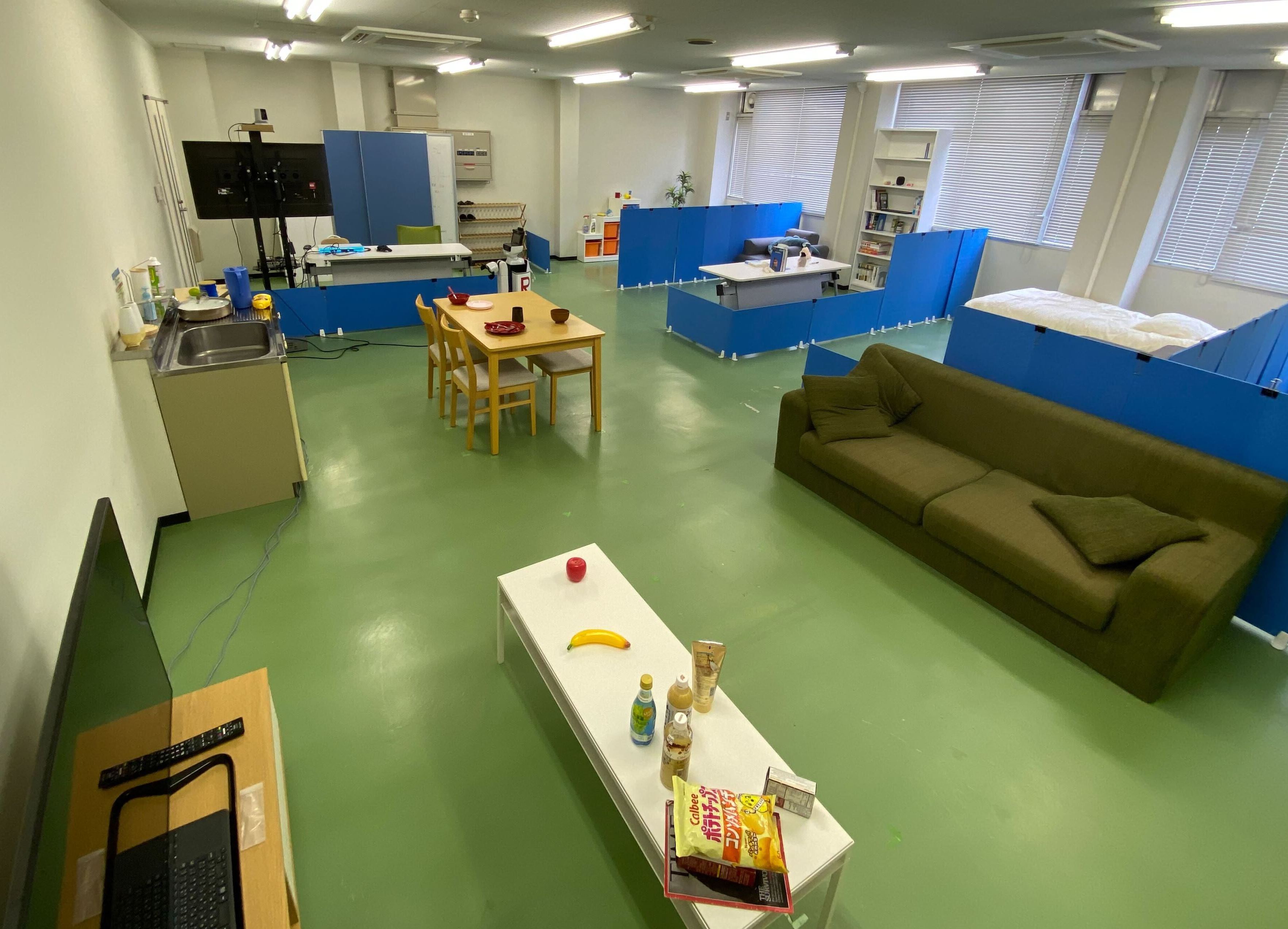}
        \label{fig:exp_2_environment_1}
   }
    \subfloat[Floor plan]{
   	    \includegraphics[width=0.32\linewidth]{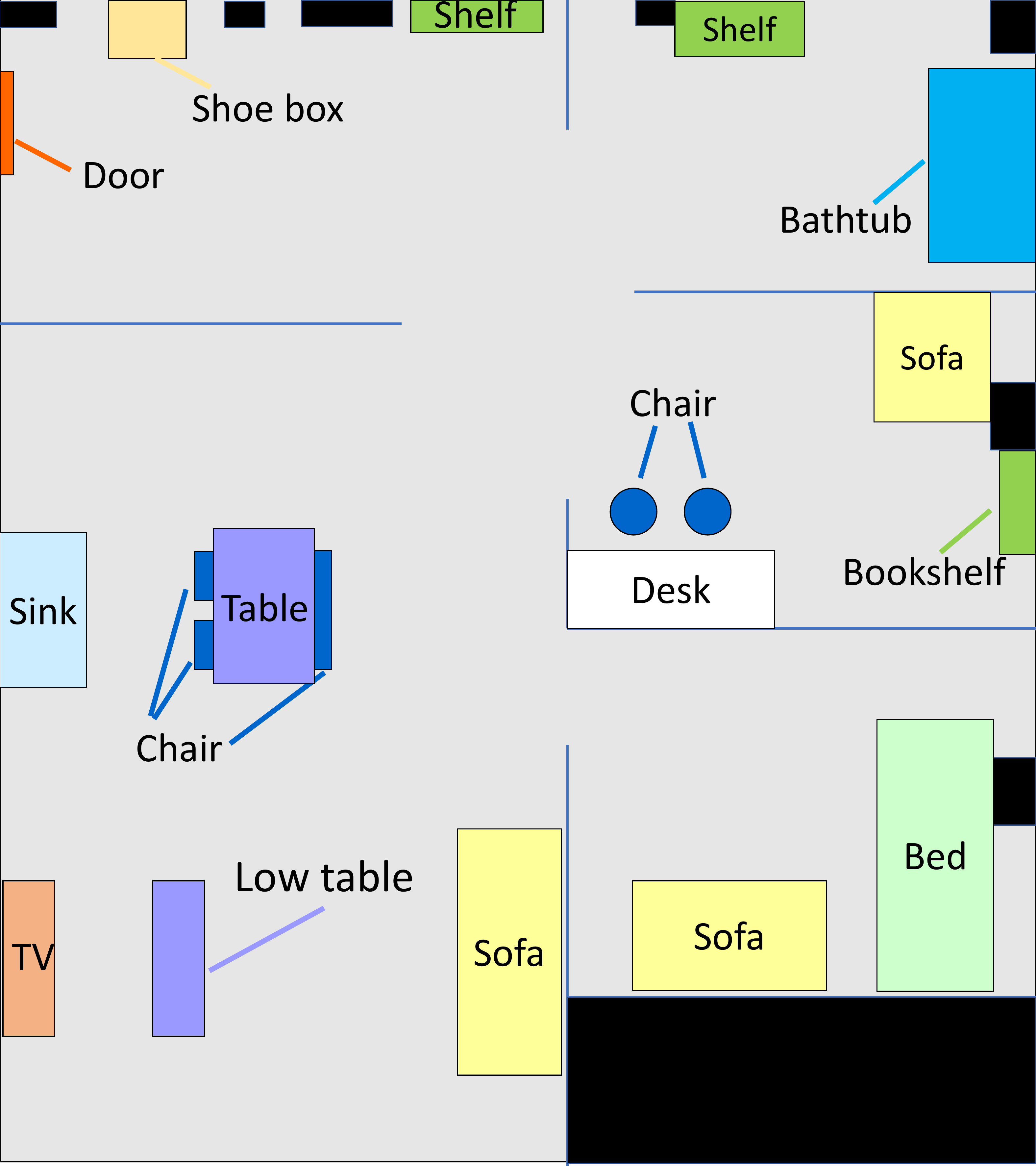}
        \label{fig:exp_2_floor_plan}
    }
   	\caption{Real robot and real experimental environment in Experiment II:
        The size of the environment is 8 m $\times$ 9.4 m with an area of approximately 74 m$^{2}$.
        In this environment, six areas are assumed.
        The number of candidate search points is 53.
    }
    \label{fig:exp_2_environment}
\end{figure}

A room simulating a home environment was used as the experimental environment.
Figure~\ref{fig:exp_2_environment} shows the actual environment and the robot.
This experiment was set up in a more difficult setting than Experiment I owing to the increased uncertainty.
First, the linguistic information provided by the user comprises multiword sentences. 
We assume that the sentence is the response given by the user when the robot asks, `Tell me what kind of place is this?'
A different sentence is provided for each candidate point.
Examples of these sentences are as follows: 
`A low table is placed between the TV and the sofa, where one can place drinks and other refreshments.' 
`The room in which you sleep at night is called a bedroom.' 
`The bathroom is used after first filling the bathtub with hot water for washing the body.'
This linguistic information is converted into a bag-of-words representation by preprocessing as follows:
(i) Nonletters have been removed.
(ii) The hyphens were removed to separate them before and after. 
(iii) All uppercase letters were converted to lowercase letters.
(iv) Nouns are unified into singular forms and
(v) Stop words were removed.

Secondly, in this experiment, a single candidate was explored several times as a challenge.
This is because visiting a single candidate point once may not provide sufficient information.
In addition, visiting candidate points with a low priority should not be explored.
In this case, the robot explored up to 100 steps because it could explore infinitely.

The number of particles was set to $R=1000$, and the number of pseudo-observations was set to $J=10$.
The hyperparameters are set as follows: $\alpha=1.0,\,\beta=0.1,\,\gamma=0.01,\,m_0=[0.0,0.0]^{\rm T},\,\kappa_0=0.001,\,V_0={\rm diag}(1.0,1.0),\,\nu_0=5.0$.
The upper limits for the number of categories were set as $K=10$ and $L=10$.

%

\subsection{Result}

\begin{figure}[tb]
    \centering
    \subfloat[Step 5]{
    	\includegraphics[width=0.32\linewidth,clip]{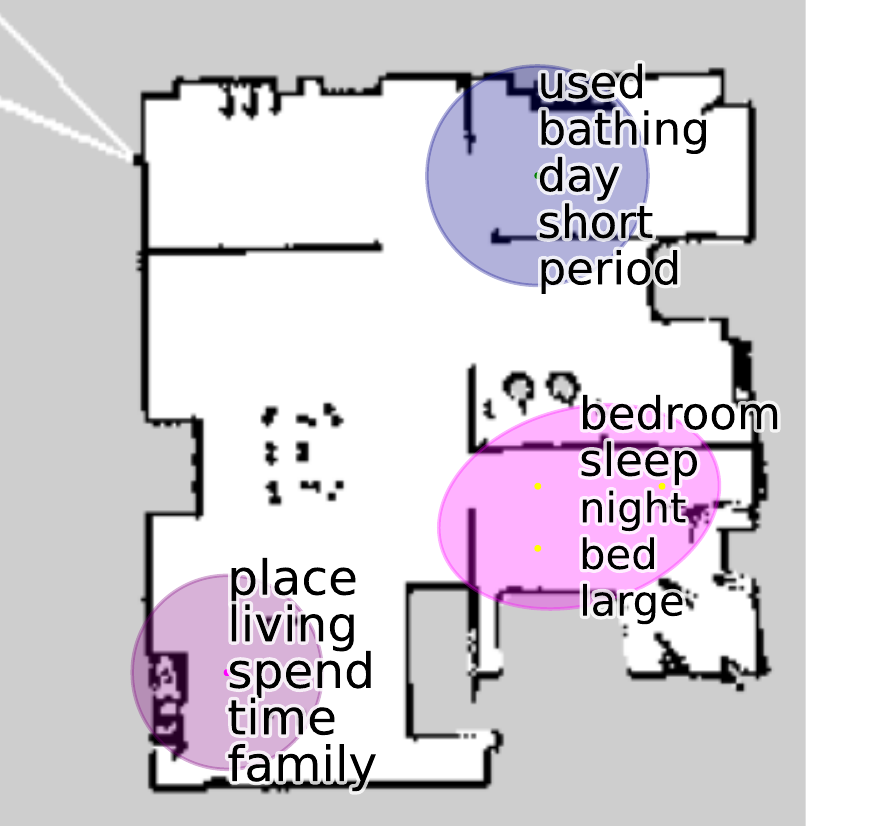}
     \label{fig:result_real_learning_5}
	}
    \subfloat[Step 20]{
    	\includegraphics[width=0.32\linewidth,clip]{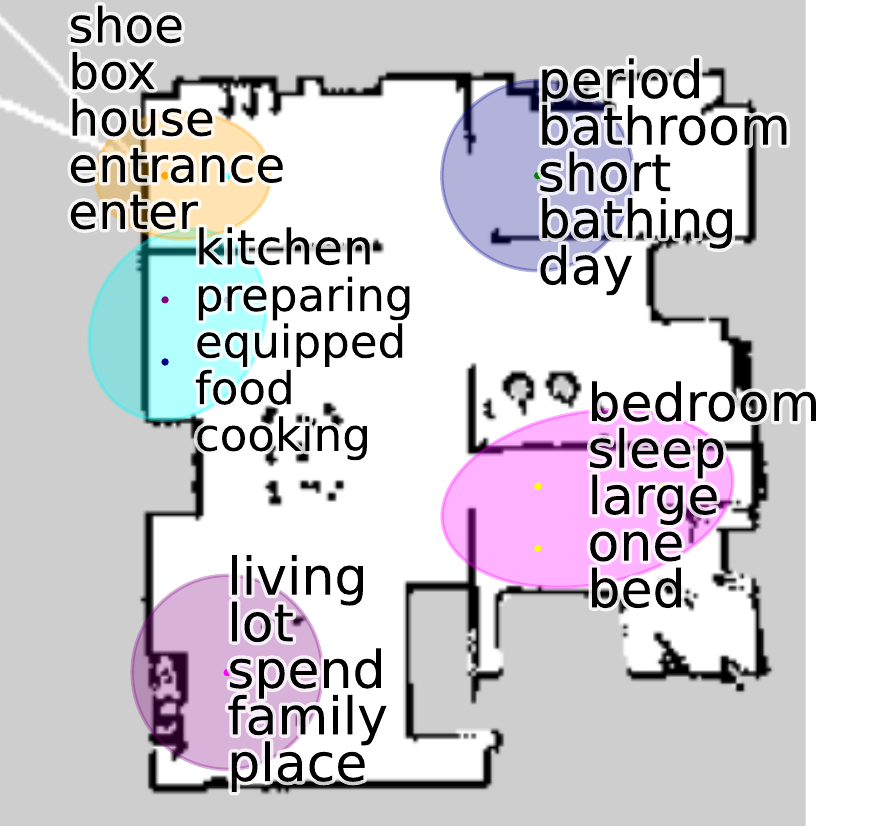}
     \label{fig:result_real_learning_20}
	}
	\\
    \subfloat[Step 25]{
    	\includegraphics[width=0.32\linewidth,clip]{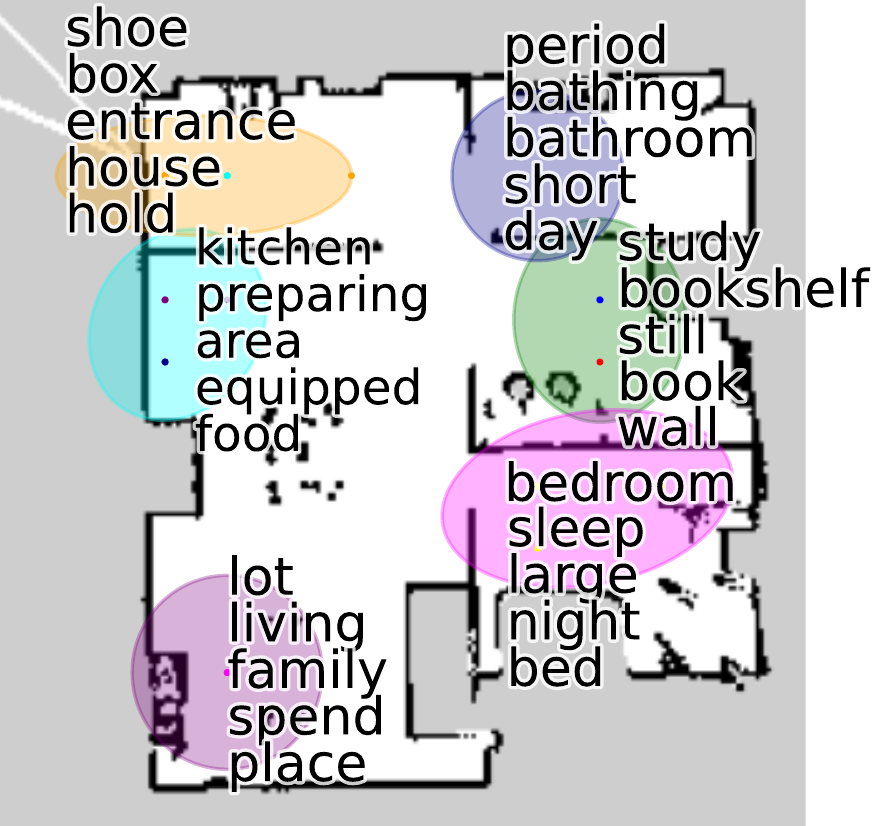}
     \label{fig:result_real_learning_25}
	}
    \subfloat[Step 80]{
    	\includegraphics[width=0.32\linewidth,clip]{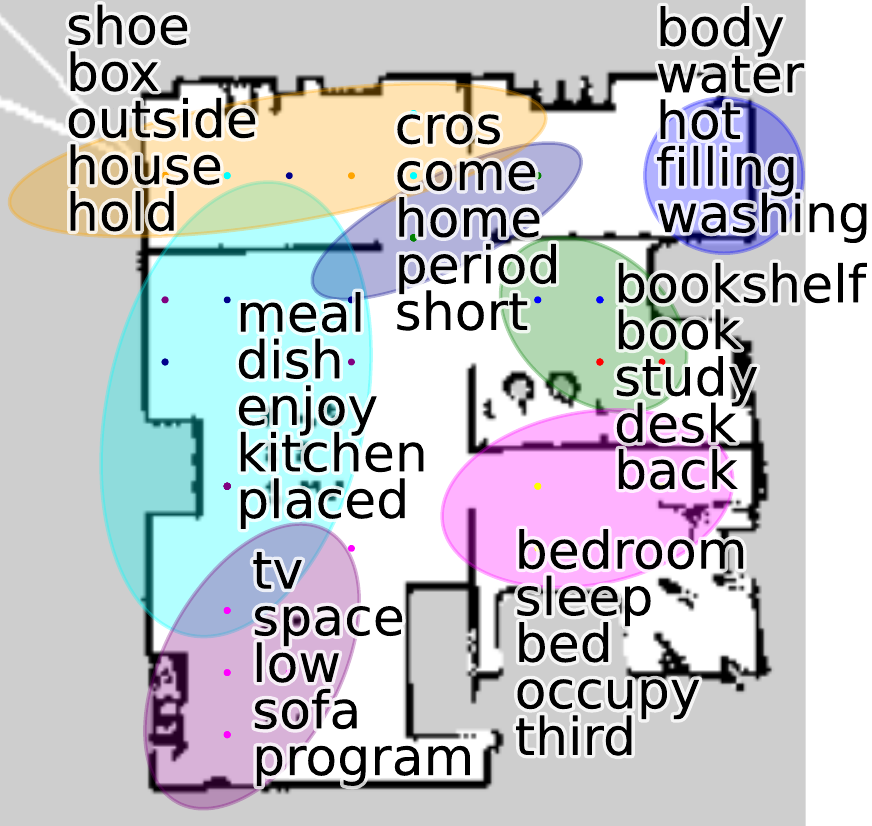}
     \label{fig:result_real_learning_80}
	}
	\caption{
		Process of spatial concept formation by active exploration in several steps (Experiment II).
		Top five words according to the value of pointwise mutual information (PMI) with word $s$ for each position distribution $k$ are obtained, i.e., {$\text{PMI}(S_{n}=s; i_{n}=k) =\log ( p( S_{n}=s \mid i_{n}=k, \Theta ) / p(S_{n}=s \mid \Theta) )$}.
    	PMI is an indicator that weights word distributions such that the words characteristic for a particular position distribution exhibit higher values.
	}
	\label{fig:result_real_learning}
\end{figure}

\begin{figure}[tb]
    \centering
	\includegraphics[width=0.70\linewidth]{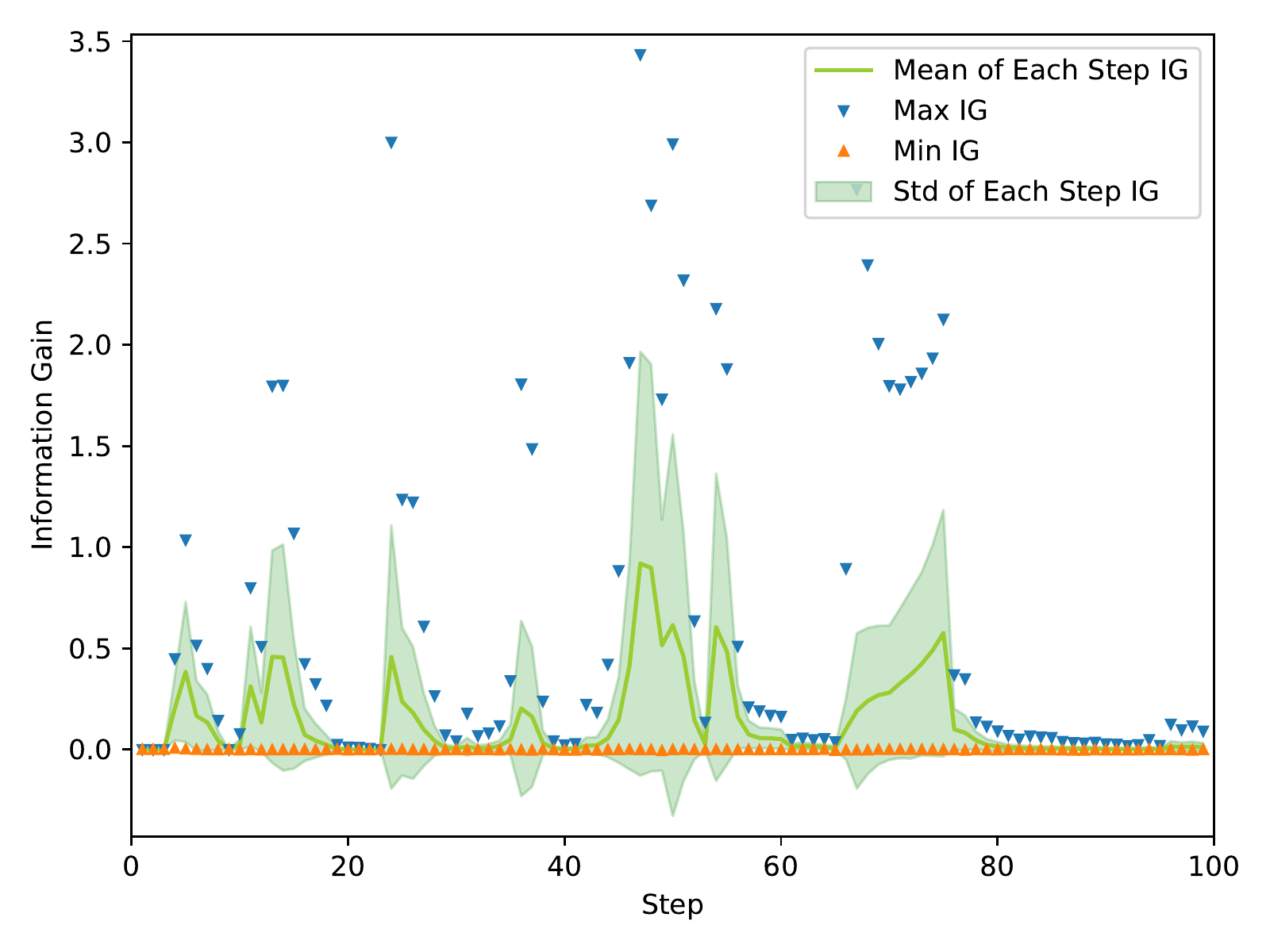}
	\caption{
	    Trends in IG values per step (Experiment II);
        Maximum, minimum, mean, and standard deviation of IG values obtained for all candidate search points at each step are drawn.
    }
	\label{fig:real_ig}
\end{figure}

Figure~\ref{fig:result_real_learning} illustrates the learning process for active exploration in SpCoAE.
For each position distribution, a place-related word corresponded to each location.
In approximately 25 steps, the exploration of all rooms was completed, and the word distribution became stable.
For example, `bedroom' and `sleep' in the lower right distribution and `shoe' and `box' in the upper left distribution are learned as words that characterize places.
Not only are place labels predefined, but the proposed method can also appropriately assign words that are called in the on-site environment and words related to the place in the environment.

Figure~\ref{fig:real_ig} shows the transition of the IG values for each step.
The value of IG forms several peaks, rising and falling, and then converges to almost 0 after step 80.
This suggests that terminating the number of exploration steps according to the IG may be effective.
In other words, this suggests that the robot may be able to determine whether or not to explore new areas based on the value of IG.

Furthermore, when the results in Figures~\ref{fig:result_real_learning} and \ref{fig:real_ig} are combined, the following can be inferred:
In the initial steps, after IG decreased, an increase in IG values was observed as a new distribution was created (e.g., steps 9, 22, and 33).
After Step 33, the number of position distributions does not increase, and each is estimated as an existing distribution.
Even after the number of distributions stops increasing, the area indicated by the position distribution tends to cover more of the entire area as exploration progresses.
As the IG values changed, the word distribution varied.
This may be attributable to the exploration of words that better represent places.
Steps 80--100 exhibit little change.
Because the candidate point is not explored, even in Step 100, a place is sufficiently informative owing to the already learned spatial concept.

In this experiment, the exploration was repeated for candidate points on an advanced trial basis. 
This approach may be particularly effective when dealing with dynamic environments.
In other words, when the robot revisits a previously explored location, and the environment has changed, the entropy of the location's distribution increases, suggesting a greater possibility that the robot will further explore the area.
In conclusion, we demonstrated that SpCoAE works in real-world environments with multiword utterances.

\section{Conclusion}
\label{sec:conclusion}

In this study, we proposed an active exploration algorithm for the spatial concept formation of a robot.
The proposed method achieves AIF by selecting candidate points with the maximum IG and performing online learning from observations obtained at the destination.
Simulation experiments demonstrate the effectiveness of active exploration in terms of more accurate spatial concept formation.
In addition to the IG, introducing the cost of travel distance was effective.
In real-world experiments in which multiple words were provided, exploration proceeded to clarify the word corresponding to a place.

We realized AIF with models in which the position and word observations were used for place categorization.
Further valid categorization using visual images available at the location, as in SpCoSLAM, is possible.
The prospects include the application of active exploration to the PGM of SpCoSLAM.
To this end, we plan to study candidate selection, including image features and integration with active SLAM in unknown and unmapped environments.

In this study, the robot learned spatial concepts from scratch by asking the question, `What kind of place is this?'.
The use of generalized prior knowledge in multiple environments~\cite{Hagiwara2021transfer,Katsumata2020SpCoMapGAN,Katsumata2021slamgan} and large-scale language models~\cite{Brown2020gpt3,Ahn2022palm_saycan} accelerates learning through active exploration.
This is expected to further reduce the number of questions and diversify the question generation.
Integration with such an approach is an emphasized future task.

It is also closely related to AIF based on the FEP~\cite{fep3,friston2017active} and control as an inference~\cite{levine2018reinforcement}.
We will also consider exploring and learning the applications of these principles in the future.
This is expected to enable an integrated generalized formulation that includes not only exploits for learning, but also moves for task execution~\cite{Taniguchi2021}.


\section*{Funding}
This study was supported by 
JST CREST, grant number JPMJCR15E3; 
JST Moonshot R\&D Program, grant number JPMJMS2011; and
JSPS KAKENHI, grant numbers JP20K19900 and JP23K16975, Japan.

\bibliographystyle{tADR}
\bibliography{./SpCoAE}  

\begin{thebibliography}{10}
\providecommand{\url}[1]{\normalfont{#1}}
\providecommand{\urlprefix}{Available from: }
\providecommand{\eprint}[2][]{\url{#2}}

\bibitem{kostavelis2015semantic}
Kostavelis I, Gasteratos A. {Semantic mapping for mobile robotics tasks: A
  survey}. Robotics and Autonomous Systems. 2015;\hspace{0pt}66:86--103.
  \urlprefix\url{http://dx.doi.org/10.1016/j.robot.2014.12.006}.

\bibitem{Garg2020}
Garg S, S{\"{u}}nderhauf N, Dayoub F, Morrison D, Cosgun A, Carneiro G, Wu Q,
  Chin TJ, Reid I, Gould S, Corke P, Milford M. {Semantics for Robotic Mapping,
  Perception and Interaction: A Survey}. Foundations and
  Trends{\textregistered} in Robotics. 2020 jan;\hspace{0pt}8(1–2):1--224.
  \eprint{2101.00443}. \urlprefix\url{http://arxiv.org/abs/2101.00443
  http://dx.doi.org/10.1561/2300000059}.

\bibitem{ataniguchi_IROS2017}
Taniguchi A, Hagiwara Y, Taniguchi T, Inamura T. {Online Spatial Concept and
  Lexical Acquisition with Simultaneous Localization and Mapping}. In:
  Proceedings of the {IEEE}/{RSJ} international conference on intelligent
  robots and systems ({IROS}). 2017. p. 811--818.

\bibitem{ataniguchi2020spcoslam2}
Taniguchi A, Hagiwara Y, Taniguchi T, Inamura T. {Improved and Scalable Online
  Learning of Spatial Concepts and Language Models with Mapping}. Autonomous
  Robots. 2020;\hspace{0pt}44(6):927--946.

\bibitem{friston2017active}
Friston KJ. {Active Inference : A Process Theory}. Neural Computation.
  2017;\hspace{0pt}49:1--49.

\bibitem{friston2021woldmodel}
Friston KJ, Moran RJ, Nagai Y, Taniguchi T, Gomi H, Tenenbaum J. {World model
  learning and inference}. Neural Networks. 2021 dec;\hspace{0pt}144:573--590.

\bibitem{Lanillos2021}
Lanillos P, Meo C, Pezzato C, Meera AA, Baioumy M, Ohata W, Tschantz A,
  Millidge B, Wisse M, Buckley CL, Tani J. {Active Inference in Robotics and
  Artificial Agents: Survey and Challenges}. arXiv. 2021
  dec;\hspace{0pt}\eprint{2112.01871}.
  \urlprefix\url{https://arxiv.org/abs/2112.01871v1
  http://arxiv.org/abs/2112.01871}.

\bibitem{thrun2005probabilistic}
Thrun S, Burgard W, Fox D. {Probabilistic Robotics}. Cambridge, MA: MIT Press.
  2005.

\bibitem{Stachniss2005activeslam}
Stachniss C, Grisetti G, Burgard W. {Information Gain-based Exploration Using
  Rao-Blackwellized Particle Filters.} In: Robotics: Science and systems.
  Vol.~2. 2005. p. 65--72.

\bibitem{Mu2016}
Mu B, Giamou M, Paull L, Agha-Mohammadi AA, Leonard J, How J.
  {Information-based Active SLAM via topological feature graphs}. In: 2016
  {IEEE} 55th conference on decision and control, cdc 2016. Cdc. 2016. p.
  5583--5590. \eprint{1509.08155}.

\bibitem{Placed2022}
Placed JA, Strader J, Carrillo H, Atanasov N, Indelman V, Carlone L,
  Castellanos JA. {A Survey on Active Simultaneous Localization and Mapping:
  State of the Art and New Frontiers}. IEEE Transactions on Robotics. 2023
  jul;\hspace{0pt}:1--20.

\bibitem{Taniguchi2015yoshino}
Taniguchi T, Yoshino R, Takano T. {Multimodal Hierarchical Dirichlet
  Process-Based Active Perception by a Robot}. Frontiers in Neurorobotics. 2018
  may;\hspace{0pt}12(MAY):22. \eprint{1510.00331}.
  \urlprefix\url{https://www.frontiersin.org/article/10.3389/fnbot.2018.00022
  http://arxiv.org/abs/1510.00331
  https://www.frontiersin.org/article/10.3389/fnbot.2018.00022/full}.

\bibitem{Yoshino2021}
Yoshino R, Takano T, Tanaka H, Taniguchi T. {Active Exploration for
  Unsupervised Object Categorization Based on Multimodal Hierarchical Dirichlet
  Process}. In: {IEEE/SICE} international symposium on system integrations
  {(SII)}. 2021.

\bibitem{Veiga2019}
Veiga TS, Silva M, Ventura R, Lima PU. {A hierarchical approach to active
  semantic mapping using probabilistic logic and information reward POMDPs}.
  In: Proceedings international conference on automated planning and scheduling
  {(ICAPS)}. Icaps. 2019. p. 773--781.

\bibitem{Chaplot2021}
Chaplot DS, Dalal M, Gupta S, Malik J, Salakhutdinov R. {SEAL: Self-supervised
  Embodied Active Learning using Exploration and 3D Consistency}. In:
  Proceedings of the advances in neural information processing systems
  {(NeurIPS)}. Vol.~16. 2021. p. 13086--13098. \eprint{2112.01001}.

\bibitem{anderson2018vision}
Anderson P, Wu Q, Teney D, Bruce J, Johnson M, S{\"{u}}nderhauf N, Reid I,
  Gould S, van~den Hengel A. {Vision-and-language navigation: Interpreting
  visually-grounded navigation instructions in real environments}. In:
  Proceedings of the {IEEE} conference on computer vision and pattern
  recognition ({CVPR}). 2018. p. 3674--3683.

\bibitem{Chen2020a}
Chen K, Chen JK, Chuang J, V{\'{a}}zquez M, Savarese S. {Topological Planning
  with Transformers for Vision-and-Language Navigation}. In: Proceedings of the
  {IEEE} computer society conference on computer vision and pattern
  recognition. Nashville, TN, USA. 2021. p. 11271--11281. \eprint{2012.05292}.
  \urlprefix\url{http://arxiv.org/abs/2012.05292}.

\bibitem{park2022vlnsurvey}
Park SM, Kim YG. {Visual language navigation: a survey and open challenges}.
  Artificial Intelligence Review 2022. 2022
  mar;\hspace{0pt}:1--63\urlprefix\url{https://link.springer.com/article/10.1007/s10462-022-10174-9}.

\bibitem{ulker2010sequential}
Ulker Y, G{\"{u}}nsel B, Cemgil T. {Sequential Monte Carlo samplers for
  Dirichlet process mixtures}. In: Proceedings of the international conference
  on artificial intelligence and statistics {(AISTATS)}. 2010. p. 876--883.

\bibitem{asada2009cognitive}
Asada M, Hosoda K, Kuniyoshi Y, Ishiguro H, Inui T, Yoshikawa Y, Ogino M,
  Yoshida C. {Cognitive developmental robotics: a survey}. {IEEE} Transactions
  on Autonomous Mental Development. 2009;\hspace{0pt}1(1):12--34.

\bibitem{cangelosi2015developmental}
Cangelosi A, Schlesinger M. {Developmental Robotics: From Babies to Robots}.
  Intelligent Robotics and Autonomous Agents Series. MIT Press. 2015.
  \urlprefix\url{https://books.google.co.jp/books?id=AbKPoAEACAAJ}.

\bibitem{taniguchi2018TCDSsurvey}
Taniguchi T, Piater J, Worgotter F, Ugur E, Hoffmann M, Jamone L, Nagai T,
  Rosman B, Matsuka T, Iwahashi N, Oztop E. {Symbol Emergence in Cognitive
  Developmental Systems: A Survey}. {IEEE} Transactions on Cognitive and
  Developmental Systems. 2019;\hspace{0pt}11(4):494--516. \eprint{1801.08829}.

\bibitem{Ha2018}
Ha D, Schmidhuber J. {World models}. 2018. {T}ech {R}ep. \eprint{1803.10122}.
  \urlprefix\url{https://worldmodels.github.io}.

\bibitem{fep}
Friston KJ, Kilner J, Harrison L. {A free energy principle for the brain}.
  Journal of Physiology-Paris. 2006 jul;\hspace{0pt}100(1):70--87.

\bibitem{fep1}
Friston KJ, Adams RA, Perrinet L, Breakspear M. {Perceptions as Hypotheses:
  Saccades as Experiments}. Frontiers in Psychology. 2012;\hspace{0pt}3:151.
  \urlprefix\url{https://www.frontiersin.org/article/10.3389/fpsyg.2012.00151}.

\bibitem{fep2}
Friston KJ. {The free-energy principle: a unified brain theory?} Nature reviews
  neuroscience. 2010;\hspace{0pt}11(2):127.

\bibitem{fep3}
Friston KJ, Rigoli F, Ognibene D, Mathys C, Fitzgerald T, Pezzulo G. {Active
  inference and epistemic value}. Cognitive Neuroscience.
  2015;\hspace{0pt}6(4):187--214.

\bibitem{rangel2016lextomap}
Rangel JC, Martinez-Gomez J, Garcia-Varea I, Cazorla M. {LexToMap:
  lexical-based topological mapping}. Advanced Robotics.
  2017;\hspace{0pt}31(5):268--281.

\bibitem{Grinvald2019}
Grinvald M, Furrer F, Novkovic T, Chung JJ, Cadena C, Siegwart R, Nieto J.
  {Volumetric Instance-Aware Semantic Mapping and 3D Object Discovery;
  Volumetric Instance-Aware Semantic Mapping and 3D Object Discovery}. IEEE
  Robotics and Automation Letters. 2019;\hspace{0pt}4(3).
  \urlprefix\url{http://www.ieee.org/publications_standards/publications/rights/index.html}.

\bibitem{Rosinol2021}
Rosinol A, Violette A, Abate M, Hughes N, Chang Y, Shi J, Gupta A, Carlone L.
  {Kimera: From SLAM to spatial perception with 3D dynamic scene graphs}. The
  International Journal of Robotics Research. 2021;\hspace{0pt}40:1510--1546.
  \eprint{2101.06894v1}. \urlprefix\url{www.sagepub.com/
  https://github.com/MIT-}.

\bibitem{Tolman1948a}
Tolman EC. {Cognitive maps in rats and men}. Psychological Review.
  1948;\hspace{0pt}55(4):189--208.

\bibitem{Okeefe1978placecells}
O'keefe J, Nadel L. {The Hippocampus as a Cognitive Map}. Vol.~27. Cambridge
  University Press. 1978.

\bibitem{Taniguchi2021hpf-pgm}
Taniguchi A, Fukawa A, Yamakawa H. {Hippocampal formation-inspired
  probabilistic generative model}. Neural Networks. 2022
  mar;\hspace{0pt}151:317--335. \eprint{2103.07356}.
  \urlprefix\url{http://creativecommons.org/licenses/by/4.0/
  http://arxiv.org/abs/2103.07356}.

\bibitem{ataniguchi2020TidyUpHere}
Taniguchi A, Isobe S, {El Hafi} L, Hagiwara Y, Taniguchi T. {Autonomous
  planning based on spatial concepts to tidy up home environments with service
  robots}. Advanced Robotics. 2021;\hspace{0pt}35(8):471--489.
  \eprint{2002.03671}.

\bibitem{ataniguchi2020spconavi}
Taniguchi A, Hagiwara Y, Taniguchi T, Inamura T. {Spatial Concept-Based
  Navigation with Human Speech Instructions via Probabilistic Inference on
  Bayesian Generative Model}. Advanced Robotics. 2020
  sep;\hspace{0pt}34(19):1213--1228.
  \urlprefix\url{https://www.tandfonline.com/doi/full/10.1080/01691864.2020.1817777}.

\bibitem{Katsumata2020SpCoMapGAN}
Katsumata Y, Taniguchi A, {El Hafi} L, Hagiwara Y, Taniguchi T. {SpCoMapGAN :
  Spatial Concept Formation-based Semantic Mapping with Generative Adversarial
  Networks}. In: Proceedings of the {IEEE}/{RSJ} international conference on
  intelligent robots and systems ({IROS}). Las Vegas, USA: Institute of
  Electrical and Electronics Engineers Inc.. 2020 oct. p. 7927--7934.

\bibitem{Katsumata2021slamgan}
Katsumata Y, Kanechika A, Taniguchi A, {El Hafi} L, Hagiwara Y, Taniguchi T.
  {Map completion from partial observation using the global structure of
  multiple environmental maps}. Advanced Robotics. 2022
  mar;\hspace{0pt}00(00):1--12. \eprint{2103.09071}.
  \urlprefix\url{https://arxiv.org/abs/2103.09071v1}.

\bibitem{Hagiwara2021transfer}
Hagiwara Y, Taguchi K, Ishibushi S, Taniguchi A, Taniguchi T. {Hierarchical
  Bayesian model for the transfer of knowledge on spatial concepts based on
  multimodal information}. Advanced Robotics. 2022
  mar;\hspace{0pt}36(1-2):33--53. \eprint{2103.06442}.
  \urlprefix\url{https://arxiv.org/abs/2103.06442v1}.

\bibitem{Sagara2021}
Sagara R, Taguchi R, Taniguchi A, Taniguchi T, Hattori K, Hoguro M, Umezaki T.
  {Unsupervised lexical acquisition of relative spatial concepts using spoken
  user utterances}. Advanced Robotics. 2021 jun;\hspace{0pt}36(1-2):54--70.
  \eprint{2106.08574}.
  \urlprefix\url{https://www.tandfonline.com/action/journalInformation?journalCode=tadr20
  https://arxiv.org/abs/2106.08574v1}.

\bibitem{wang2019vln}
Wang X, Huang Q, Celikyilmaz A, Gao J, Shen D, Wang YF, Wang WY, Zhang L.
  {Reinforced cross-modal matching and self-supervised imitation learning for
  vision-language navigation}. In: Proceedings of the {IEEE} conference on
  computer vision and pattern recognition ({CVPR}). Vol. 2019-June. 2019. p.
  6622--6631. \eprint{1811.10092}.

\bibitem{montemerlo2003fastslam}
Montemerlo M, Thrun S, Koller D, Wegbreit B, Others. {FastSLAM 2.0: An improved
  particle filtering algorithm for simultaneous localization and mapping that
  provably converges}. In: Proceedings of the international joint conference on
  artificial intelligence {(IJCAI)}. Acapulco, Mexico. 2003. p. 1151--1156.

\bibitem{gridbasedfastslam2005}
Grisetti G, Stachniss C, Burgard W. {Improving grid-based slam with
  Rao-Blackwellized particle filters by adaptive proposals and selective
  resampling}. In: Proceedings of the {IEEE} international conference on
  robotics and automation {(ICRA)}. April. IEEE. 2005. p. 2432--2437.

\bibitem{frontier}
Yamauchi B. {Frontier-based Exploration Using Multiple Robots}. In: Agents.
  1998. p. 47--53. \urlprefix\url{http://doi.acm.org/10.1145/280765.280773}.

\bibitem{Placed2021}
Placed JA, Castellanos JA. {Fast Autonomous Robotic Exploration Using the
  Underlying Graph Structure}. In: Proceedings of the {IEEE}/{RSJ}
  international conference on intelligent robots and systems ({IROS}). Section
  IV. 2021. p. 6649--6656.

\bibitem{Chaplot2020}
Chaplot DS, Gandhi D, Gupta S, Gupta A, Salakhutdinov R. {Learning to Explore
  using Active Neural SLAM}. In: Proceedings of the international conference on
  learning representations {(ICLR)}. 2020. \eprint{arXiv:2004.05155v1}.

\bibitem{Catal2021a}
{\c{C}}atal O, Verbelen T, {Van de Maele} T, Dhoedt B, Safron A. {Robot
  navigation as hierarchical active inference}. Neural Networks. 2021
  oct;\hspace{0pt}142:192--204.
  \urlprefix\url{https://linkinghub.elsevier.com/retrieve/pii/S0893608021002021
  https://doi.org/10.1016/j.neunet.2021.05.010}.

\bibitem{Asgharivaskasi2023}
Asgharivaskasi A, Atanasov N. {Semantic OcTree Mapping and Shannon Mutual
  Information Computation for Robot Exploration}. IEEE Transactions on
  Robotics. 2023;\hspace{0pt}\eprint{2112.04063}.
  \urlprefix\url{https://doi.org/10.1109/TRO.2023.3245986.}

\bibitem{Paul2018}
Paul R, Arkin J, Aksaray D, Roy N, Howard TM. {Efficient grounding of abstract
  spatial concepts for natural language interaction with robot platforms}.
  International Journal of Robotics Research.
  2018;\hspace{0pt}37(10):1269--1299.

\bibitem{Patki2019}
Patki S, Daniele AF, Walter MR, Howard TM. {Inferring compact representations
  for efficient natural language understanding of robot instructions}. In:
  Proceedings of the {IEEE} international conference on robotics and automation
  {(ICRA)}. 2019. p. 6926--6933.

\bibitem{Sugiura2010}
Sugiura K, Iwahashi N, Kashioka H, Nakamura S. {Active learning of confidence
  measure function in robot language acquisition framework}. In: Proceedings of
  the {IEEE}/{RSJ} international conference on intelligent robots and systems
  ({IROS}). 2010. p. 1774--1779.

\bibitem{chen2016experimental}
Chen Y, Bordes JBB, Filliat D. {An experimental comparison between NMF and LDA
  for active cross-situational object-word learning}. Proceedings of the Joint
  {IEEE} International Conference on Development and Learning and Epigenetic
  Robotics (ICDL-EpiRob). 2016;\hspace{0pt}:217--222.

\bibitem{nakamura2011}
Nakamura T, Araki T, Nagai T, Iwahashi N. {Grounding of Word Meanings in Latent
  Dirichlet Allocation-Based Multimodal Concepts}. Advanced Robotics.
  2011;\hspace{0pt}25(17):2189--2206.

\bibitem{Chaplot2020c}
Chaplot DS, Jiang H, Gupta S, Gupta A. {Semantic Curiosity for Active Visual
  Learning}. Lecture Notes in Computer Science (including subseries Lecture
  Notes in Artificial Intelligence and Lecture Notes in Bioinformatics).
  2020;\hspace{0pt}12351 LNCS:309--326. \eprint{2006.09367}.
  \urlprefix\url{http://arxiv.org/abs/2006.09367}.

\bibitem{Georgakis2022iclr}
Georgakis G, Bucher B, Schmeckpeper K, Singh S, Daniilidis K. {Learning to Map
  for Active Semantic Goal Navigation}. Proceedings of the International
  Conference on Learning Representations {(ICLR)}. 2022
  mar;\hspace{0pt}\eprint{2106.15648}.
  \urlprefix\url{http://arxiv.org/abs/2106.15648}.

\bibitem{Ramakrishnan2020a}
Ramakrishnan SK, Jayaraman D, Grauman K. {An Exploration of Embodied Visual
  Exploration}. International Journal of Computer Vision.
  2021;\hspace{0pt}129(5):1616--1649. \eprint{2001.02192}.
  \urlprefix\url{http://arxiv.org/abs/2001.02192}.

\bibitem{Oliver2022}
Oliver G, Lanillos P, Cheng G. {An Empirical Study of Active Inference on a
  Humanoid Robot}. {IEEE} Transactions on Cognitive and Developmental Systems.
  2022;\hspace{0pt}14(2):462--471. \eprint{1906.03022}.
  \urlprefix\url{http://arxiv.org/abs/1906.03022{\%}0Ahttp://dx.doi.org/10.1109/TCDS.2021.3049907}.

\bibitem{Horii2021}
Horii T, Nagai Y. {Active Inference Through Energy Minimization in Multimodal
  Affective Human–Robot Interaction}. Frontiers in Robotics and AI. 2021
  nov;\hspace{0pt}8:361.

\bibitem{doya2007bayesian}
Doya K, Ishii S, Pouget A, Rao RPN. {Bayesian Brain: Probabilistic Approaches
  to Neural Coding.} MIT Press. 2007.

\bibitem{Hafner2019}
Hafner D, Lillicrap T, Fischer I, Villegas R, Ha D, Lee H, Davidson J.
  {Learning latent dynamics for planning from pixels}. Proceedings of the
  International Conference on Machine Learning {(ICML)}.
  2019;\hspace{0pt}2019-June:4528--4547. \eprint{1811.04551}.

\bibitem{Okada2020a}
Okada M, Kosaka N, Taniguchi T. {PlaNet of the Bayesians: Reconsidering and
  Improving Deep Planning Network by Incorporating Bayesian Inference}. In:
  Proceedings of the {IEEE}/{RSJ} international conference on intelligent
  robots and systems ({IROS}). 2020. \eprint{arXiv:2003.00370v1}.

\bibitem{Ball2020}
Ball PJ, Holder JP, Pacchiano A, Choromanski K, Roberts S. {Ready policy one:
  World building through active learning}. In: 37th international conference on
  machine learning, {(ICML)}. Vol. PartF16814. 2020. p. 568--578.
  \eprint{2002.02693}. \urlprefix\url{http://arxiv.org/abs/2002.02693}.

\bibitem{Rao1999}
Rao RPN, Ballard DH. {Predictive coding in the visual cortex: A functional
  interpretation of some extra-classical receptive-field effects}. Nature
  Neuroscience. 1999 jan;\hspace{0pt}2(1):79--87.
  \urlprefix\url{https://pubmed.ncbi.nlm.nih.gov/10195184/}.

\bibitem{murphy2012machine}
Murphy KP. {Machine learning: a probabilistic perspective}. Cambridge, MA: MIT
  Press. 2012.

\bibitem{sethuraman1994constructive}
Sethuraman J. {A constructive definition of Dirichlet priors}. Statistica
  Sinica. 1994;\hspace{0pt}4:639--650.

\bibitem{ref:ishwaran2001gibbs}
Ishwaran H, James LF. {Gibbs sampling methods for stick-breaking priors}.
  Journal of the American Statistical Association.
  2001;\hspace{0pt}96(453):161--173.

\bibitem{fox2011sticky}
Fox EB, Sudderth EB, Jordan MI, Willsky AS, Fox BEB, Sudderth EB, Jordan MI,
  Willsky AS. {A sticky HDP-HMM with application to speaker diarization}. The
  Annals of Applied Statistics. 2011;\hspace{0pt}5(2A):1020--1056.
  \eprint{arXiv:0905.2592v4}.

\bibitem{doucet2000rao}
Doucet A, {De Freitas} N, Murphy K, Russell S. {Rao-Blackwellised particle
  filtering for dynamic Bayesian networks}. In: Proceedings of the 16th
  conference on uncertainty in artificial intelligence. Morgan Kaufmann
  Publishers Inc.. 2000. p. 176--183. \eprint{1301.3853}.

\bibitem{Kong1994}
Kong A, Liu JS, Wong WH. {Sequential Imputations and Bayesian Missing Data
  Problems}. Journal of the American Statistical Association. 1994
  mar;\hspace{0pt}89(425):278.

\bibitem{Liu1996}
Liu JS. {Metropolized independent sampling with comparisons to rejection
  sampling and importance sampling}. Statistics and Computing.
  1996;\hspace{0pt}6(2):113--119.
  \urlprefix\url{http://130.203.136.95/viewdoc/summary?doi=10.1.1.31.6718}.

\bibitem{HSR2019}
Yamamoto T, Terada K, Ochiai A, Saito F, Asahara Y, Murase K. {Development of
  Human Support Robot as the research platform of a domestic mobile
  manipulator}. ROBOMECH Journal. 2019;\hspace{0pt}6(1):4.
  \urlprefix\url{https://doi.org/10.1186/s40648-019-0132-3}.

\bibitem{inamura2021sigverse}
Inamura T, Mizuchi Y. {SIGVerse: A Cloud-Based VR Platform for Research on
  Multimodal Human-Robot Interaction}. Frontiers in Robotics and AI. 2021
  may;\hspace{0pt}8:158.

\bibitem{ROS}
Quigley M, Conley K, Gerkey BP, Faust J, Foote T, Leibs J, Wheeler R, Ng AY.
  {ROS: an open-source Robot Operating System}. In: Proceedings of the icra
  workshop on open source software. Kobe, Japan. 2009.

\bibitem{ataniguchi2022spcotmhp}
Taniguchi A, Ito S, Taniguchi T. {Spatial Concept-based Topometric Semantic
  Mapping for Hierarchical Path-planning from Speech Instructions}. arXiv. 2022
  mar;\hspace{0pt}\eprint{2203.10820}.
  \urlprefix\url{https://arxiv.org/abs/2203.10820v1
  http://arxiv.org/abs/2203.10820}.

\bibitem{Brown2020gpt3}
Brown TB, Mann B, Ryder N, Subbiah M, Kaplan J, Dhariwal P, Neelakantan A,
  Shyam P, Sastry G, Askell A, Agarwal S, Herbert-Voss A, Krueger G, Henighan
  T, Child R, Ramesh A, Ziegler DM, Wu J, Winter C, Hesse C, Chen M, Sigler E,
  Litwin M, Gray S, Chess B, Clark J, Berner C, McCandlish S, Radford A,
  Sutskever I, Amodei D. {Language models are few-shot learners}. In: Advances
  in neural information processing systems. Vol. 2020-December. Neural
  information processing systems foundation. 2020 may. p. 1877--1901.
  \eprint{2005.14165}. \urlprefix\url{https://commoncrawl.org/the-data/
  https://arxiv.org/abs/2005.14165v4}.

\bibitem{Ahn2022palm_saycan}
Ahn M, Brohan A, Brown N, Chebotar Y, Cortes O, David B, Finn C, Fu C,
  Gopalakrishnan K, Hausman K, Herzog A, Ho D, Hsu J, Ibarz J, Ichter B, Irpan
  A, Jang E, Ruano RJ, Jeffrey K, Jesmonth S, Joshi NJ, Julian R, Kalashnikov
  D, Kuang Y, Lee KH, Levine S, Lu Y, Luu L, Parada C, Pastor P, Quiambao J,
  Rao K, Rettinghouse J, Reyes D, Sermanet P, Sievers N, Tan C, Toshev A,
  Vanhoucke V, Xia F, Xiao T, Xu P, Xu S, Yan M, Zeng A. {Do As I Can, Not As I
  Say: Grounding Language in Robotic Affordances}. In: arxiv preprint. 2022
  apr. \eprint{2204.01691}. \urlprefix\url{http://arxiv.org/abs/2204.01691}.

\bibitem{levine2018reinforcement}
Levine S. {Reinforcement Learning and Control as Probabilistic Inference:
  Tutorial and Review}. arXiv preprint arXiv:180500909. 2018;\hspace{0pt}.

\bibitem{Taniguchi2021}
Taniguchi T, {El Hafi} L, Hagiwara Y, Taniguchi A, Shimada N, Nishiura T.
  {Semiotically adaptive cognition: toward the realization of remotely-operated
  service robots for the new normal symbiotic society}. Advanced Robotics.
  2021;\hspace{0pt}35(11):664--674.
  \urlprefix\url{https://www.tandfonline.com/doi/abs/10.1080/01691864.2021.1928552}.

\end{thebibliography}

\clearpage
\appendix
\pagenumbering{roman} 

\section{Formula derivation in online learning of spatial concepts}
\label{apdx:online}

\subsection{Derivation of target distribution $P_n$}
\label{apdx:online:P_n}

The target distribution $P_n$ is transformed as follows:
\begin{align}
P_{n} &= p(C_{1:n}, i_{1:n} \mid x_{1:n}, S_{1:n}, h) \\
&= p(C_{n}, i_{n} \mid C_{1:n-1}, i_{1:n-1}, x_{1:n}, S_{1:n}, h)p(C_{1:n-1}, i_{1:n-1} \mid x_{1:n}, S_{1:n}, h) \\
&\overset{\text{Bayes' rule}}{=} p(C_{n}, i_{n} \mid C_{1:n-1}, i_{1:n-1}, x_{1:n}, S_{1:n}, h) \nonumber \\ 
&\qquad \frac{p(x_{n} \mid  C_{1:n-1}, i_{1:n-1}, x_{1:n-1}, S_{1:n}, h)p(C_{1:n-1}, i_{1:n-1} \mid x_{1:n-1}, S_{1:n}, h)}{p(x_{n} \mid x_{1:n-1}, S_{1:n}, h)} \\
&\overset{\text{Markovian}}{\propto} p(C_{n}, i_{_{\tau}n} \mid C_{1:n-1}, i_{1:n-1}, x_{1:n}, S_{1:n}, h)p(x_{n} \mid C_{1:n-1}, i_{1:n-1}, x_{1:n-1}, h) \nonumber \\
&\qquad p(C_{1:n-1}, i_{1:n-1} \mid x_{1:n-1}, S_{1:n}, h) \\
&\overset{\text{Bayes' rule}}{=} p(C_{n}, i_{n} \mid C_{1:n-1}, i_{1:n-1}, x_{1:n}, S_{1:n}, h)p(x_{n} \mid C_{1:n-1}, i_{1:n-1}, x_{1:n-1}, h) \nonumber \\
&\qquad \frac{p(S_{n} \mid C_{1:n-1}, i_{1:n-1}, x_{1:n-1}, S_{1:n-1}, h)p(C_{1:n-1}, i_{1:n-1} \mid x_{1:n-1}, S_{1:n-1}, h)}{p(S_{n} \mid x_{1:n-1}, S_{1:n-1}, h)} \\
&\overset{\text{Markovian}}{\propto} p(C_{n}, i_{n} \mid C_{1:n-1}, i_{1:n-1}, x_{1:n}, S_{1:n}, h)p(x_{n} \mid C_{1:n-1}, i_{1:n-1}, x_{1:n-1}, h) \nonumber \\
&\qquad p(S_{n} \mid S_{1:n-1}, C_{1:n-1}, \alpha, \beta)P_{n-1}.
\end{align}

\subsection{Derivation of proposal distribution $q_n$}
\label{apdx:online:q_n}

The proposal distribution $q_{n}$ is transformed as follows:
\begin{align}
q_{n} &= p(C_{n}, i_{n} \mid C_{1:n-1}, i_{1:n-1}, x_{1:n}, S_{1:n}, h) \\
&\overset{\text{Bayes' rule}}{=} \frac{p(x_{1:n} \mid C_{1:n}, i_{1:n}, S_{1:n}, h)p(C_{n}, i_{n} \mid C_{1:n-1}, i_{1:n-1}, S_{1:n}, h)}{p(x_{1:n} \mid C_{1:n-1}, i_{1:n-1}, S_{1:n}, h)} \\
&\overset{\text{Markovian}}{\propto} p(x_{1:n} \mid i_{1:n}, h)p(C_{n}, i_{n} \mid C_{1:n-1}, i_{1:n-1}, S_{1:n}, h) \\
&\overset{\text{Bayes' rule}}{=} p(x_{1:n} \mid i_{1:n}, h)\frac{p(S_{1:n} \mid C_{1:n}, i_{1:n}, h)p(C_{n}, i_{n} \mid C_{1:n-1}, i_{1:n-1}, h)}{p(S_{1:n} \mid C_{1:n-1}, i_{1:n-1}, h)} \\
&\overset{\text{Markovian}}{\propto} p(x_{1:n} \mid i_{1:n}, h)p(S_{1:n} \mid C_{1:n}, \beta)p(C_{n}, i_{n} \mid C_{1:n-1}, i_{1:n-1}, \alpha, \gamma) \\
&{\propto} p(x_{n} \mid x_{1:n-1},i_{1:n}, h)p(S_{n} \mid S_{1:n-1}, C_{1:n}, \beta) \nonumber \\
&\qquad p(C_{n}, i_{n} \mid C_{1:n-1}, i_{1:n-1}, \alpha, \gamma).
\end{align}

\subsection{Simultaneous sampling of $C_{n}$ and $i_{n}$}
\label{subsec:C_n_tau,i_n_tau}

Here, each term of Eq.~(\ref{eq:qn}).
The probability distribution for $x_{n}$ is 
\begin{align}
p(x_{n} \mid x_{1:n-1},i_{1:n}, h) 
= \operatorname{St}\left( x_{n} \mid m_k, \frac{V_k(\kappa_k +1)}{\kappa_k(\nu_k-d+1)}, \nu_k-d+1 \right)
\label{eq:x_n}
\end{align}
where $\operatorname{St}()$ is the Student's t-distribution.
The parameters of the posterior distribution of Eq.~(\ref{eq:x_n}) is calculated as follows:
\begin{align}
\bar{x}_k &= \frac{1}{t_{n-1}^{(k)}}\sum_{x_j\in x_k}x_j\\
m_k &= \frac{t_{n-1}^{(k)}\bar{x}_k+\kappa_0m_0}{t_{n-1}^{(k)}+\kappa_0} \\
\kappa_k &=  \,t_{n-1}^{(k)}+\kappa_0 \\
\nu_k &= \nu_0+t_{n-1}^{(k)} \\
V_k &= V_0 + \sum_{x_j\in x_k} x_jx_j^T + \kappa_0m_0m_0^T - \kappa_km_km_k^T
\end{align}
where $t_{n-1}^{(k)}$ denotes the index of the data point assigned to the $k$-th position distribution when the $n-1$-th data point is obtained.

The probability distribution for the bag-of-words $S_{n}$ is as follows:
\begin{align}
&p(S_{n} \mid S_{1:n-1}, C_{1:n}, \beta)  \nonumber \\
&= \prod_{g=1}^{G} p(S_{n,g} \mid S_{1:n-1}, C_{1:n-1}, C_{n}=l, \beta) \\
&= \prod_{g=1}^{G} \left( \frac{t_{n-1}^{(l,g)}+\beta}{\sum_{g^{\prime}=1}^G (t_{n-1}^{(l,g^{\prime})}+\beta)} \right)^{S_{n,g}}
\end{align}
where $t_{n-1}^{(l,g)}$ is the count number of the $g$-th word $s_g$ of the $l$-th word distribution in $S_{1:n-1}$ when the $n-1$-th data point is obtained.
In addition, $G$ is the number of word types (the number of dimensions of the word distribution), and $B_t$ is the number of words in an utterance.

The probability distribution with respect to $i_{n}$ and $C_{n}$ is 
\begin{align}
&p(C_{n}, i_{n} \mid C_{1:n-1}, i_{1:n-1}, \alpha, \gamma)  \nonumber \\
&= p(i_{n}=k \mid C_{n}=l, C_{1:n-1}, i_{1:n-1}, \gamma)p(C_{n}=l \mid C_{1:n-1}, \alpha) \\
&= \,\frac{t_{n-1}^{(l,k)}+\gamma/K}{t_{n-1}^{(l)}+\gamma}\frac{t_{n-1}^{(l)}+\alpha/L}{t_{n-1}+\alpha}
\end{align}
where $t_{n-1}^{(l)}$ denotes the count number of data assigned to the $l$-th spatial concept when the $n-1$-th data point is obtained.
In addition, $t_{n-1}^{(l,k)}$ denotes the count number of data assigned to the $l$-th spatial concept and $k$-th position distribution when the $n-1$-th data are obtained.

From the above, Eq.~(\ref{eq:qn}) can be expressed as follows:
\begin{align}
q_{n} 
&\propto \operatorname{St}\left( x_{n} \mid m_k, \frac{V_k(\kappa_k +1)}{\kappa_k(\nu_k-d+1)}, \nu_k-d+1 \right) \nonumber \\
&\qquad \prod_{g=1}^{G} \left( \frac{t_{n-1}^{(l,g)}+\beta}{\sum_{g^{\prime}=1}^G (t_{n-1}^{(l,g^{\prime})}+\beta)} \right)^{S_{n,g}}
\frac{t_{n-1}^{(l,k)}+\gamma/K}{t_{n-1}^{(l)}+\gamma}  \frac{t_{n-1}^{(l)}+\alpha/L}{t_{n-1}+\alpha}.
\label{eq:ci}
\end{align}

\subsection{Posterior distribution of model parameters $\Theta$}
\label{subsec:Theta}

The first term $p(\Theta \mid C_{1:n}, i_{1:n}, x_{1:n}, S_{1:n}, h)$ in Eq.~(\ref{eq:learning_algorithm}) is the posterior distribution of the parameter $\Theta$, as follows:
\begin{align}
&p(\Theta \mid C_{1:n}, i_{1:n}, x_{1:n}, S_{1:n}, h)  \nonumber \\
&= \prod_{k=1}^{K} \mathcal{NIW}(\mu_k, \Sigma_k \mid m_{k}, \kappa_{k}, V_{k}, \nu_{k}) 
\left[  \prod_{l=1}^{L} 
\operatorname{Dir}(W_l \mid ( t_{n}^{(l,g)}+ \beta )  ) \operatorname{Dir}(\phi_l \mid ( t_{n}^{(l,k)} + \gamma/K ) ) 
\right]  \nonumber \\ 
&\qquad \operatorname{Dir}(\pi \mid ( t_{n}^{(l)} + \alpha/L ) ) 
\label{eq:param}
\end{align}
where $\mathcal{NIW}()$ denotes a Gaussian inverse Wishart distribution.
In practice, this set of parameters $\Theta$ is obtained for each particle sampled using the RBPF.
Each term in Eq.~(\ref{eq:param}) can be computed independently for the corresponding parameters and categories.
In addition, each term is a posterior distribution for each parameter, which can be computed by the conjugacy of the prior distribution and the likelihood function.
To obtain specific estimates for each parameter, the expected value of each posterior distribution was obtained.
Refer to \cite{murphy2012machine} for the specific formulas for the above probability distributions.

\section{Formula derivation in SpCoAE}
\label{apdx:active}

\subsection{Derivation of IG for destination selection}
\label{apdx:active:IG_formula}

This section describes the transformation details of Eqs.~(\ref{eq:KL_min})--(\ref{eq:argmax_IG}).
A similar derivation has been employed in MHDP-based active perception/learning methods~\cite{Taniguchi2015yoshino,Yoshino2021}.
The index $a^{\ast}$ of the candidate position for exploration observed at the next destination can be obtained using the expected value of the KL divergence as follows:
\begin{align}
a^{\ast}&=\argmin_{a} \mathbb{E}_{X_{\{1:N\} \backslash n_0}  \mid X_{n_0}}\left[ D_{\rm{KL}}[p(Z \mid X_{1:N}),p(Z \mid X_{n_0\cup a})]\right] \\
&=\argmin_{a}\sum_{X_{\{1:N\} \backslash n_0}} \sum_Z \left[p(X_{\{1:N\} \backslash n_0} \mid X_{n_0}) p(Z \mid X_{1:N})\log\frac{p(Z \mid X_{1:N})}{p(Z \mid X_a,X_{n_0})} \right] .
\label{eq:log}
\end{align}

In Eq.~(\ref{eq:log}), the numerator of the log function $p(Z \mid X_{1:N})$ is eliminated because it does not depend on $a$, and the denominator and numerator are inverted as follows:
\begin{align}
\text{Eq.~(\ref{eq:log})} &=\argmax_{a}\sum_{X_{\{1:N\} \backslash n_0}} \sum_Z \left[p(X_{\{1:N\} \backslash n_0} \mid X_{n_0})p(Z \mid X_{1:N})\log p(Z \mid X_a,X_{n_0})\right] \\
&=\argmax_{a}\sum_{X_{\{1:N\} \backslash n_0}} \sum_Z \left[p(Z,X_{\{1:N\} \backslash n_0} \mid X_{n_0})\log p(Z \mid X_a,X_{n_0})\right] . \label{eq:tikan}
\end{align}

In Eq.~(\ref{eq:tikan}), the replacement of $\{1:N\} \backslash n_0$ with $a$ can be expressed as follows:
\begin{align}
\text{Eq.~(\ref{eq:tikan})} =\argmax_{a}& \sum_{X_a} \sum_Z \left[p(Z,X_{a} \mid X_{n_0})\log p(Z \mid X_a,X_{n_0})\right] \\
=\argmax_{a}& \sum_{X_a} \sum_Z \left[p(Z,X_{a} \mid X_{n_0})\log p(Z \mid X_a,X_{n_0})\right] - \underbrace{\sum_Z \left[p(Z \mid X_{n_0})\log p(Z \mid X_{n_0})\right]}_{\mbox{Add the constant term}} \\
{=}\argmax_{a}& \left[\sum_{X_a} \sum_Z \left[p(X_{a} \mid X_{n_0})p(Z \mid X_{n_0},X_{a})\log p(Z \mid X_a,X_{n_0})\right] \right. \nonumber \\
&-\sum_{X_a} \left. \sum_Z  \left[p(X_{a} \mid X_{n_0})p(Z \mid X_{n_0},X_{a})\log p(Z \mid X_{n_0})\right]\right] \\
{=}\argmax_{a}& \sum_{X_a} p(X_{a} \mid X_{n_0}) \underbrace{\sum_Z p(Z \mid X_{n_0},X_{a}) \frac{\log p(Z \mid X_a,X_{n_0})}{p(Z \mid X_{n_0})}}_{\mbox{KL divergence}}  \label{eq:tokl}\\
{=}\argmax_{a}& \sum_{X_a} p(X_{a} \mid X_{n_0}) D_{\rm{KL}}[p(Z \mid X_a,X_{n_0})  \| p(Z \mid X_{n_0})]\label{eq:kl_definition}\\
=\argmax_{a} & \, \mathbb{E}_{X_a \mid X_{n_0}}\left[ D_{\rm{KL}}[p(Z \mid X_{{n_0}\cup a}) \| p(Z \mid X_{n_0})]\right]  \label{eq:kl_to_ig}\\
=\argmax_{a} & \, \mbox{IG}(Z;X_a \mid X_{n_0}). \label{eq:ig_definition}
\end{align}
Based on information-theoretic definitions, the deformation from Eq.~(\ref{eq:tokl}) to  Eq.~(\ref{eq:kl_definition}) uses the definition of the KL divergence and deformation from Eq.~(\ref{eq:kl_to_ig}) to  Eq.~(\ref{eq:ig_definition}) defines the IG.
IG is known as mutual information~\cite{Placed2022}.

\subsection{Another derivation: entropy-based IG derivation}
\label{apdx:active:another}

The algorithm for deriving the IG calculation using an entropy-based procedure similar to the active SLAM~\cite{Stachniss2005activeslam} is shown in Algorithm~\ref{alg:ActiveSpCoA_2}.
The derivation of the algorithm is presented in this section. 

The IG function $\mbox{IG}(Z;X_a \mid X_{n_0})$ in terms of the entropy difference is as follows:
\begin{align}
\mbox{IG}(Z;X_a \mid X_{n_0})=&H(p(Z \mid X_{n_0}))-H(p(Z \mid X_{n_0\cup a})) \\
=&H(p(\Theta,C_{1:n},i_{1:n} \mid x_{n_0},S_{n_0},h)) \nonumber \\
& -H(p(\Theta,C_{1:n},i_{1:n} \mid x_{n_0\cup a},S_{n_0\cup a},h))
\end{align}

Thus, Eq.~(\ref{eq:argmax_IG}) becomes 
\begin{align}
\text{Eq.~(\ref{eq:argmax_IG})}&=\argmax_{a}[\underbrace{H(p(Z \mid X_{n_0}))}_{\mbox{Constant}}-H(p(Z \mid X_{n_0\cup a}))] \\
&=\argmin_{a}H(p(Z \mid X_{n_0}, X_a)).
\end{align}

Thus, $H(p(Z \mid X_{n_0}, X_a))$ can be expressed using the definition of conditional entropy as follows:
\begin{align}
H(p(Z \mid X_{n_0}, X_a))&=\sum_{X_a}[p(X_a \mid X_{n_0})H(p(Z \mid X_{n_0},X_a))] \\
&\approx\sum_{j=1}^{J} H(p(Z \mid X_{n_0},X_a^{[j]})),\quad X_a^{[j]} \sim p(X_a \mid X_{n_0}).
\label{eq:conditonal_entropy}
\end{align}

Therefore, from Eqs.~(\ref{eq:conditonal_entropy}) and (\ref{eq:pred_dist_X}), $H(p(Z \mid X_{n_0}, X_a))$ is expressed as:
\begin{align}
H(p(Z \mid X_{n_0}, X_a)) %
\approx& \sum_{r=1}^{R} \left[ \omega_{n_0}^{[r]}\sum_{j=1}^{J}  H(p(Z \mid X_{n_0},X_a^{[j]})) \right], \nonumber \\
&\quad Z^{[r]} \sim q(Z \mid X_{n_0}), \quad X_a^{[j]} \sim p(X_a \mid Z^{[r]}, X_{n_0}). 
\end{align}

In addition, $H(p(Z \mid X_{n_0},X_a^{[j]}))$ is 
\begin{align}
H(p(Z \mid X_{n_0},X_a^{[j]}))&=H(p(\Theta, C_{1:n}, i_{1:n} \mid x_{n_0}, S_{n_0}, x_a^{[j]}, S_a^{[j]}, h)) \\
&\approx H(p(C_{n_0\cup a}, i_{n_0\cup a} \mid x_{n_0\cup a}^{[j]}, S_{n_0\cup a}^{[j]}, h)) \nonumber \\
&\quad + \sum_{r^{\prime}=1}^{R} \omega_{n_0\cup a}^{[r^{\prime},j]}H(p(\Theta \mid C_{n_0\cup a}^{[r^{\prime},j]}, i_{n_0\cup a}^{[r^{\prime},j]}, x_{n_0\cup a}^{[j]}, S_{n_0\cup a}^{[j]}, h)).
\end{align}
This suggests that entropy is computed after turning the particle filters using the pseudo-observation $X_a^{[j]} = \{ x_a^{[j]}, S_a^{[j]} \}$.

\begin{algorithm}[tb]
    \caption{
        Entropy-based active exploration algorithm for spatial concept formation. 
    } 
    \label{alg:ActiveSpCoA_2}          
    \begin{algorithmic}[1]
            \State{Initialize $n_{0}=\emptyset$, $ Z_{0}=\emptyset$, $X_{n_{0}}=\emptyset$}
            \For{$\tau=1$ to $\mathcal{T}$} \Comment{The number of action trials} 
                \For{\textbf{all} $ \{ x_a \in \mathbb{R}^{D} \mid \text{free-space in a map} \}$} \Comment{The number of candidate positions} 
                    \For{$r=1$ to $R$} \Comment{The number of particles}
                        \For{$j=1$ to $J$} \Comment{The number of samples} 
                            \State{$X_{a}^{[r,j]} \sim p(X_{a} \mid Z_{\tau-1}^{[r]}, X_{n_0})$}
                            \State{$\bar{Z}_{\tau}^{[r,j]} = \textsc{Online\_Learning}(Z_{\tau-1}, \{ X_{n_0},X_a^{[r,j]} \} )$} 
                        \EndFor
                    \EndFor
                    \State{$\displaystyle {H}_{a} = \sum_{r=1}^{R} \sum_{j=1}^{J} H(p(Z \mid X_{n_0},X_a^{[r,j]}))$} 
                \EndFor                    
                \State{$a^{\ast} = \argmin_{a} {{H}}_{a} $} 
                \Comment{Select position $x_{a^{\ast}}$}
                \State{$n_{0} = n_{0} \cup \{ a^{\ast} \}$ }
                \State{Move to the position $x_{a^{\ast}}$, and observe words $S_{a^{\ast}}$}
                \Comment{Observe $X_{n_{0}}$}
                \State{$Z_{\tau} = \textsc{Online\_Learning}(Z_{\tau-1},X_{n_0})$} 
            \EndFor
    \end{algorithmic}
\end{algorithm}

\subsection{Exploration of active inference algorithms in two derivations}

Algorithm~\ref{alg:ActiveSpCoA} is based on the IG-based active perception methods~\cite{Taniguchi2015yoshino,Yoshino2021}, and Algorithm~\ref{alg:ActiveSpCoA_2} is based on the method proposed for active SLAM~\cite{Stachniss2005activeslam}.
These two algorithms are identical, except for lines 9 and 11 in Algorithm~\ref{alg:ActiveSpCoA} and lines 7, 10, and 12 in Algorithm~\ref{alg:ActiveSpCoA_2}.
In Algorithm~\ref{alg:ActiveSpCoA}, the calculation results in line 7 can be reused in line 10.
Compared with Algorithm~\ref{alg:ActiveSpCoA_2}, which requires entropy calculations, Algorithm~\ref{alg:ActiveSpCoA} is easier to implement and faster.
Although Algorithm~\ref{alg:ActiveSpCoA_2} appears to require more processing in line 8, the results of the processing in line 7 can be directly reused in the sampling and weight calculations of the online learning algorithm in line 8.
In addition, if $X_{a^{\ast}} = \{ x_{a^{\ast}}, S_{a^{\ast}} \} = X_{a}^{[r,j]}$, the estimated value, $\bar{Z}_{\tau}^{[r,j]}$, can be reused as $Z_{\tau}$.

This study demonstrated the theoretical connectivity between SpCoAE and active SLAM.
In the future, Algorithm~\ref{alg:ActiveSpCoA_2} will be more compatible with the theoretical integration of SpCoAE and active SLAM.
However, because the two algorithms were derived from the same IG, an integrated reinterpretation based on Algorithm~\ref{alg:ActiveSpCoA} was possible.
This allows for an almost intact extension of the algorithm and implementation.

Comparing the proposed and conventional methods, the number of samples $K$ in the Monte Carlo approximation~\cite{Taniguchi2015yoshino,Yoshino2021} corresponds to $R$ when $J=1$ in the proposed algorithm.
Therefore, the proposed algorithm is more generalizable.
Furthermore, as described in Section~\ref{sec:proposed:ig}, the Monte Carlo approximation is described in detail~\cite{Taniguchi2015yoshino,Yoshino2021}.
However, the proposed method can use the estimation results of the existing particle filters. 
In addition, whereas~\cite{Taniguchi2015yoshino,Yoshino2021} assumed batch learning for Gibbs sampling for each exploration, the proposed method is computationally more efficient than existing methods because of online learning using particle filters.

\section{Preliminary experiment: Comparison of accuracy and computation time for spatial concept learning}
\label{apdx:pre_exp}

The purpose of this experiment was to compare the accuracies of spatial concept learning and computation time when the number of particles and pseudo-observations in SpCoAE were varied.

\subsection{Condition}
\label{subsec:exp1_condition}

Table~\ref{tab:exp_1_condition} lists the number of particles $R$ and the number of pseudo-observations $J$ for each experimental pattern.
Ten trials were conducted for each experimental pattern.
Environment 5 was used as the experimental environment in this study.
The conditions related to the candidate points were identical to those in Experiment I.
A total of 103 candidates were available for observation.
In all the experimental patterns,
The hyperparameters were set as $\alpha=1.0,\,\beta=0.01,\,\gamma=0.1,\,m_0=[0.0,0.0],\,\kappa_0=0.001,\,V_0={\rm diag}(1.5,1.5),\,\nu_0=4.0$.
The upper limits for the number of categories were set as $K=10$ and $L=10$.
The simulations were performed on a laptop with the following specifications: Intel Core i7-7820HK CPU, 32 GB of DDR4 memory, and Nvidia GeForce GTX 1080 GPU.
The robot was operated using the ROS kinetic Kame running on Ubuntu 16.04 LTS.

The evaluation metrics were the same as those in Experiment I.
In addition, the computation time required for each step and the computation time required for all steps were evaluated.
The computation time excludes the time required to obtain observations or travel time.

\begin{table}[tb]
    \tbl{
        Experimental condition pattern.
    }{
    \label{tab:exp_1_condition}
    \begin{tabular}{ccc} \hline
        \textbf{Pattern} & \textbf{Number of particles $(R)$} & \textbf{Number of pseudo-observations $(J)$} \\ \hline 
        A & 1000 & 1\\ 
        B & 100  & 1\\ 
        C & 10   & 1\\ 
        D & 1    & 1\\ 
        E & 1000 & 10\\ 
        F & 100  & 10\\ 
        G & 10   & 10\\ 
        H & 1    & 10\\ 
        K & 1500 & 10\\ 
        L & 500  & 10\\ 
        \hline
    \end{tabular}
    }
\end{table}

\subsection{Results}
\label{subsec:exp1_result}

\begin{figure}[tb]
	\centering
    \includegraphics[width=0.97\linewidth]{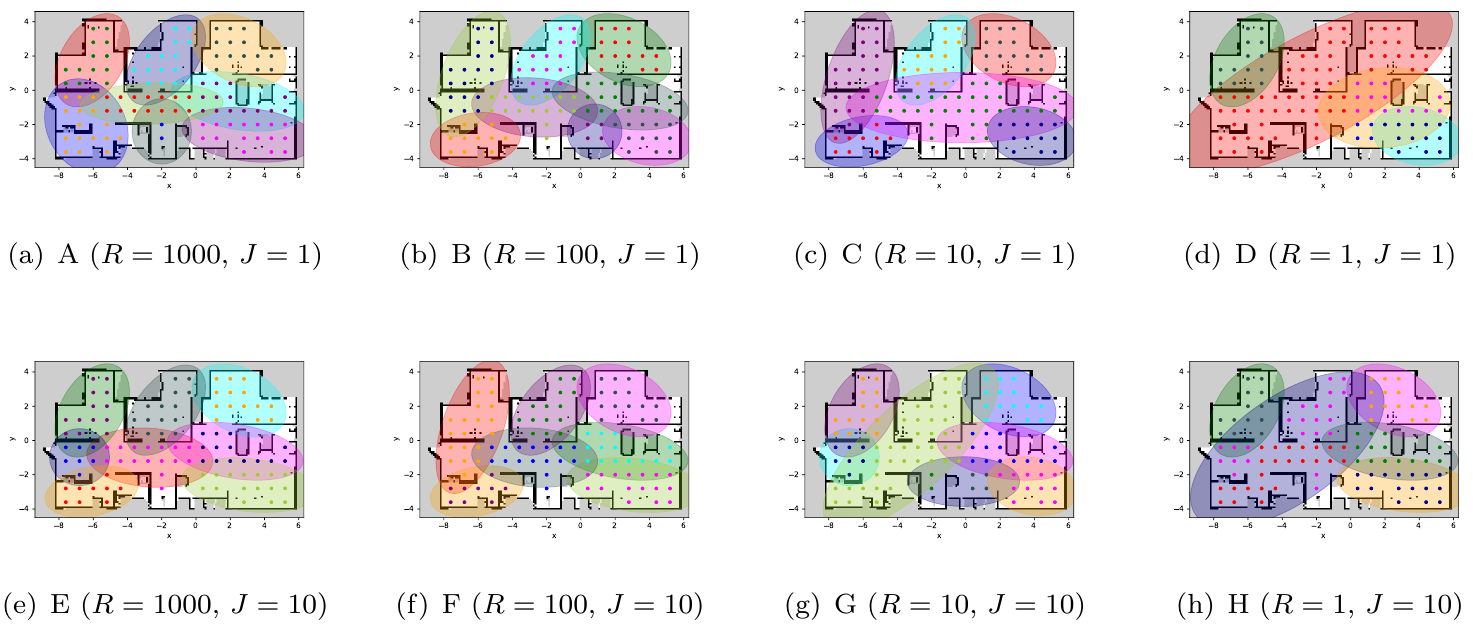}
    \caption{
        Examples of learning results for position distributions and index~$C_{n}$ of spatial concept for each coordinate in experimental patterns A--H.
    }
    \label{fig:exp_1_result_A_H_map_C}
\end{figure}

Figure~\ref{fig:exp_1_result_A_H_map_C} shows examples of the learning results of the spatial concept index $C$ and the position distribution for each position coordinate in experimental patterns A--H.
Additionally, the color changed randomly for each trial because the category numbers assigned to index~$C$ and index~$i$ changed randomly.
For each experimental pattern, the mean and standard deviation of each evaluation value for all ten trials are shown in the graphs below.

\begin{figure}[tb]
	\centering
    \includegraphics[width=\linewidth]{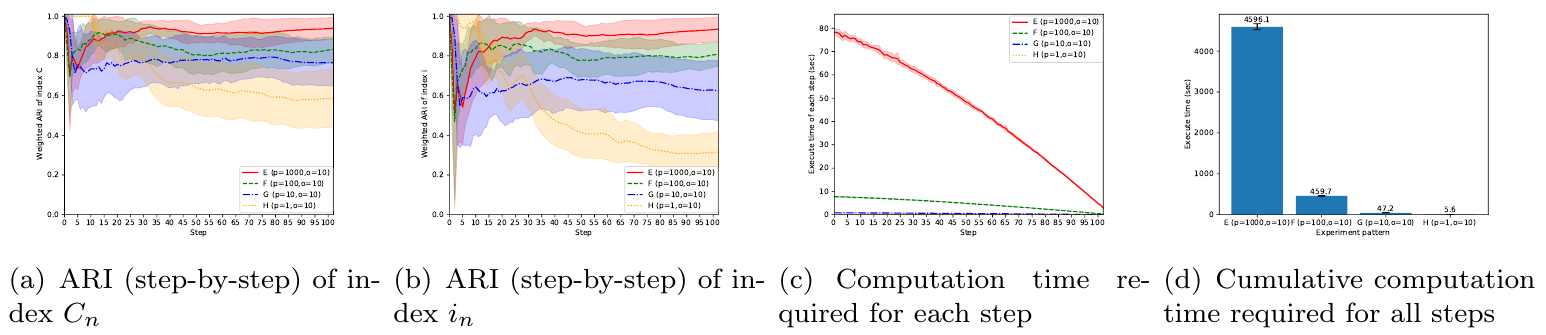}
    \caption{Weighted ARI values and computation time in experimental patterns E--H.}
    \label{fig:exp_1_result_E_H}
\end{figure}

Figure~\ref{fig:exp_1_result_E_H} shows the results of comparing experimental patterns E--H with $R=(1000,\,100,\,10,\,1)$ number of particles and $J=10$ pseudo-observations.
In terms of the weighted ARI for indices $C$ and $i$ in experimental patterns A--H, Figures~\ref{fig:exp_1_result_E_H}(a) and \ref{fig:exp_1_result_E_H}(b) indicate that the higher the number of particles, the higher the value from approximately step 30 to the final step.
In addition, an example of the learning results (Figure~\ref{fig:exp_1_result_A_H_map_C}) indicates that the higher the number of particles, the closer the result is to an ideal position distribution.
Figures~\ref{fig:exp_1_result_E_H}(c) and \ref{fig:exp_1_result_E_H}(d) show that the larger the number of particles, the longer is the computation time required.
These results confirm a tradeoff between the number of particles and the accuracy of spatial concept learning.

\begin{figure}[tb]
	\centering
    \includegraphics[width=\linewidth]{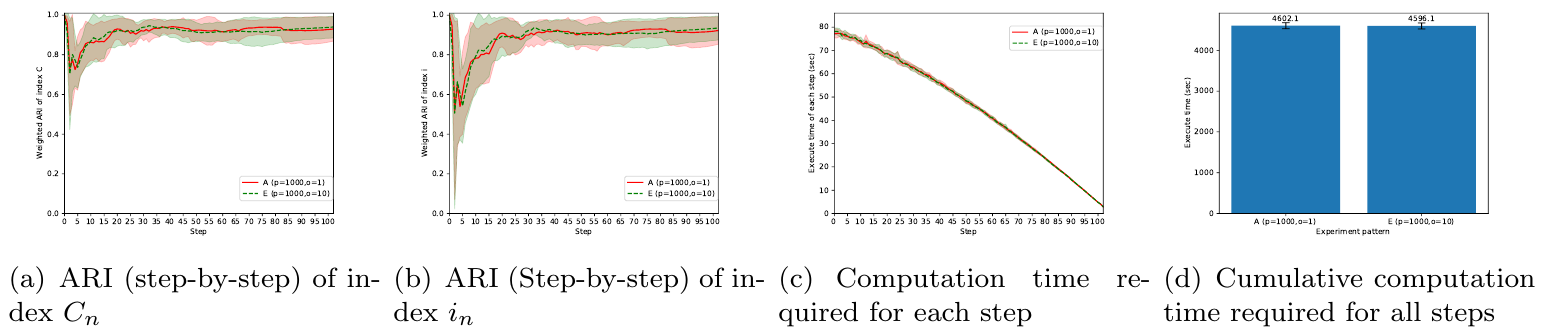}
    \caption{Weighted ARI values and computation time in experimental patterns A and E.}
    \label{fig:exp_1_result_AE}
\end{figure}

\begin{figure}[tb]
	\centering
    \includegraphics[width=\linewidth]{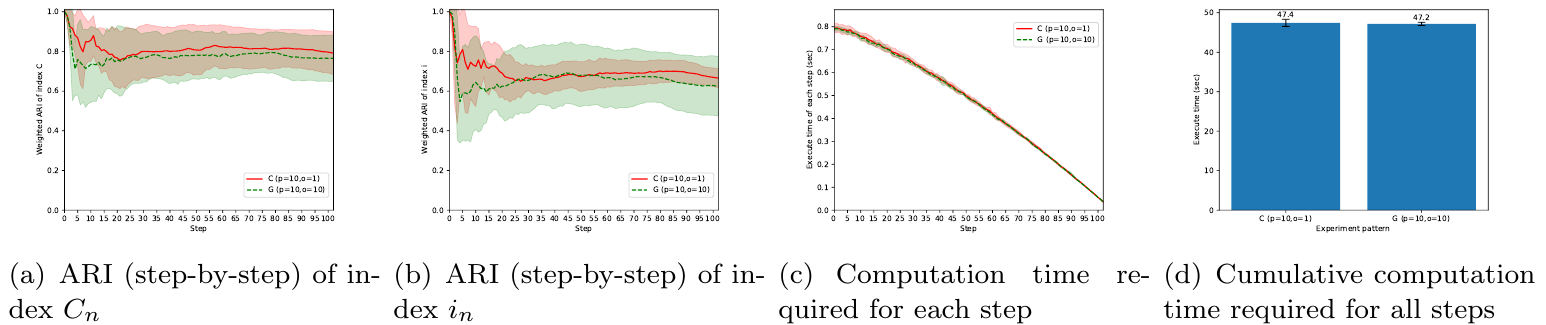}
    \caption{Weighted ARI values and computation times in experimental patterns C and G.}
    \label{fig:exp_1_result_CG}
\end{figure}

The comparison of experimental results A and E with the number of particles $R=1000$ and the number of pseudo-observations $J=(1,\,10)$ are shown in Figure~\ref{fig:exp_1_result_AE}.
The results of comparing experimental patterns C and G with the number of particles $R=10$ and $J=(1,\,10)$ are shown in Figure~\ref{fig:exp_1_result_CG}.
These results show that, regardless of the number of particles, the difference in the number of pseudo-observations has little effect on the accuracy and computation time of spatial concept learning.
This can be attributed to the present condition setting in which one word is assigned.
The effect of the number of pseudo-observations is more pronounced when there are many-to-many correspondences between places and words.

\begin{figure}[tb]
	\centering
    \includegraphics[width=\linewidth]{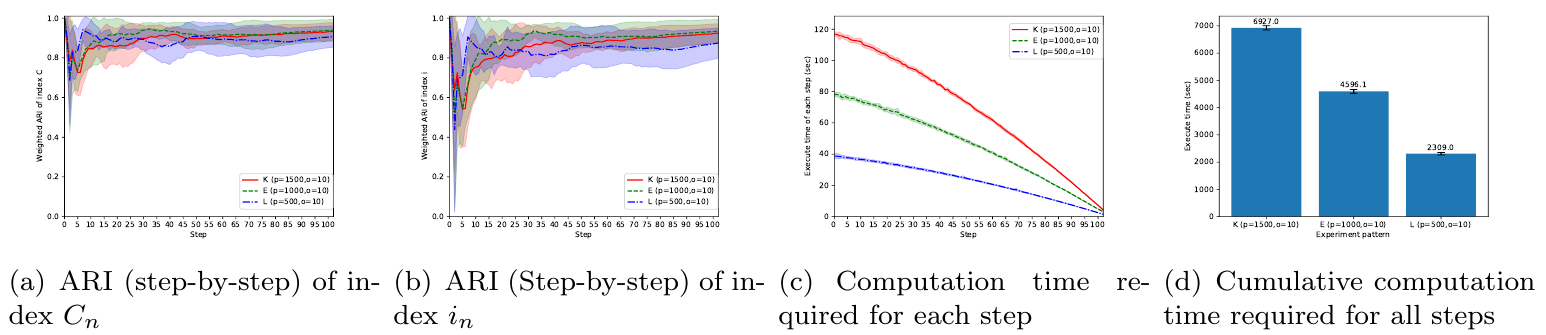}
    \caption{Weighted ARI values and computation time in experimental patterns of K, E, L.}
    \label{fig:exp_1_result_KEL}
\end{figure}

Finally, we show the results of the experiments when the number of particles was $1500, 500$, based on the experimental pattern with the number of particles $R = 1000$, which was the most accurate in spatial concept learning in previous experiments.
The results of comparing experimental patterns K, E, and L with $R=(1500,\,1000,\,500)$ number of particles and $J=10$ pseudo-observations are shown in Figure~\ref{fig:exp_1_result_KEL}.
The weighted ARI for experimental patterns K, E, and L, as shown in Figure~\ref{fig:exp_1_result_KEL}(a) and \ref{fig:exp_1_result_KEL}(b) indicate that the ARI of 500 particles is slightly lower than that of 1500 or 1000 particles.
Combining Figures~\ref{fig:exp_1_result_KEL}(c) and \ref{fig:exp_1_result_KEL}(d), when 1500 particles were used, the computation time was approximately 1.5 times longer than when 1000 particles were used, even though the accuracies of spatial concept learning were similar.
Therefore, the number of particles, $R=1000$, and the number of pseudo-observations, $J=10$ were determined in Experiment I.

The experiment was performed on a single-core CPU to verify the computational speed.
In reality, it is possible to parallelize and implement each sample during Monte Carlo sampling using particle filters.
Therefore, in an actual operation, parallelization allows for accelerated calculations.

\clearpage

\section{Results of Experiment I in ten environments}
\label{apdx:exp1:results}

\begin{figure}[tb]
	\centering
   	\subfloat[Environment 1]{
       	\includegraphics[width=0.17\linewidth]{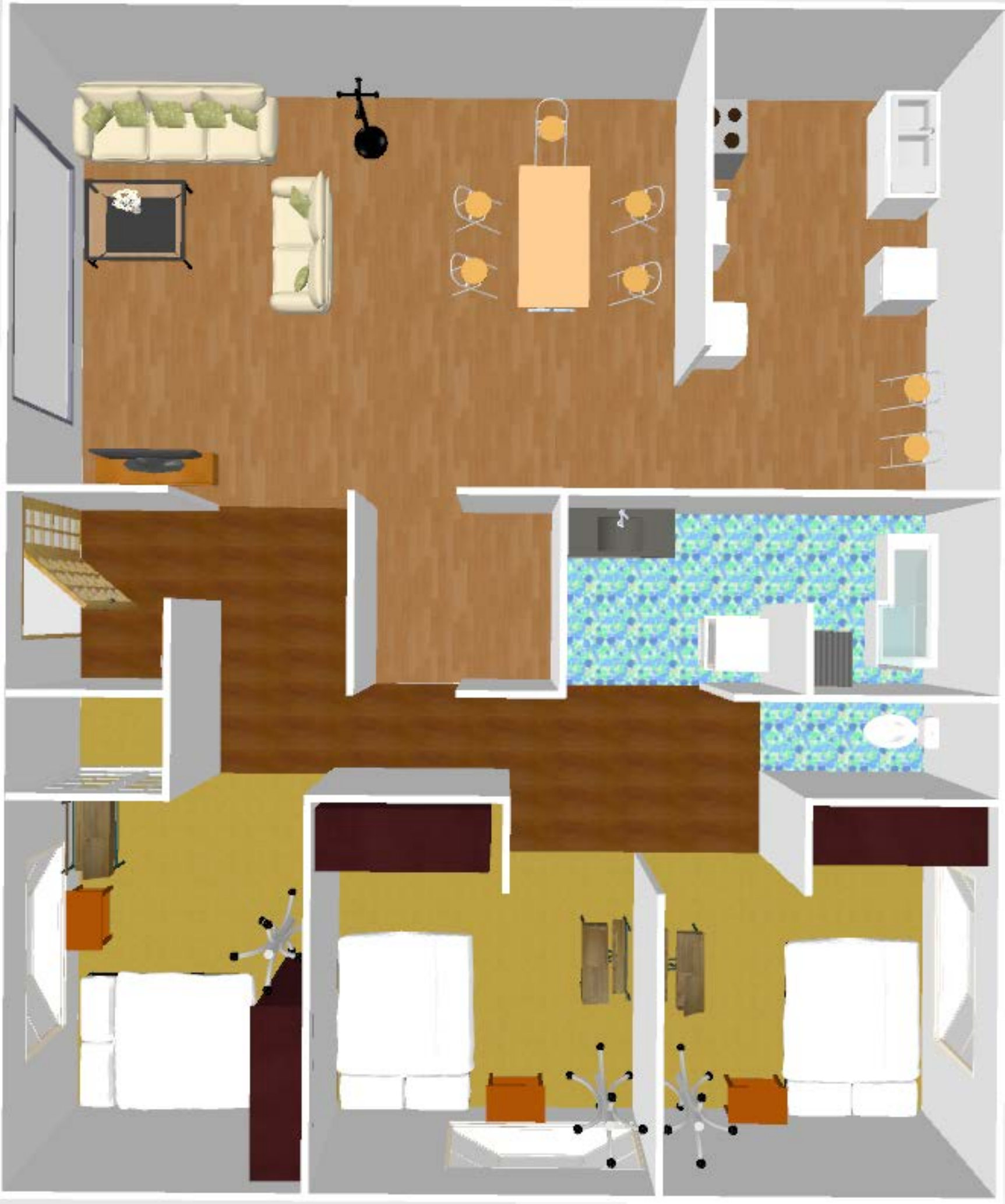}
        \label{fig:exp_2_environment_sim_1}}
   	\subfloat[Environment 2]{
       	\includegraphics[width=0.15\linewidth]{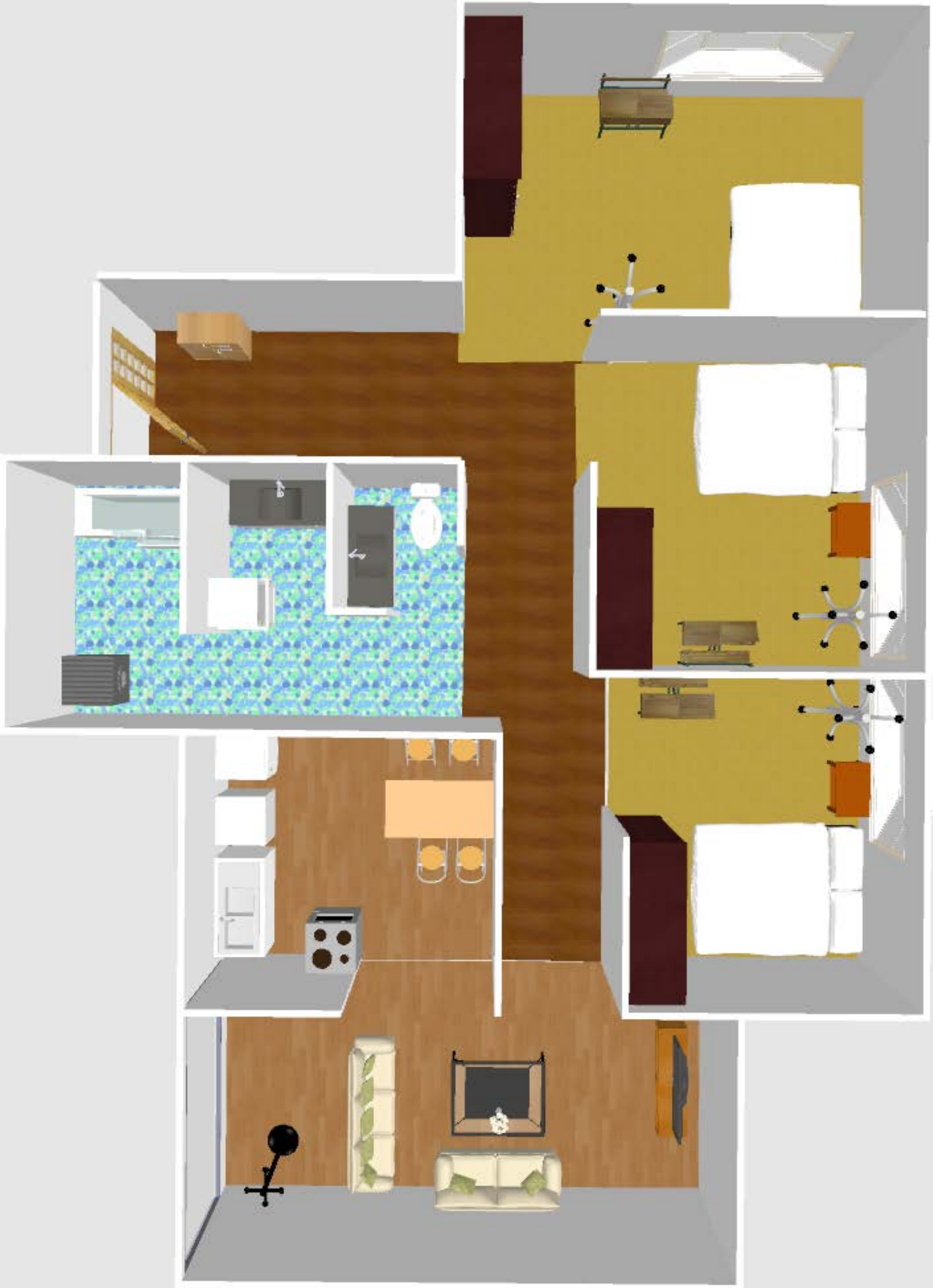}
        \label{fig:exp_2_environment_sim_2}}
   	\subfloat[Environment 3]{
       	\includegraphics[width=0.18\linewidth]{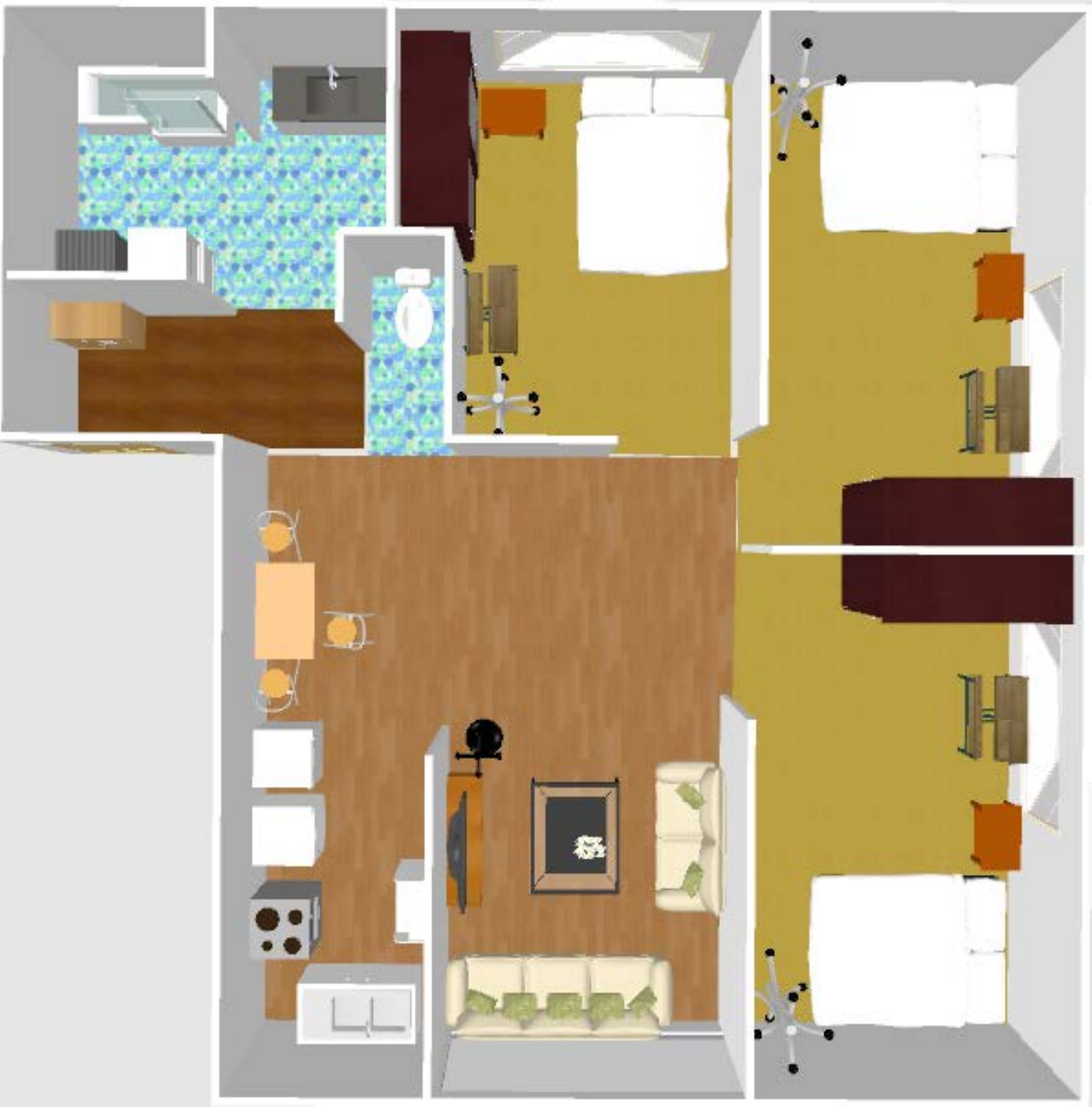}
        \label{fig:exp_2_environment_sim_3}}
   	\subfloat[Environment 4]{
       	\includegraphics[width=0.18\linewidth]{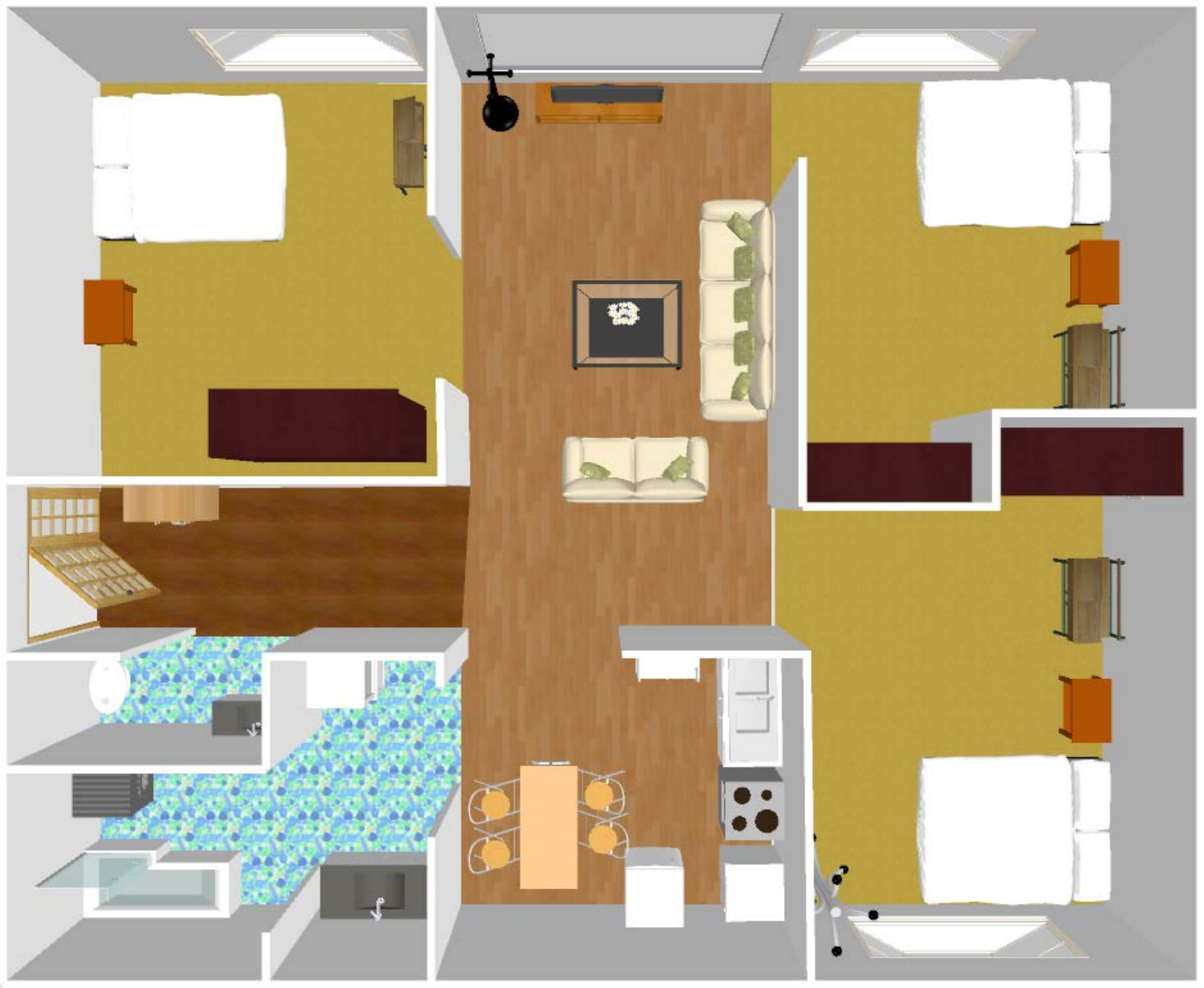}
        \label{fig:exp_2_environment_sim_4}}
   	\subfloat[Environment 5]{
       	\includegraphics[width=0.21\linewidth]{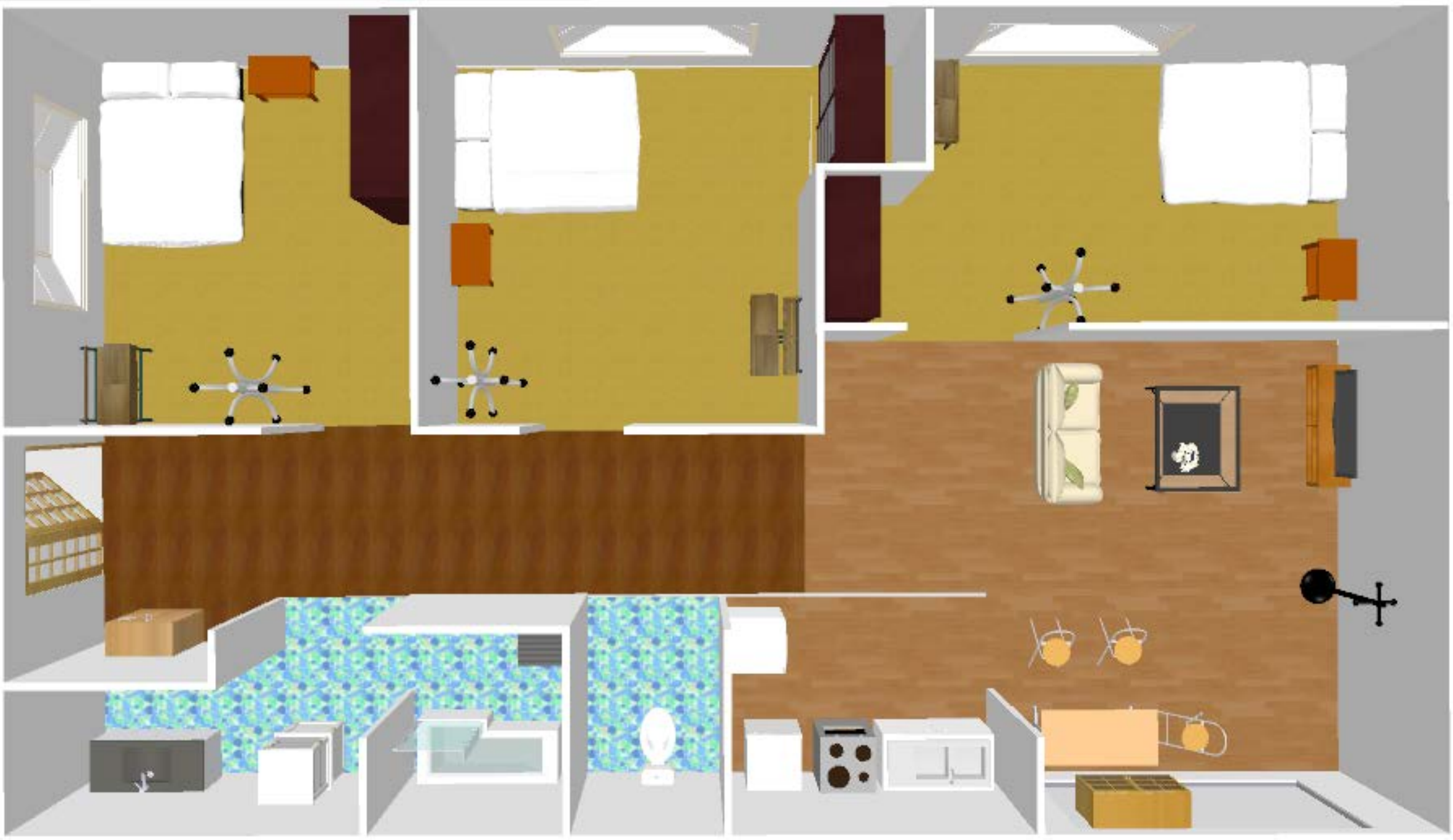}
        \label{fig:exp_2_environment_sim_5}} 
        \\
   	\subfloat[Environment 6]{
       	\includegraphics[width=0.17\linewidth]{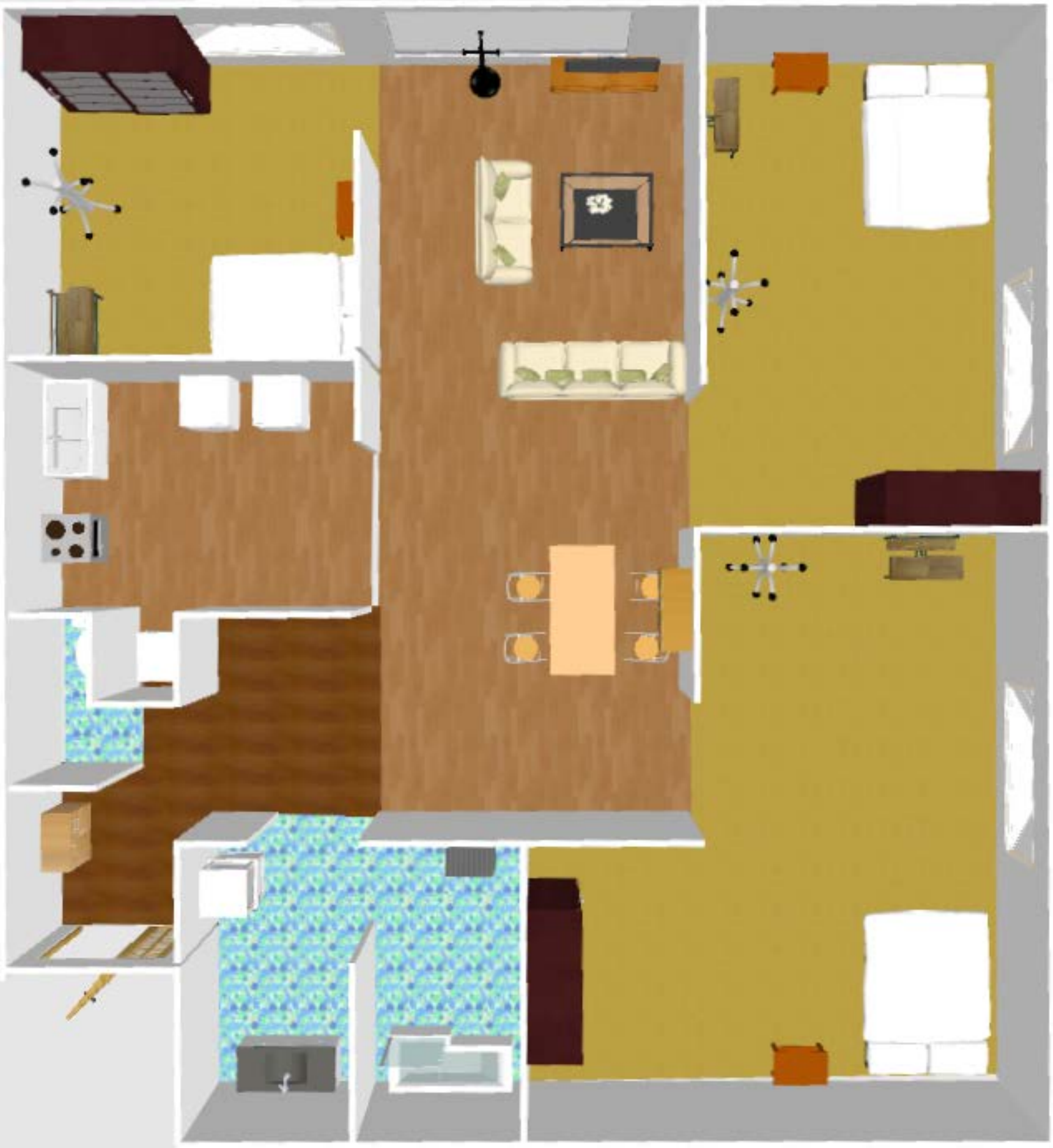}
        \label{fig:exp_2_environment_sim_6}}
   	\subfloat[Environment 7]{
       	\includegraphics[width=0.18\linewidth]{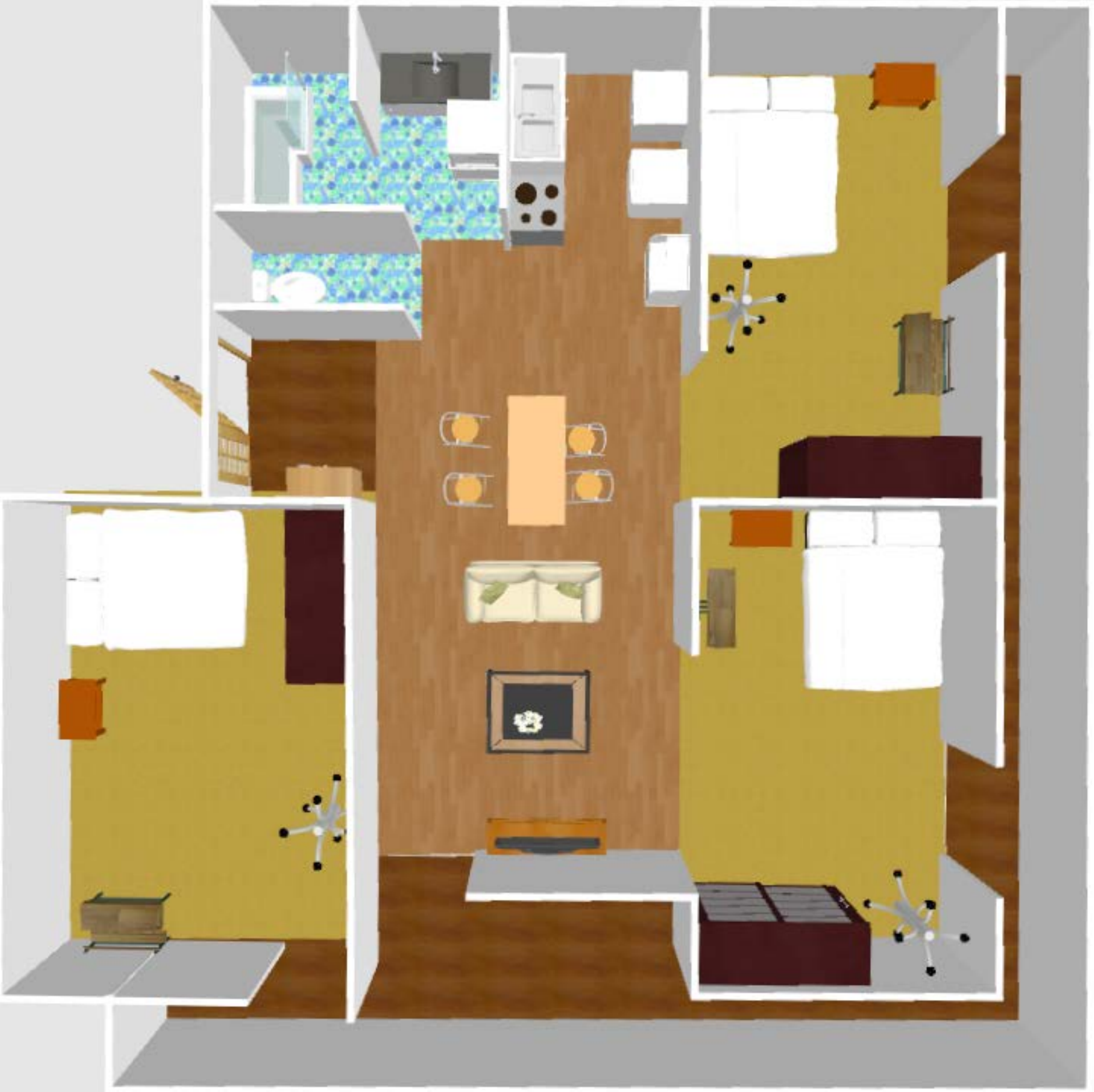}
        \label{fig:exp_2_environment_sim_7}}
   	\subfloat[Environment 8]{
       	\includegraphics[width=0.20\linewidth]{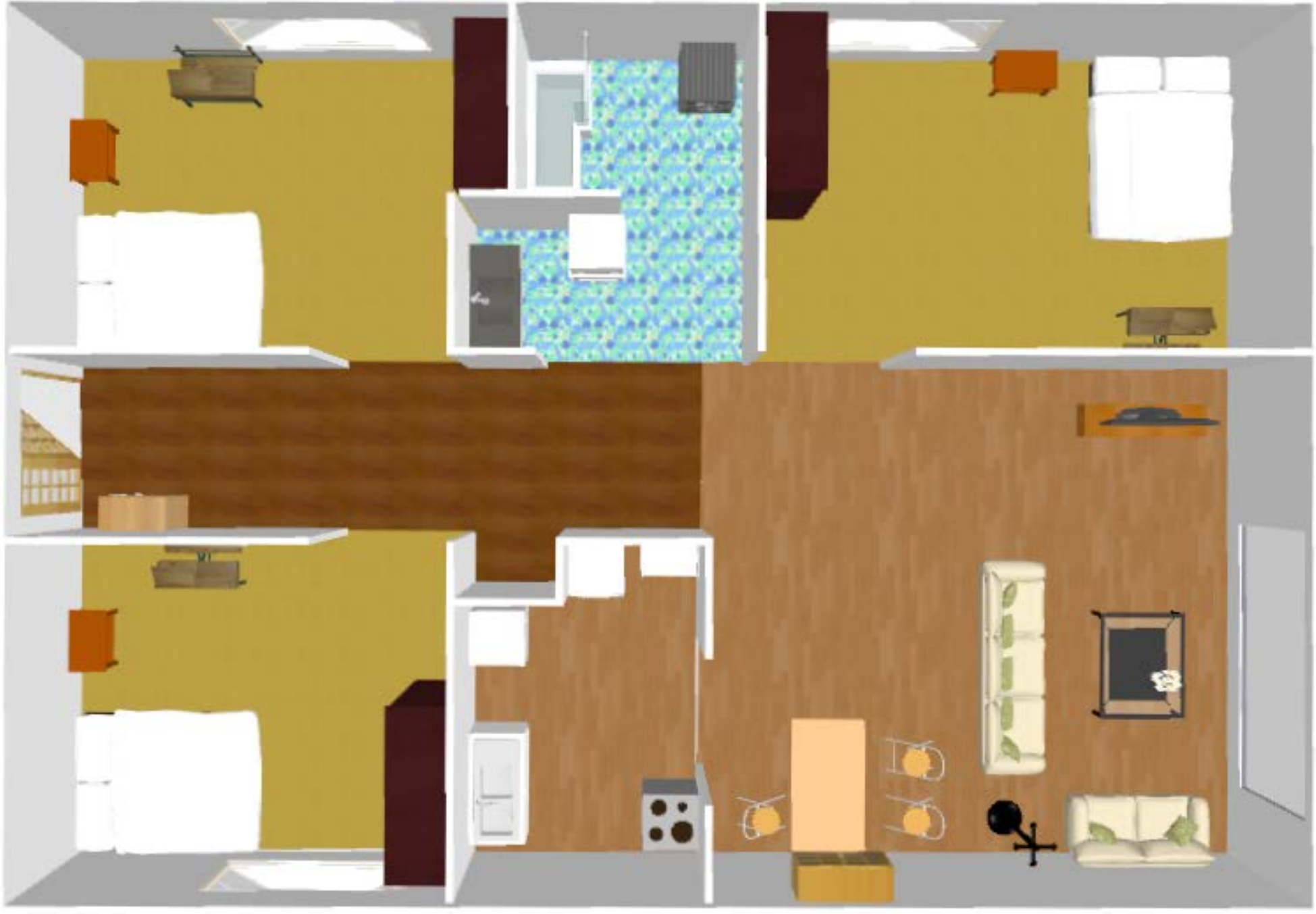}
        \label{fig:exp_2_environment_sim_8}}
   	\subfloat[Environment 9]{
       	\includegraphics[width=0.16\linewidth]{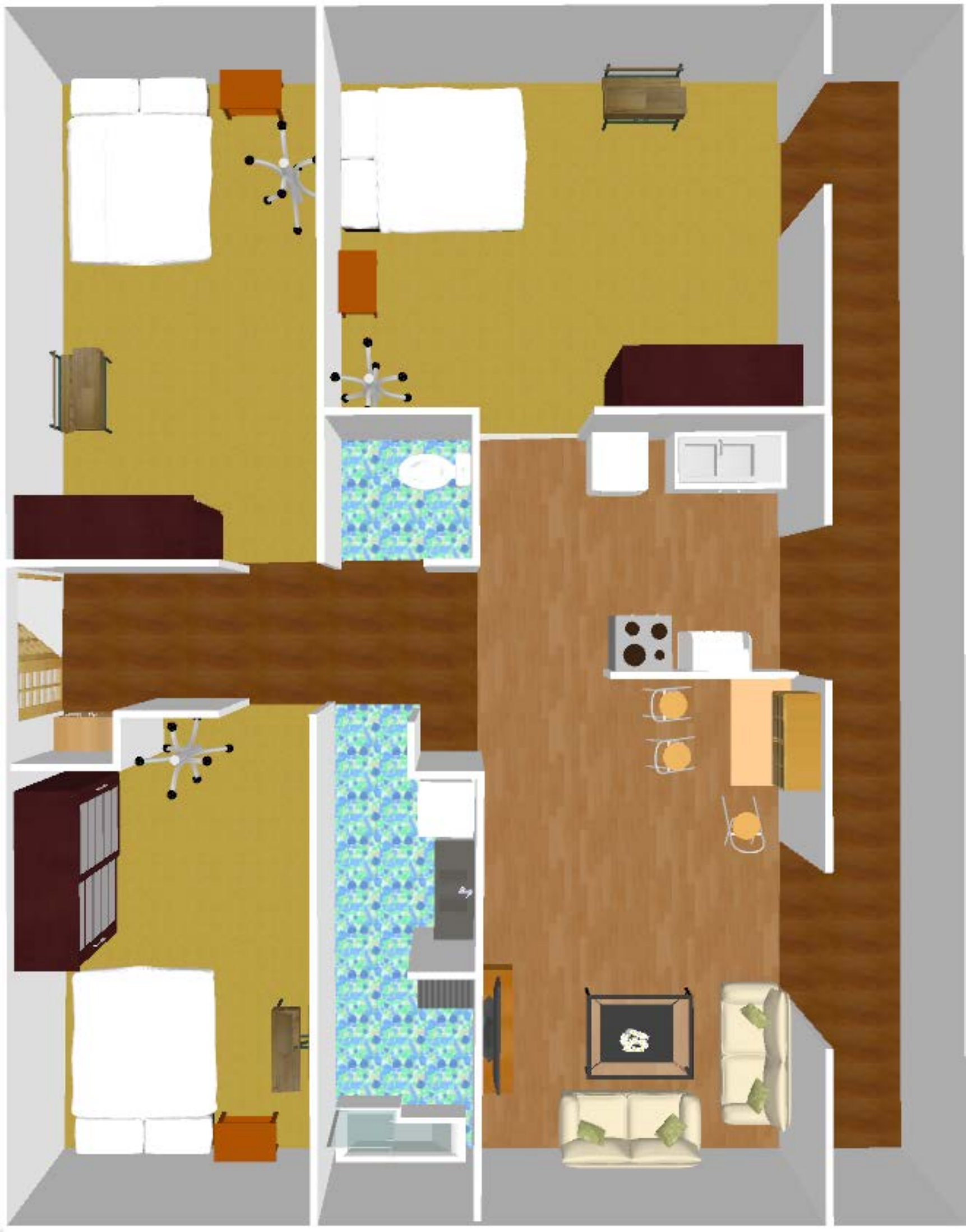}
        \label{fig:exp_2_environment_sim_9}}
   	\subfloat[Environment 10]{
       	\includegraphics[width=0.18\linewidth]{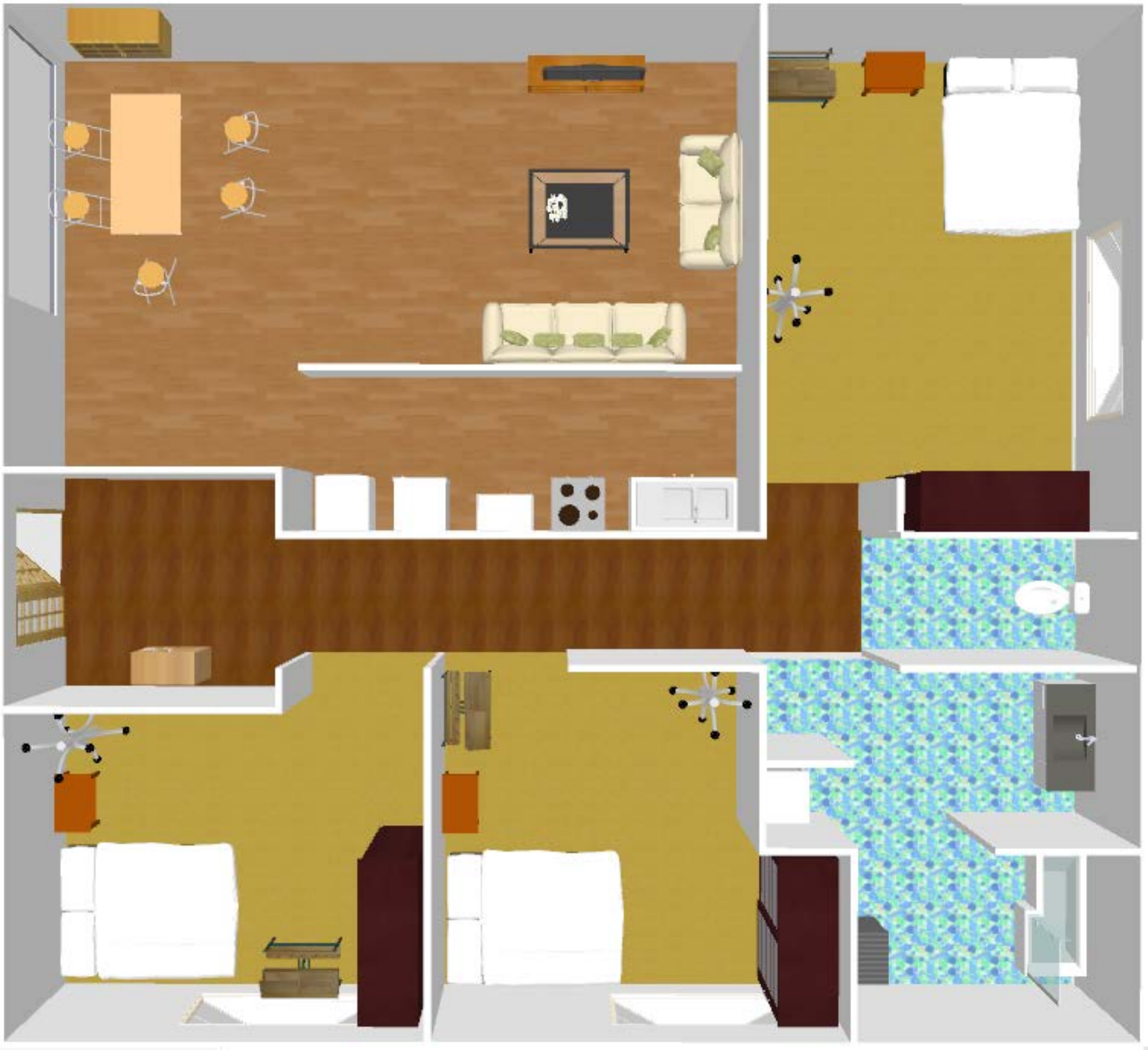}
        \label{fig:exp_2_environment_sim_10}}
	\caption{Home environments created with SIGVerse in Experiment I (ten environments)}
	\label{fig:exp_2_environment_10}
\end{figure}

\begin{figure}[tb]
	\centering
    \includegraphics[width=0.90\linewidth]{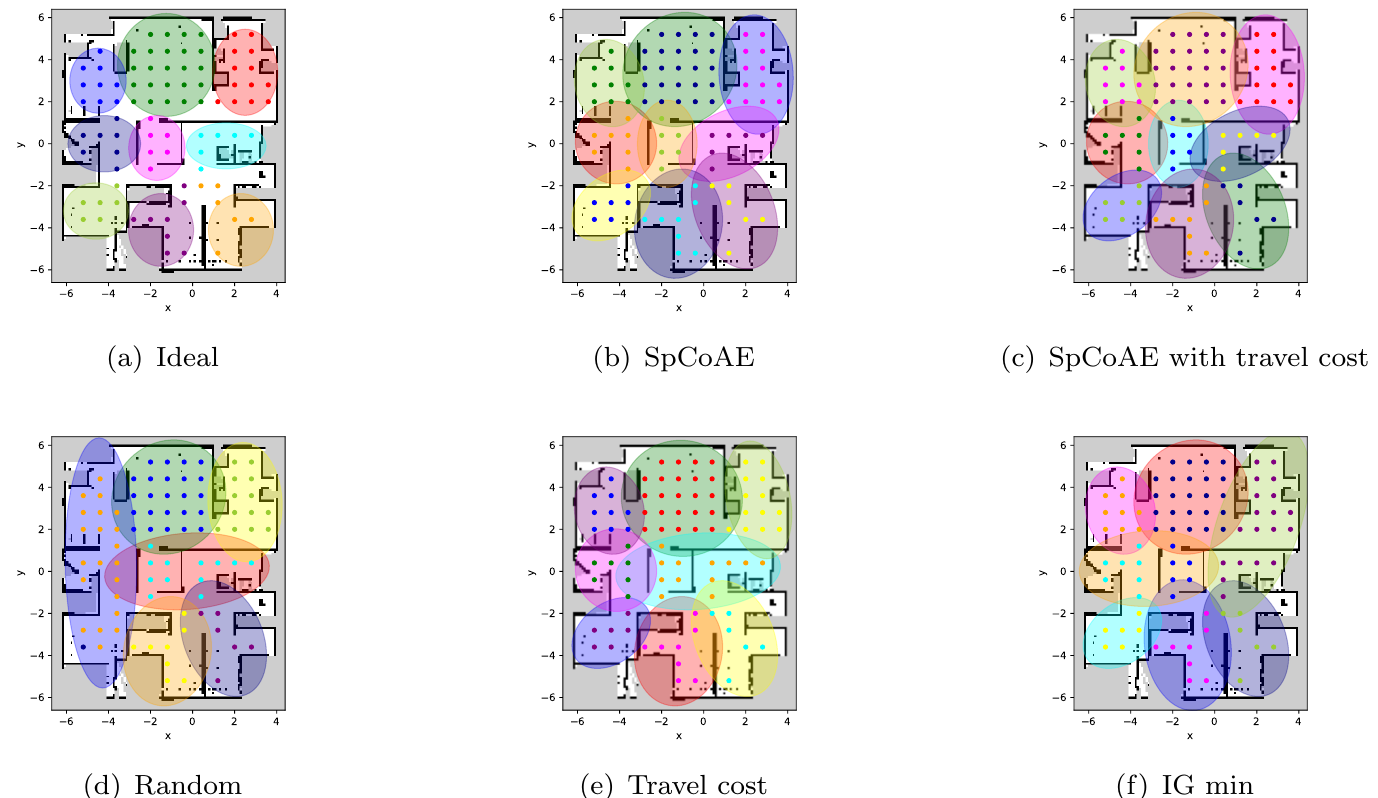}
		\caption{Examples of learning results in environment 1: Position distributions and indices~$C_{1:n}$ of spatial concepts drawn on the map.
    }
    \label{fig:exp_1_result_env1_map}
\end{figure}

\begin{figure}[tb]
	\centering
    \includegraphics[width=0.90\linewidth]{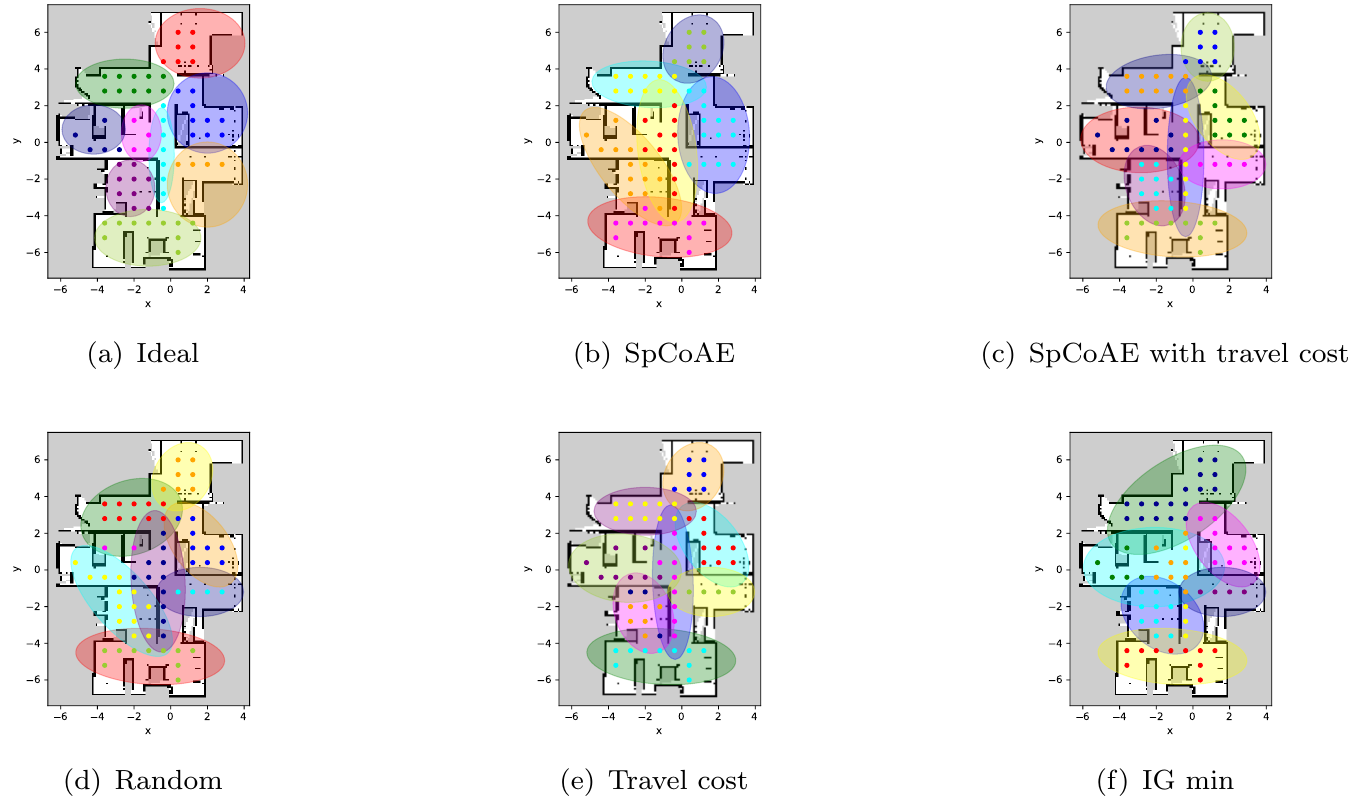}
		\caption{Examples of learning results in environment 2: Position distributions and indices~$C_{1:n}$ of spatial concepts drawn on the map.}
    \label{fig:exp_1_result_env2_map}
\end{figure}

\begin{figure}[tb]
	\centering
    \includegraphics[width=0.90\linewidth]{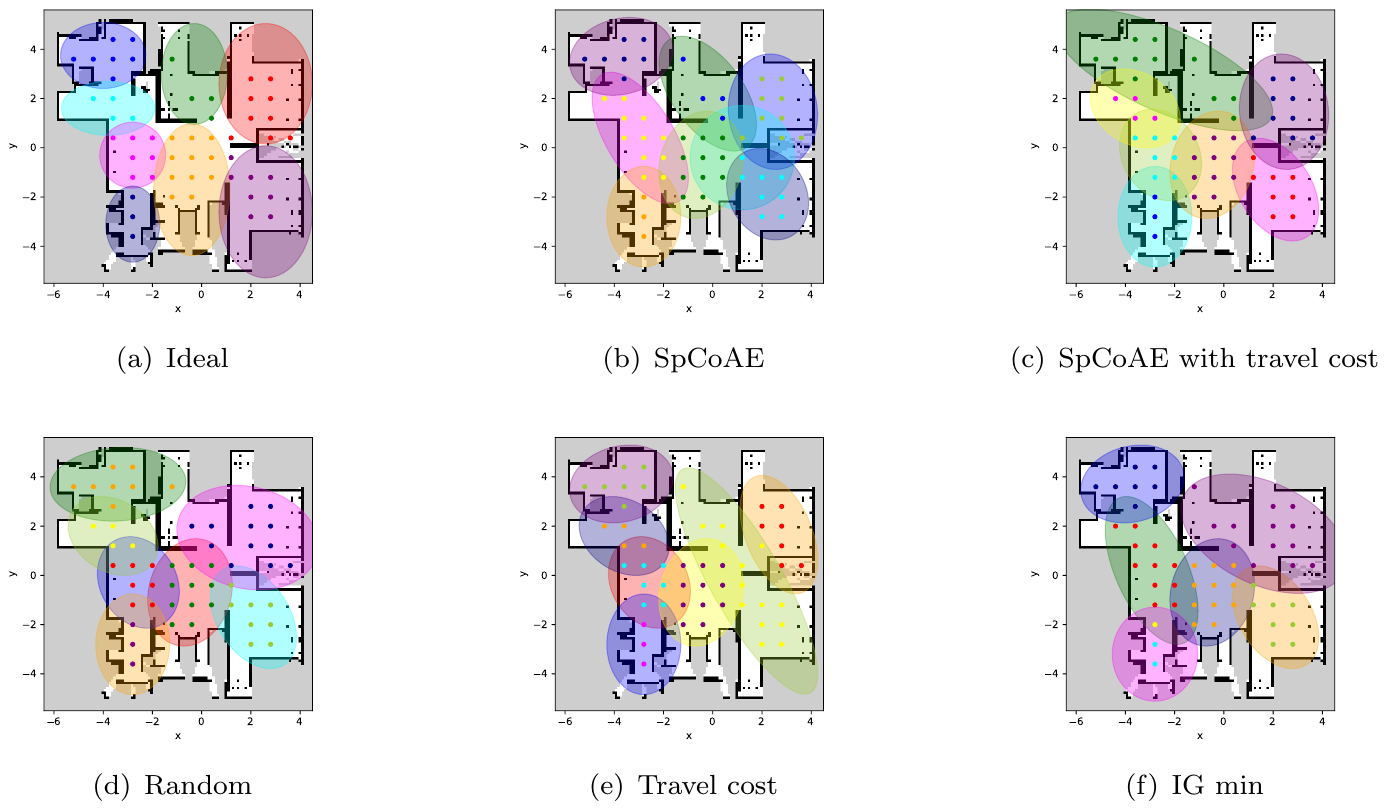}
		\caption{Examples of learning results in environment 3: Position distributions and indices~$C_{1:n}$ of spatial concepts drawn on the map.}
    \label{fig:exp_1_result_env3_map}
\end{figure}

\begin{figure}[tb]
	\centering
    \includegraphics[width=0.90\linewidth]{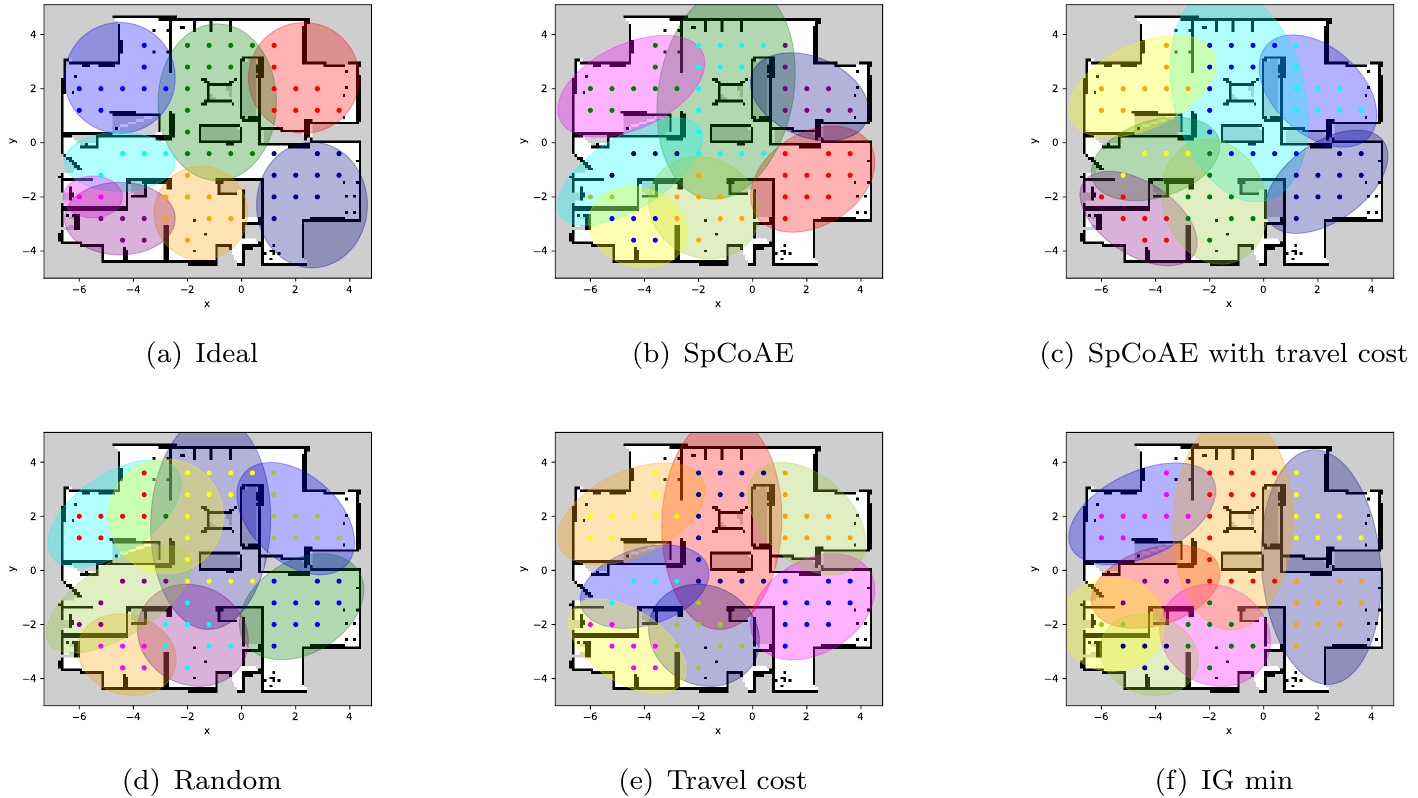}
		\caption{Examples of learning results in environment 4: Position distributions and indices~$C_{1:n}$ of spatial concepts drawn on the map.}
    \label{fig:exp_1_result_env4_map}
\end{figure}

\begin{figure}[tb]
	\centering
    \includegraphics[width=0.90\linewidth]{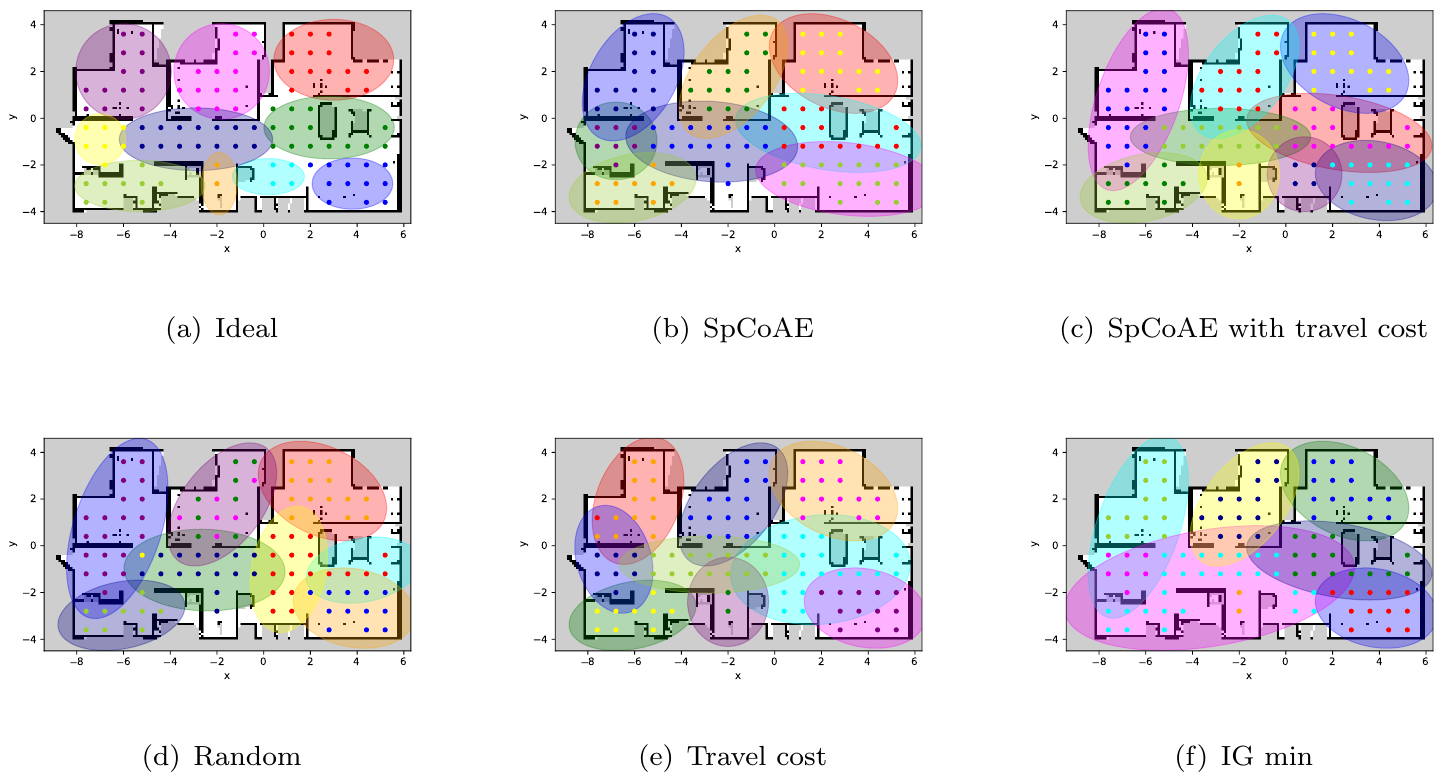}
		\caption{Examples of learning results in environment 5: Position distributions and indices~$C_{1:n}$ of spatial concepts drawn on the map.}
    \label{fig:exp_1_result_env5_map}
\end{figure}

\begin{figure}[tb]
	\centering
    \includegraphics[width=0.90\linewidth]{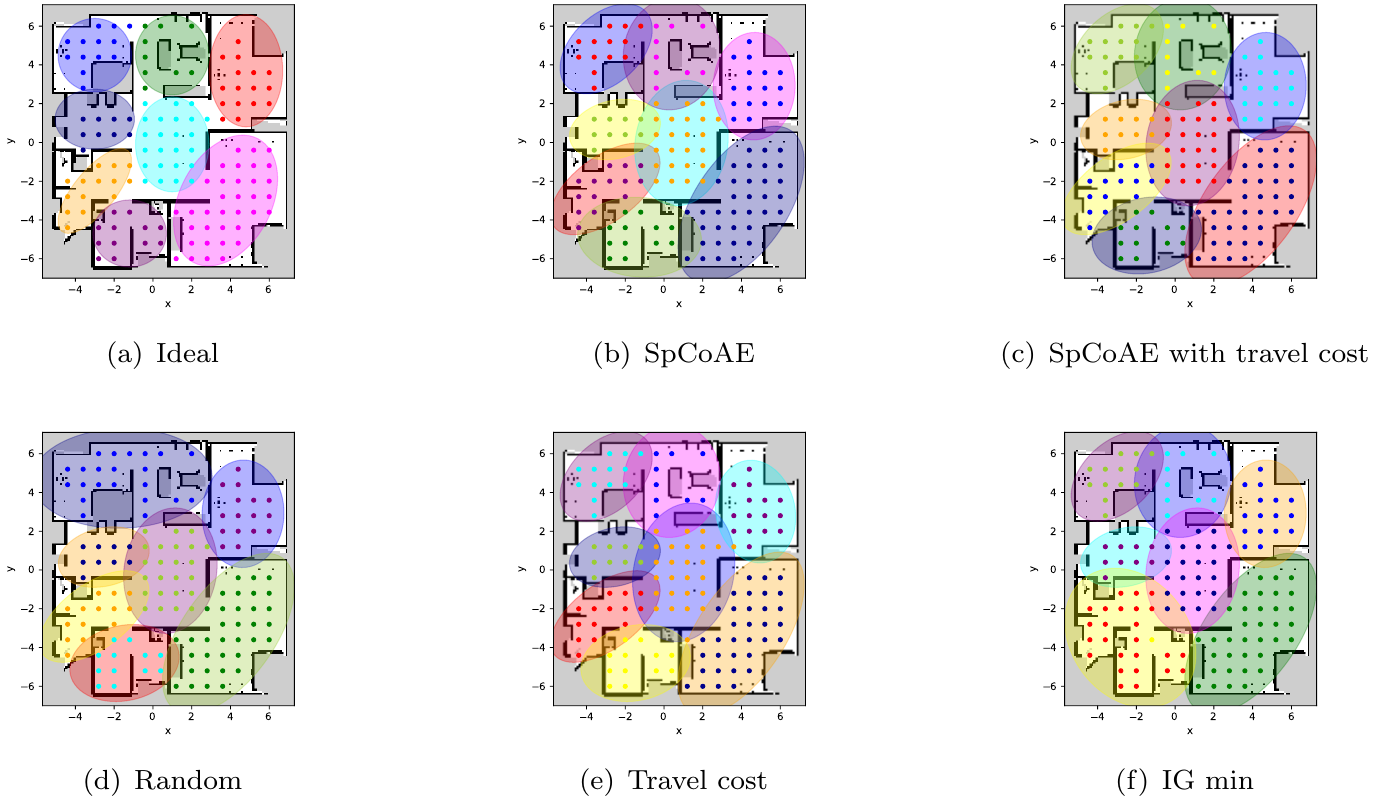}
		\caption{Examples of learning results in environment 6: Position distributions and indices~$C_{1:n}$ of spatial concepts drawn on the map.}
    \label{fig:exp_1_result_env6_map}
\end{figure}

\begin{figure}[tb]
	\centering
    \includegraphics[width=0.90\linewidth]{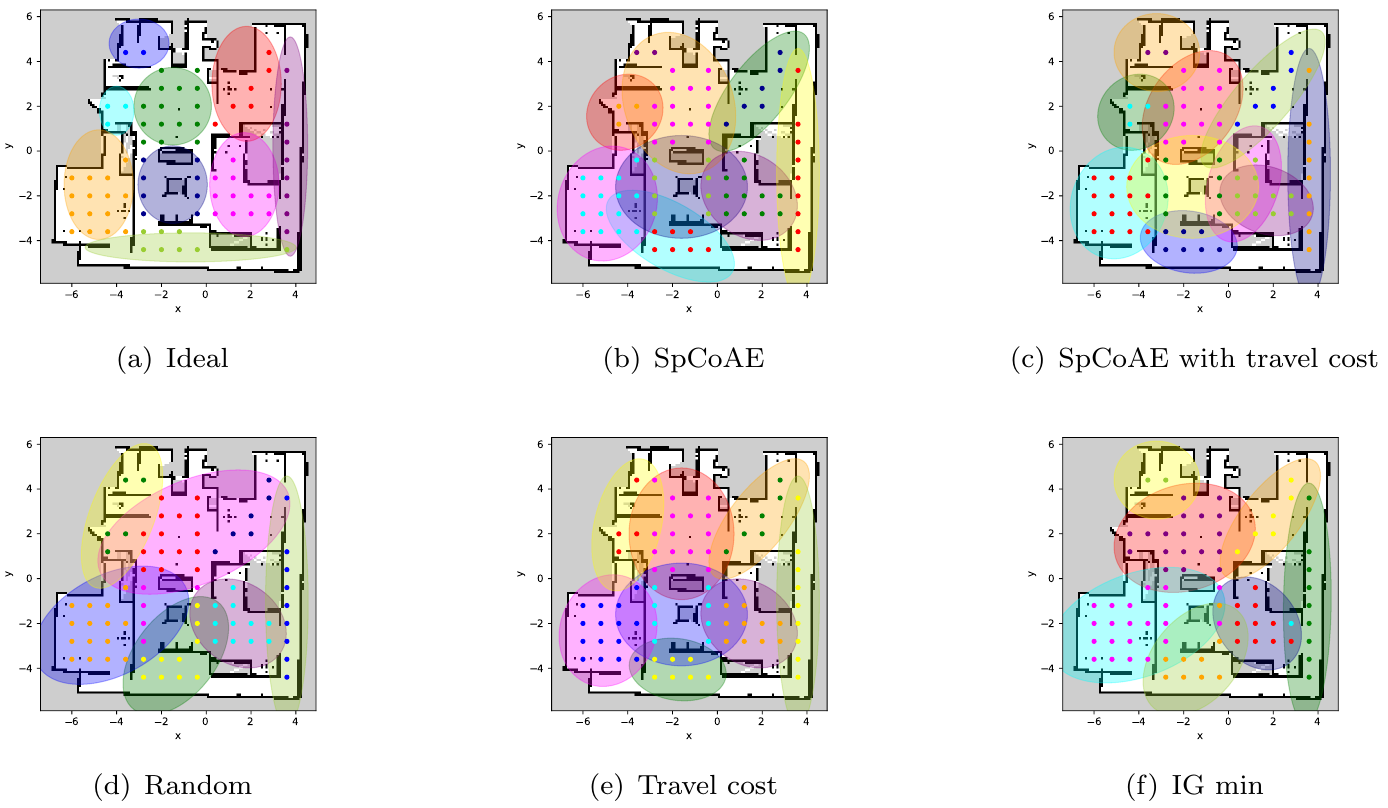}
		\caption{Examples of learning results in environment 7: Position distributions and indices~$C_{1:n}$ of spatial concepts drawn on the map.}
    \label{fig:exp_1_result_env7_map}
\end{figure}

\begin{figure}[tb]
	\centering
    \includegraphics[width=0.90\linewidth]{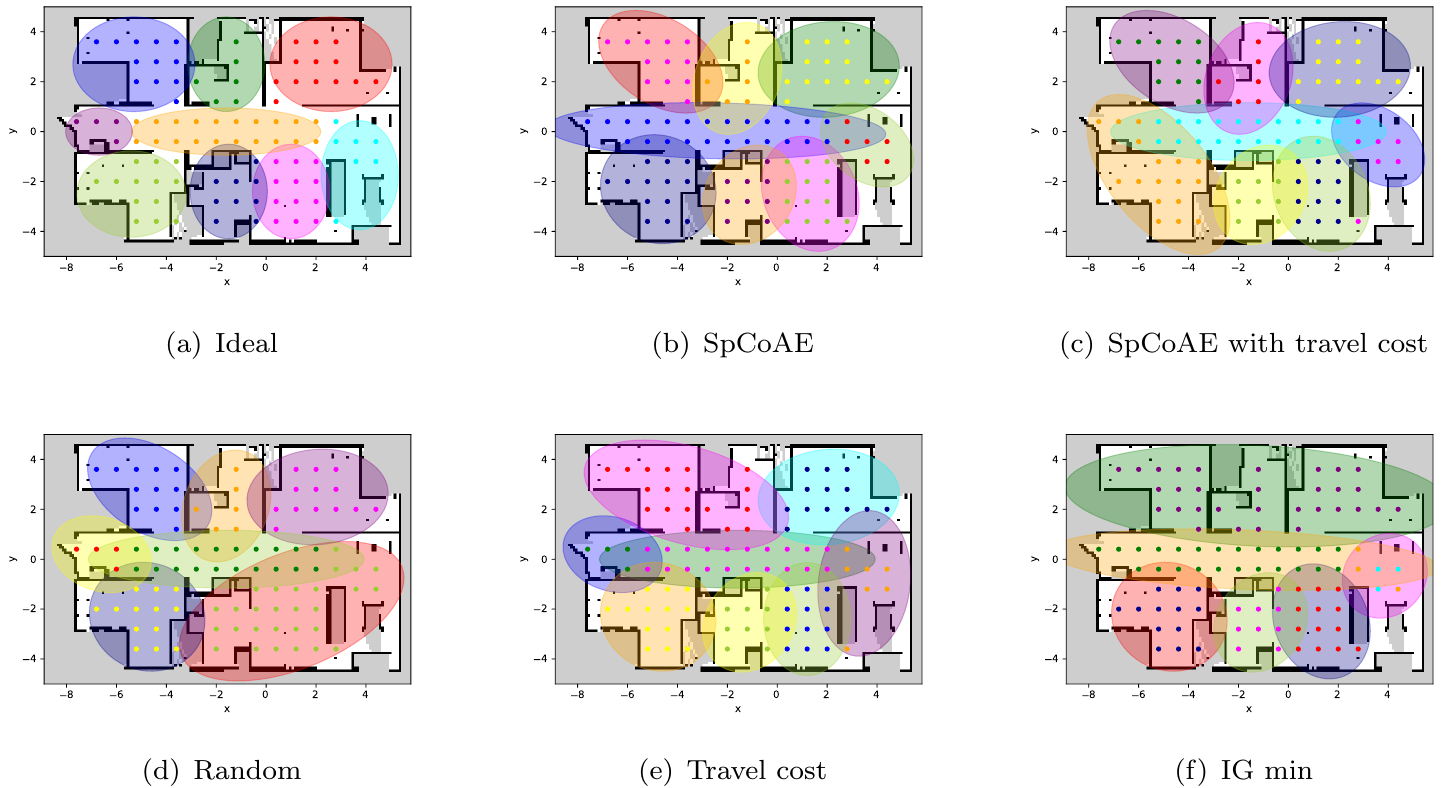}
		\caption{Examples of learning results in environment 8: Position distributions and indices~$C_{1:n}$ of spatial concepts drawn on the map.}
    \label{fig:exp_1_result_env8_map}
\end{figure}

\begin{figure}[tb]
	\centering
    \includegraphics[width=0.90\linewidth]{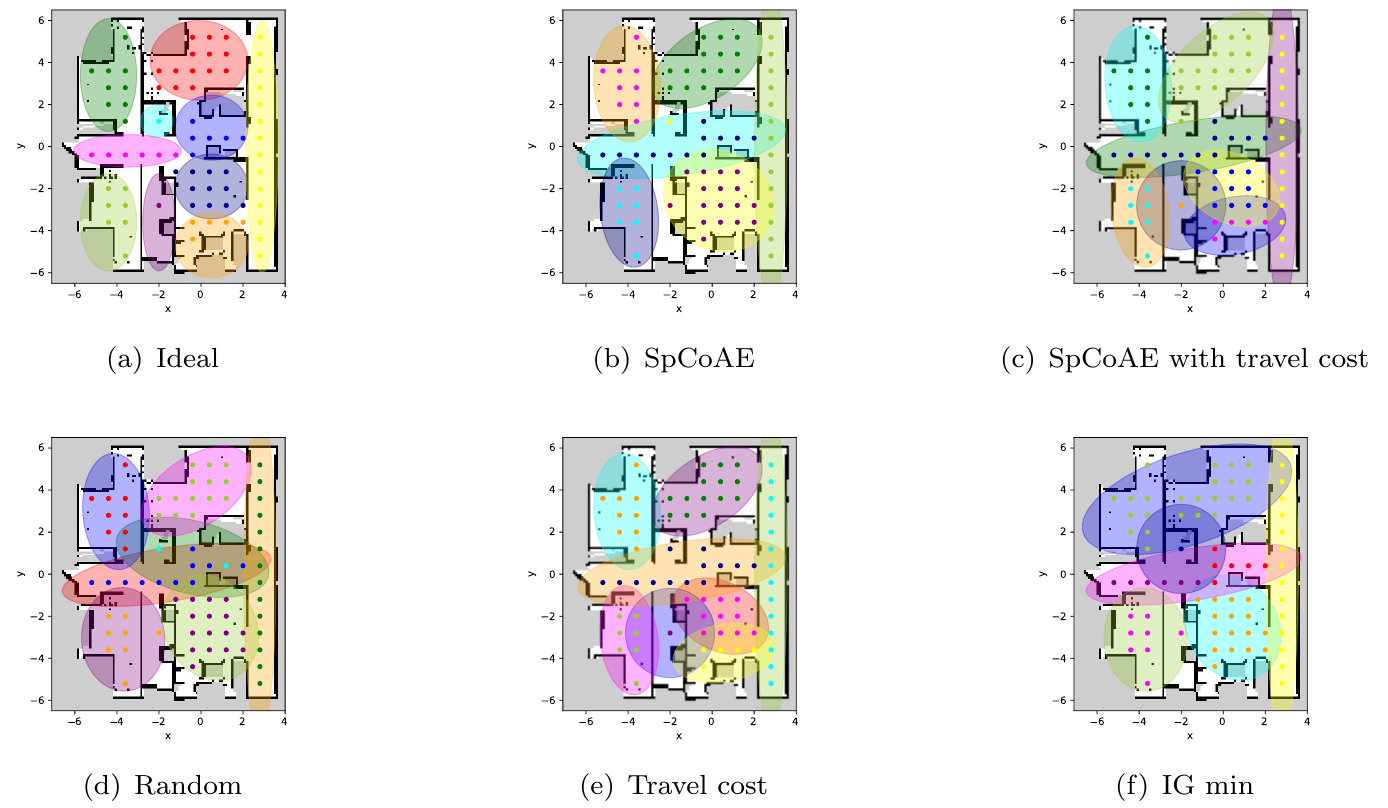}
		\caption{Examples of learning results in environment 9: Position distributions and indices~$C_{1:n}$ of spatial concepts drawn on the map.}
    \label{fig:exp_1_result_env9_map}
\end{figure}

\begin{figure}[tb]
	\centering
    \includegraphics[width=0.90\linewidth]{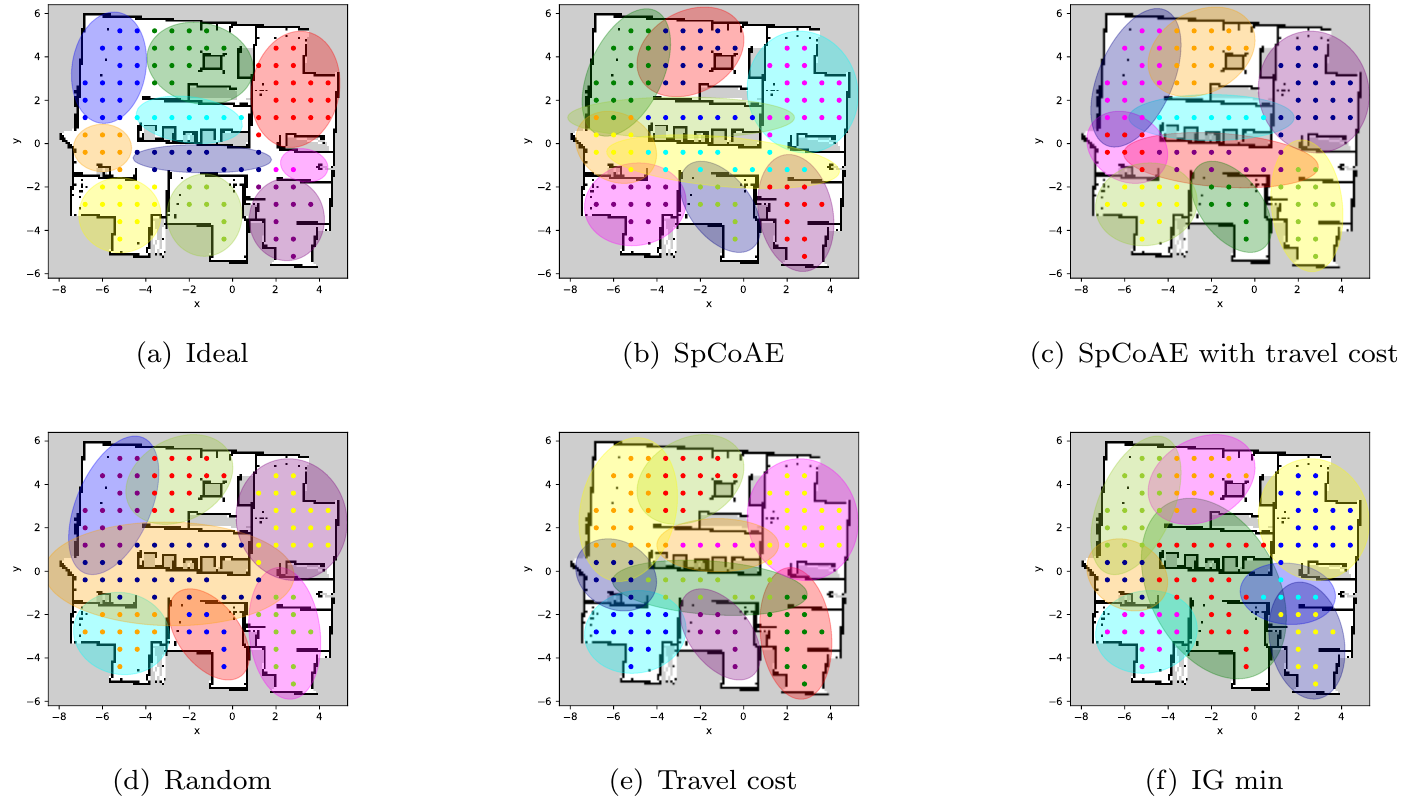}
		\caption{Examples of learning results in environment 10: Position distributions and indices~$C_{1:n}$ of spatial concepts drawn on the map.}
    \label{fig:exp_1_result_env10_map}
\end{figure}

\begin{figure}[tb]
	\centering
    \includegraphics[width=\linewidth]{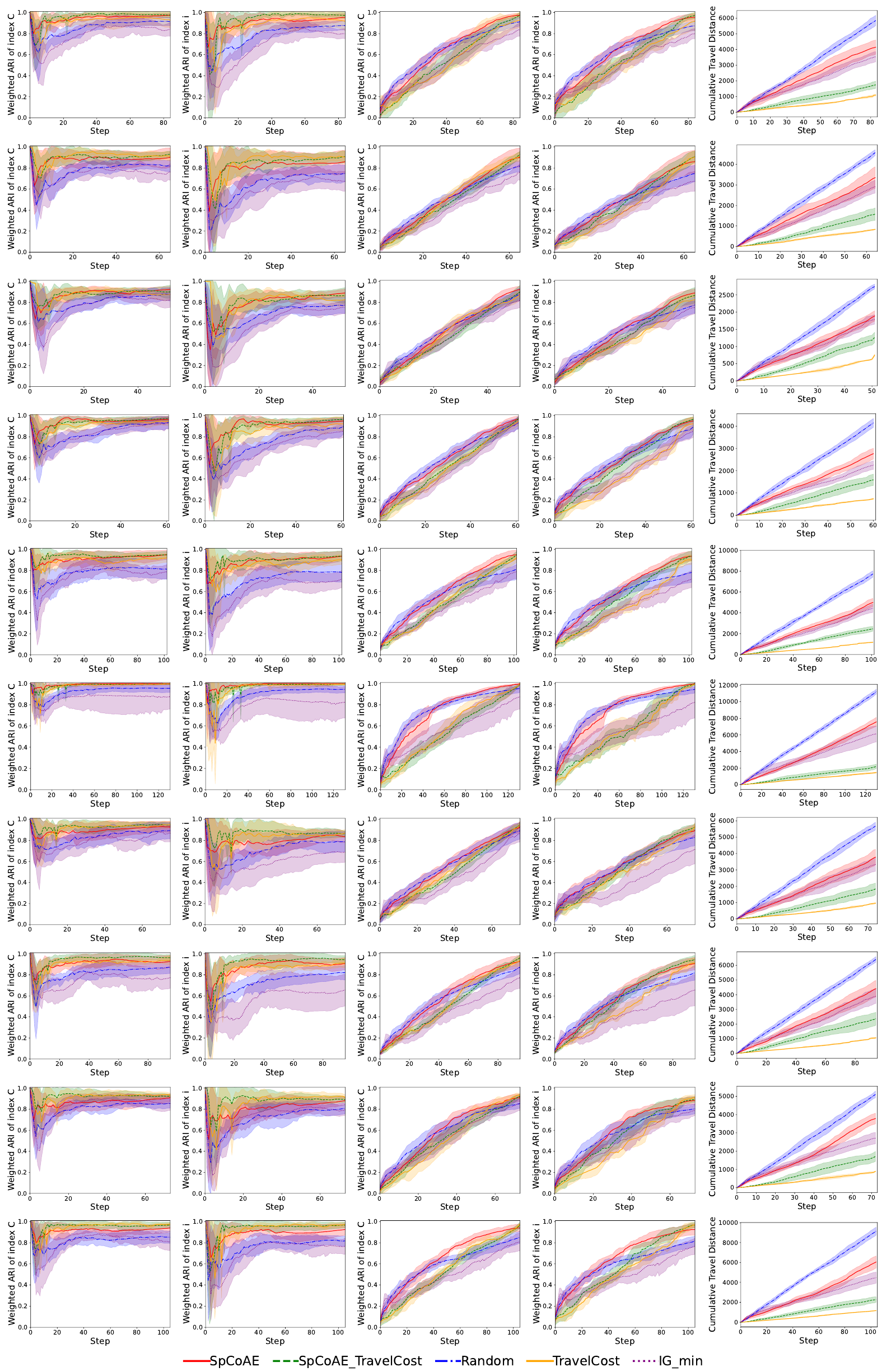}
   	\caption{
		Evaluation values in ten environments: from left to right, ARI (step-by-step) values of $C$ and $i$, ARI (predictive padding) values of $C$ and $i$, and cumulative travel distance.
    }
    \label{fig:exp_1_result_10env}
\end{figure}

In this section, we present the results of Experiment I using ten simulated home environments created in SIGVerse as the experimental environment.
The experimental setup is shown in Figure~\ref{fig:exp_2_environment_10}.
Maps previously created by SLAM were used in all the environments.
Examples of the learning results for all the methods are shown in Figures~\ref{fig:exp_1_result_env1_map}--\ref{fig:exp_1_result_env10_map}.
The index of the spatial concept is colored according to the position coordinates where the data are observed in each environment.
Each ellipse represents the position distribution.
The colors of each ellipse and candidate point were randomly determined for each trial.
Sub-figure (a) shows the ideal form envisioned by the tutor for reference.
Finally, the progress of the results of the evaluation values in the ten experimental environments is shown in Figure~\ref{fig:exp_1_result_10env}.

%


\end{document}